\documentclass[10pt,twocolumn,twoside]{IEEEtran}
\usepackage{cite,graphicx,epstopdf,float,url,amssymb,amsthm,threeparttable,multirow,algorithmic,nameref,paralist}
\usepackage[tight,footnotesize]{subfigure}
\usepackage[ruled,vlined]{algorithm2e}
\usepackage[usenames]{color}
\usepackage[normalem]{ulem}
\usepackage[cmex10]{amsmath}

\graphicspath{{Figures/}}

\theoremstyle{plain}
\newtheorem{lemma}{Lemma}
\newtheorem{theorem}{Theorem}
\newtheorem{corollary}{Corollary}

\theoremstyle{definition}
\newtheorem{definition}{Definition}
\theoremstyle{remark}
\newtheorem{remark}{Remark}

\newcommand{\E}{\mathbb{E}}

\newcommand{\cF}{\mathcal{F}}

\newcommand{\cG}{\mathcal{G}}

\newcommand{\cM}{\mathcal{M}}

\newcommand{\R}{\mathbb{R}}

\newcommand{\cS}{\mathcal{S}}

\newcommand{\cV}{\mathcal{V}}

\DeclareMathOperator*{\argmin}{arg\,min}
\DeclareMathOperator*{\argmax}{arg\,max}

\newcommand{\diag}{\mathrm{diag}}
\newcommand{\orth}{\mathrm{orth}}
\newcommand{\tr}{\mathrm{tr}}
\newcommand{\rank}{\mathrm{rank}}

\begin{document}
%
\title{Learning the nonlinear geometry of high-dimensional data: Models and algorithms}

\author{Tong Wu,~\IEEEmembership{Student Member,~IEEE,} and
        Waheed U. Bajwa,~\IEEEmembership{Senior Member,~IEEE}
\thanks{This work is supported in part by the NSF under grants CCF-1218942 and CCF-1453073,
by the Army Research Office under grant W911NF-14-1-0295, and by an Army Research Lab Robotics CTA subaward.
Preliminary versions of parts of this work have been presented at the IEEE
International Conference on Acoustics, Speech and Signal Processing
(ICASSP 2014) \cite{WuB.ICASSP2014}, IEEE Workshop on Statistical Signal
Processing (SSP 2014) \cite{WuB.SSP2014}, and IEEE
International Conference on Acoustics, Speech and Signal Processing
(ICASSP 2015) \cite{WuB.ICASSP2015}.}%
\thanks{The authors are with the Department of Electrical and Computer Engineering,
Rutgers University, Piscataway, NJ 08854, USA (E-mails: {\tt \{tong.wu.ee, waheed.bajwa\}@rutgers.edu}).}}

%

\maketitle

\begin{abstract}
Modern information processing relies on the axiom that high-dimensional
data lie near low-dimensional geometric structures. This paper revisits the
problem of data-driven learning of these geometric structures and puts
forth two new nonlinear geometric models for data describing ``related''
objects/phenomena. The first one of these models straddles the two extremes
of the subspace model and the union-of-subspaces model, and is termed the
\emph{metric-constrained union-of-subspaces} (MC-UoS) model. The second one
of these models---suited for data drawn from a mixture of nonlinear
manifolds---generalizes the kernel subspace model, and is termed the
\emph{metric-constrained kernel union-of-subspaces} (MC-KUoS) model.
The main contributions of this paper in this regard include the following.
First, it motivates and formalizes the problems of MC-UoS and MC-KUoS
learning. Second, it presents algorithms that efficiently learn an MC-UoS
or an MC-KUoS underlying data of interest. Third, it extends these
algorithms to the case when parts of the data are missing. Last, but not
least, it reports the outcomes of a series of numerical experiments
involving both synthetic and real data that demonstrate the superiority of
the proposed geometric models and learning algorithms over existing
approaches in the literature. These experiments also help clarify the
connections between this work and the literature on (subspace and kernel
$k$-means) clustering.
\end{abstract}


\section{Introduction}
\label{sec:intro}

\IEEEPARstart{W}{e} have witnessed an explosion in data generation in the
last decade or so. Modern signal processing, machine learning and statistics
have been relying on a fundamental maxim of information processing to cope
with this data explosion. This maxim states that while real-world data might
lie in a high-dimensional Hilbert space, relevant information within them
almost always lies near low-dimensional geometric structures embedded in the
Hilbert space. Knowledge of these low-dimensional geometric structures not
only improves the performance of many processing tasks, but it also helps
reduce computational and communication costs, storage requirements, etc.

Information processing literature includes many models for geometry of
high-dimensional data, which are then utilized for better performance in
numerous applications, such as dimensionality reduction and data compression
\cite{Hotelling.JEP1933,CoxC.2000,RoweisS.Science2000,EladGK.TIP2007,BaraniukCW.PI2010},
denoising \cite{MikaSSMSR.NIPS1998,EladA.TIP2006}, classification
\cite{TurkP.CVPR1991,SwetsW.PAMI1996,WrightYGSM.PAMI2009,MairalBP.PAMI2012}, and motion segmentation
\cite{RaoTVM.PAMI2010,ElhamifarV.PAMI2013}. These geometric models broadly fall into
two categories, namely, linear models \cite{Hotelling.JEP1933,Harman.1976,SwetsW.PAMI1996} and
nonlinear models \cite{ScholkopfSM.NC1998,RoweisS.Science2000,AharonEB.TSP2006,EldarM.TIT2009,ElhamifarV.PAMI2013}.
A further distinction can be made within each of these two categories
depending upon whether the models are prespecified \cite{MarcellinGBB.DCC2000,StarckCD.TIP2002} or learned from
the data themselves \cite{AharonEB.TSP2006,EladA.TIP2006,ZhangSL.CVW2009,GowreesunkerT.TSP2010,BalzanoSRN.SSP2012,ElhamifarV.PAMI2013}. Our focus in this paper is on the latter case, since data-driven learning of geometric models is known to outperform
prespecified geometric models \cite{EladA.TIP2006,HongWHM.TIP2006}.

Linear models, which dictate that data lie near a low-dimensional subspace of
the Hilbert space, have been historically preferred within the class of
data-driven models due to their simplicity. These models are commonly studied
under the rubrics of \emph{principal component analysis} (PCA)
\cite{Pearson.PM1901,Hotelling.JEP1933}, Karhunen--Lo\`{e}ve transform
\cite{Fukunaga.1990}, factor analysis \cite{Harman.1976}, etc. But real-world
data in many applications tend to be nonlinear. In order to better
capture the geometry of data in such applications, a few nonlinear
generalizations of data-driven linear models that remain computationally
feasible have been investigated in the last two decades. One of the most
popular generalizations is the nonlinear manifold model
\cite{ScholkopfSM.Advance1999,RoweisS.Science2000,BelkinN.NC2003,BaraniukCW.PI2010}.
The (nonlinear) manifold model can also be considered as the
\emph{kernel subspace} model, which dictates that a mapping of the data to a
higher- (possibly infinite-) dimensional Hilbert space lies near a
low-dimensional subspace \cite{HamLMS.ICML2004}. Data-driven learning of
geometric models in this case is commonly studied under the moniker of
\emph{kernel PCA} (KPCA) \cite{ScholkopfSM.Advance1999}. Another one of the
most popular generalizations of linear models is the
\emph{union-of-subspaces} (UoS) \big(resp.,
\emph{union-of-affine-subspaces} (UoAS)\big) model, which dictates that data
lie near a mixture of low-dimensional subspaces (resp., affine subspaces) in
the ambient Hilbert space. Data-driven learning of the UoS model is commonly
carried out under the rubrics of generalized PCA \cite{VidalMS.PAMI2005},
dictionary learning \cite{AharonEB.TSP2006,Zelnik-ManorRE.TSP2012}, and
subspace clustering \cite{SoltanolkotabiC.AS2012,DyerSB.JMLR2013,ElhamifarV.PAMI2013,HeckelB.arxiv2013,SoltanolkotabiEC.AS2014}.
On the other hand, data-driven learning of the UoAS model is often studied
under the umbrella of hybrid linear modeling \cite{ZhangSWL.IJCV2012},
mixture of factor analyzers \cite{GhahramaniH.tech1997}, etc.

In the literature, encouraging results have been reported for both the UoS
and the kernel subspace models in the context of a number of applications
\cite{MikaSSMSR.NIPS1998,WrightMMSHY.PI2010,MairalBP.PAMI2012,ElhamifarV.PAMI2013}.
But there remains a lot of room for improvement in both these models. The
canonical UoS model, for example, does not impose any constraint on the
collection of subspaces underlying data of interest. On the other hand, one
can intuit that subspaces describing ``similar'' data should have some
``relation'' on the Grassmann manifold. The lack of any a priori constraint
during learning on the subspaces describing ``similar'' data has the potential to
make different methods for UoS learning susceptible to errors due to low
\emph{signal-to-noise ratio} (SNR), outliers, missing data, etc. Another
limitation of the UoS model is the individual linearity of its constituent
subspaces, which limits its usefulness for data drawn from a
nonlinear manifold \cite{ScholkopfSM.Advance1999}. On the other hand, while
the kernel subspace model can handle manifold data, a single kernel
subspace requires a large dimension to capture the richness of data drawn
from a mixture of nonlinear manifolds.

Our goal in this paper is to improve the state-of-the-art data-driven
learning of geometric data models for both complete and missing data
describing similar phenomenon. We are in particular interested in
learning models for data that are either mildly or highly nonlinear.
Here, we are informally using the terms ``mildly nonlinear'' and
``highly nonlinear.'' Heuristically, nonlinear data that cannot be
represented through a mixture of linear components should be deemed ``highly
nonlinear.'' Our key objective in this regard is overcoming the
aforementioned limitations of the UoS model and the kernel subspace model for
mildly nonlinear data and highly nonlinear data, respectively.

\subsection{Our Contributions and Relation to Other Work}
\label{ssec:contribute}

One of our main contributions is introduction of a novel geometric model,
termed \emph{metric-constrained union-of-subspaces} (MC-UoS) model, for
mildly nonlinear data describing similar phenomenon. Similar to the
canonical UoS model, the MC-UoS model also dictates that data lie near a
union of low-dimensional subspaces in the ambient space. But the key
distinguishing feature of the MC-UoS model is that it also forces its
constituent subspaces to be close to each other according to a metric defined
on the Grassmann manifold. In this paper, we formulate the MC-UoS learning
problem for a particular choice of the metric and derive three novel
iterative algorithms for solving this problem. The first one of these
algorithms operates on complete data, the second one deals with the case of
unknown number and dimension of subspaces, while the third one carries out
MC-UoS learning in the presence of missing data.

One of our other main contributions is extension of our MC-UoS model for
highly nonlinear data. This model, which can also be considered a
generalization of the kernel subspace model, is termed
\emph{metric-constrained kernel union-of-subspaces} (MC-KUoS) model. The
MC-KUoS model asserts that mapping of data describing similar
phenomenon to some higher-dimensional Hilbert space (also known as the
\emph{feature space}) lies near a mixture of subspaces in the feature space
with the additional constraint that the individual subspaces are also close
to each other in the feature space. In this regard, we formulate the MC-KUoS
learning problem using the \emph{kernel trick} \cite{ScholkopfSM.NC1998},
which avoids explicit mapping of data to the feature space. In addition, we
derive two novel iterative algorithms that can carry out MC-KUoS learning in
the presence of complete data and missing data.

Our final contribution involves carrying out a series of numerical
experiments on both synthetic and real data to justify our heuristics for the
two models introduced in this paper. Our main focus in these experiments is
learning the geometry of (training) data describing similar
phenomenon in the presence of additive, white Gaussian noise and
missing entries. In the case of real data, we demonstrate the
superiority of the proposed algorithms by
focusing on the tasks of denoising of (test) data and clustering of data having
either complete or missing entries. (Other applications of our models will
be investigated in future works.) Our results confirm the superiority of our
models in comparison to a number of state-of-the-art approaches under both
the UoS and the kernel subspace models
\cite{ScholkopfSM.Advance1999,HoYLLK.CVPR2003,BalzanoSRN.SSP2012,Zelnik-ManorRE.TSP2012,ElhamifarV.PAMI2013,
HeckelB.arxiv2013,SoltanolkotabiEC.AS2014}.

We conclude this discussion by pointing out that our work is
not only related to the traditional literature on geometry learning, but it
also has connections to the literature on clustering \cite{HoYLLK.CVPR2003,ElhamifarV.PAMI2013,HeckelB.arxiv2013,SoltanolkotabiEC.AS2014}. Specifically, the mixture components within our two models can be treated as different clusters
within the data and the outputs of our algorithms automatically lead us to
these clusters. Alternatively, one could approach the MC-UoS/MC-KUoS learning
problem by first clustering the data and then learning the individual
subspaces in the ambient/feature space. However, numerical
experiments confirm that our algorithms perform better than such heuristic
approaches.

\subsection{Notation and Organization}
\label{ssec:organote}

Throughout the paper, we use bold lower-case and bold upper-case
letters to represent vectors/sets and matrices, respectively. The $i$-th
element of a vector/set $\mathbf{v}$ is denoted by $\mathbf{v}_{(i)}$, while
$a_{i,j}$ denotes the $(i,j)$-th element of a matrix $\mathbf{A}$. The
$m$-dimensional zero vector is denoted by $\boldsymbol{0}_m$ and the $m
\times m$ identity matrix is denoted by $\mathbf{I}_{m}$. Given a set
$\mathbf{\Omega}$, $[\mathbf{A}]_{\mathbf{\Omega},:}$ (resp.,
$[\mathbf{v}]_{\mathbf{\Omega}}$) denotes the submatrix of $\mathbf{A}$
(resp., subvector of $\mathbf{v}$) corresponding to the rows of $\mathbf{A}$
(resp., entries of $\mathbf{v}$) indexed by $\mathbf{\Omega}$. Given two sets
$\mathbf{\Omega}_1$ and $\mathbf{\Omega}_2$,
$[\mathbf{A}]_{\mathbf{\Omega}_1, \mathbf{\Omega}_2}$ denotes the submatrix
of $\mathbf{A}$ corresponding to the rows and columns indexed by
$\mathbf{\Omega}_1$ and $\mathbf{\Omega}_2$, respectively. Finally,
$(\cdot)^T$ and $\tr(\cdot)$ denote transpose and trace operations,
respectively, while the Frobenius norm of a matrix $\mathbf{A}$ is denoted by
${\|\mathbf{A}\|_F}$ and the $\ell_{2}$ norm of a vector $\mathbf{v}$ is
represented by $\|\mathbf{v}\|_{2}$.

The rest of the paper is organized as follows. In Sec.~\ref{sec:probform},
we formally define the metric-constrained union-of-subspaces (MC-UoS)
model and mathematically formulate the data-driven learning problems studied in this paper. Sec.~\ref{sec:linearsolver}
presents algorithms for MC-UoS learning in the presence of complete and missing data.
Sec.~\ref{sec:kernelsolver} gives the details of two algorithms for learning of an MC-UoS
in the feature space, corresponding to the cases of complete and missing data. We then
present some numerical results in Sec.~\ref{sec:experiment}, which is followed by concluding remarks in Sec.~\ref{sec:conclusion}.


\section{Problem Formulation}
\label{sec:probform}

In this section, we mathematically formulate the two problems of learning the
geometry of mildly and highly nonlinear data from training
examples. Both of our problems rely on the notion of a metric-constrained
union-of-subspaces (MC-UoS), one in the ambient space and the other in the
feature space. We therefore first begin with a mathematical characterization of the MC-UoS
model.

Recall that the canonical UoS model asserts data in an $m$-dimensional
ambient space can be represented through a union of $L$ low-dimensional
subspaces \cite{LuD.TSP2008,BaraniukCW.PI2010}: $\cM_L = \bigcup_{\ell=1}^{L}
\cS_{\ell}$, where $\cS_{\ell}$ is a subspace of $\R^m$. In here, we make the
simplified assumption that all subspaces in $\cM_L$ have the same dimension,
i.e., $\forall \ell$, $\text{dim}(\cS_{\ell}) = s \ll m$. In this case, each
subspace $\cS_{\ell}$ corresponds to a point on the Grassmann manifold
$\cG_{m,s}$, which denotes the set of all $s$-dimensional subspaces of
$\R^m$. While the canonical UoS model allows $\cS_{\ell}$'s to be arbitrary
points on $\cG_{m,s}$, the basic premise of the MC-UoS model is that
subspaces underlying \emph{similar} signals likely form a ``cluster'' on the
Grassmann manifold. In order to formally capture this intuition, we make use
of a distance metric on $\cG_{m,s}$ and define an MC-UoS according to that
metric as follows.

\begin{definition}   \label{def:MC-UoS}
\textbf{(Metric-Constrained Union-of-Subspaces.)} A UoS $\cM_L = \bigcup_{\ell=1}^{L} \cS_{\ell}$ is said to be constrained with respect to a metric
$d_u: \cG_{m,s} \times \cG_{m,s} \to [0,\infty)$ if $\max_{\ell,p: \ell \neq p} d_u(\cS_{\ell},\cS_p) \leq \epsilon$ for some positive constant $\epsilon$.
\end{definition}

The metric we use in this paper to measure distances between subspaces is
based on the Hausdorff distance between a vector and a subspace, which was
first defined in \cite{WangWF.PR2006}. Specifically, if $\mathbf{D}_{\ell} \in
\R^{m \times s}$ and $\mathbf{D}_p \in \R^{m \times s}$ denote
orthonormal bases of subspaces $\cS_{\ell}$ and $\cS_p$, respectively, then
\begin{align}    \label{eqn:dmetric}
d_u(\cS_{\ell},\cS_p) & =  \sqrt{ s - \tr(\mathbf{D}_{\ell}^{T} \mathbf{D}_{p} \mathbf{D}_{p}^{T} \mathbf{D}_{\ell}) } \nonumber \\
& = \| \mathbf{D}_{\ell} - P_{\cS_p} \mathbf{D}_{\ell} \|_{F},
\end{align}
where $P_{\cS_p}$ denotes the projection operator onto the subspace
$\cS_p$: $P_{\cS_p} = \mathbf{D}_{p} \mathbf{D}_{p}^T$. It is easy to convince oneself that
$d_u(\cdot,\cdot)$ in \eqref{eqn:dmetric} is invariant to the choice of
orthonormal bases of the two subspaces, while it was formally shown to be a
metric on $\cG_{m,s}$ in \cite{SunWF.PR2007}. Note that $d_u(\cdot,\cdot)$ in
\eqref{eqn:dmetric} is directly related to the concept of \emph{principal
angles} between two subspaces. Given two subspaces $\cS_{\ell}, \cS_p$ and
their orthonormal bases $\mathbf{D}_{\ell}, \mathbf{D}_{p}$, the cosines of the principal
angles $\cos(\theta_{\ell,p}^j)$, $j = 1,\dots,s$, between $\cS_{\ell}$ and
$\cS_p$ are defined as the ordered singular values of $\mathbf{D}_{\ell}^{T} \mathbf{D}_{p}$
\cite{SoltanolkotabiEC.AS2014}. It therefore follows that $d_u(\cS_{\ell},\cS_p) = \sqrt{s -
\sum_{j=1}^s \cos^2(\theta_{\ell,p}^j)}$. We conclude our discussion of the
MC-UoS model by noting that other definitions of metrics on the Grassmann
manifold exist in the literature that are based on different manipulations of
$\cos(\theta_{\ell,p}^j)$'s \cite{WolfS.CVPR2003}. In this paper, however, we focus only
on \eqref{eqn:dmetric} due to its ease of computation.

\subsection{Geometry Learning for Mildly Nonlinear Data}
\label{ssec:problinear}

Our first geometry learning problem corresponds to the case of
high-dimensional data that lie near an MC-UoS $\cM_L$ in the ambient space
$\R^m$. We are using the qualifier ``mildly nonlinear'' for such data since
individual components of these data are being modeled in a linear fashion. In
terms of a formal characterization, we assume access to a collection of $N$
noisy training samples, $\mathbf{Y}= [ \mathbf{y}_1, \dots, \mathbf{y}_N ]
\in \R^{m \times N}$, such that every sample $\mathbf{y}_i$ can be
expressed as $\mathbf{y}_i = \mathbf{x}_i + \boldsymbol{\xi}_i$ with
$\mathbf{x}_i$ belonging to one of the $\cS_{\ell}$'s in $\cM_L$ and
$\boldsymbol{\xi}_i \sim \mathcal{N}(\boldsymbol{0},
(\sigma_{tr}^2/m)\mathbf{I}_m)$ denoting additive noise. We assume
without loss of generality throughout this paper that
$\|\mathbf{x}_i\|_2^2=1$, which results in training SNR of
$\|\mathbf{x}_i\|_2^2/\E[\|\boldsymbol{\xi}_i\|_2^2] =\sigma_{tr}^{-2}$. To
begin, we assume both $L$ and $s$ are known a priori. Later, we relax this
assumption and extend our work in Sec.~\ref{ssec:linearcompleteprac} to the
case when these two parameters are unknown. Our goal is to learn $\cM_L$
using the training data $\mathbf{Y}$, which is equivalent to learning a
collection of $L$ subspaces that not only approximate the training data, but
are also ``close'' to each other on the Grassmann manifold (cf.~Definition~\ref{def:MC-UoS}).
Here, we pose this goal of learning an MC-UoS $\cM_L$ in
terms of the following optimization program:
\begin{align}    \label{eqn:linearproblem}
\{\cS_{\ell}\}_{\ell=1}^L & = \argmin_{\{\cS_{\ell}\} \subset \cG_{m,s}} \sum_{\substack{\ell,p=1\\\ell \neq p}}^{L} d_u^{2}(\cS_{\ell},\cS_p) \nonumber \\
&\qquad\qquad\qquad + \lambda \sum_{i=1}^{N} \| \mathbf{y}_i - P_{\cS_{l_i}} \mathbf{y}_i \|_{2}^{2},
\end{align}
where $l_i = \argmin_{\ell} \| \mathbf{y}_i - P_{\cS_{\ell}} \mathbf{y}_i
\|_{2}^{2}$ with $P_{\cS_{\ell}} \mathbf{y}_i$ denoting the (orthogonal)
projection of $\mathbf{y}_i$ onto the subspace $\cS_{\ell}$. Notice that the
first term in \eqref{eqn:linearproblem} forces the learned subspaces to be
close to each other, while the second term requires them to simultaneously
provide good approximations to the training data. The tuning parameter
$\lambda > 0$ in this setup provides a compromise between subspace closeness
and approximation error. While a discussion of finding an optimal $\lambda$
is beyond the scope of this paper, cross validation can be used to find
ranges of good values of tuning parameters in such problems
\cite{Kohavi.IJCAI1995} (also, see
Sec.~\ref{ssec:linearexperiment}). It is worth pointing out here
that \eqref{eqn:linearproblem} can be reformulated for the UoAS model through
a simple extension of the metric defined in \eqref{eqn:dmetric}. In addition,
note that \eqref{eqn:linearproblem} is mathematically similar to a related
problem studied in the clustering literature \cite{PelckmansDSD.PASCAL2005}.
In fact, it is straightforward to show that \eqref{eqn:linearproblem} reduces
to the clustering problem in \cite{PelckmansDSD.PASCAL2005} for $\cM_L$ being
a union of zero-dimensional affine subspaces.

\begin{figure}[t]
\centering
\subfigure[]{\includegraphics[width=1.6in]{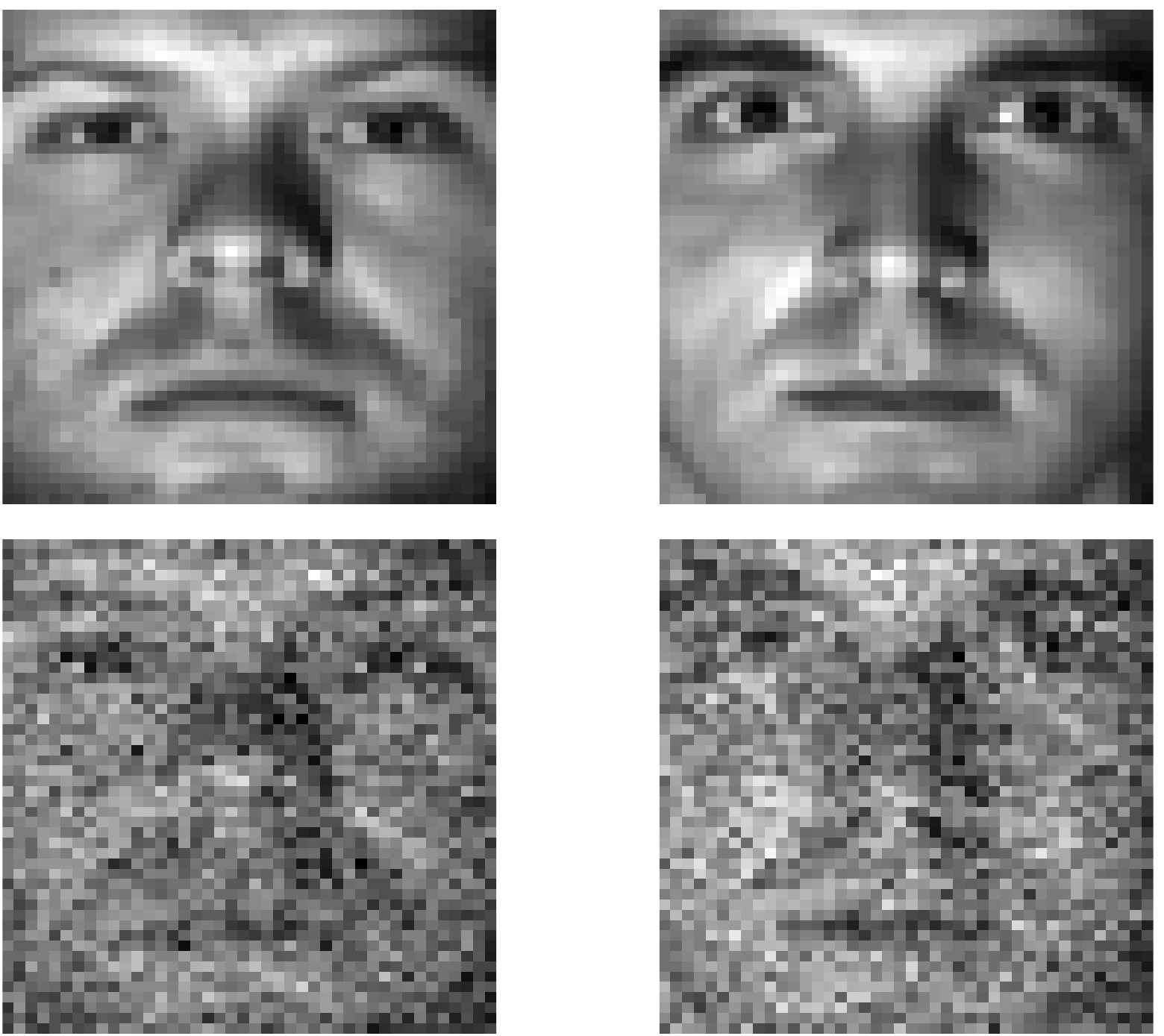} \label{fig:Example1}}
\quad
\subfigure[]{\includegraphics[width=1.6in]{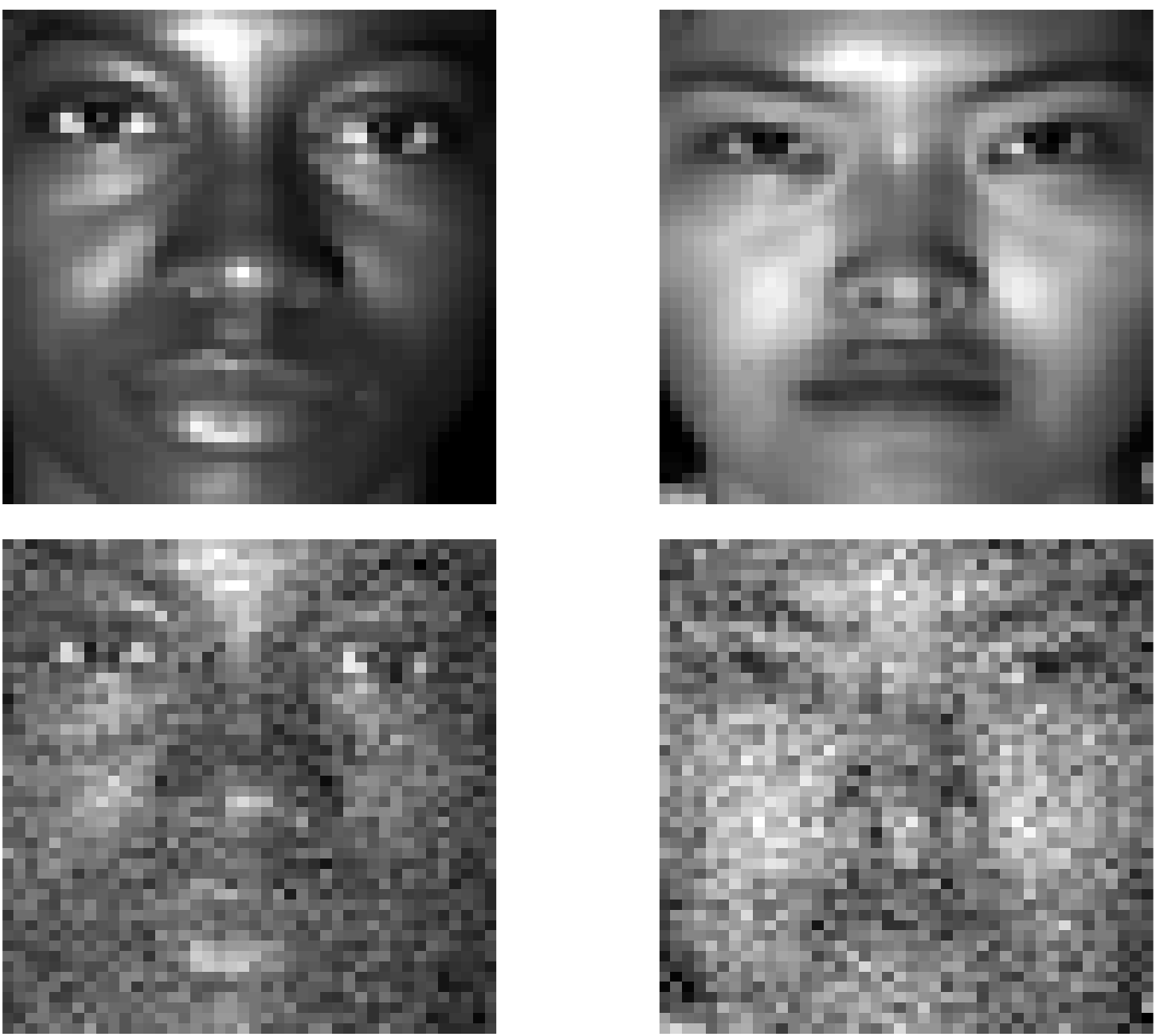} \label{fig:Example2}}
\caption{An example illustrating the limitations of existing methods for UoS learning
from noisy training data. The top row in this figure shows examples of
``clean'' facial images of four individuals in the Extended Yale B dataset
\cite{LeeHK.PAMI2005}, while the bottom row shows noisy versions of these
images, corresponding to $\sigma_{tr}^2=0.1$. The ``ground truth'' distance
between the subspaces of the individuals in (a) is $1.7953$, while it is
$2.3664$ between the subspaces of the individuals in (b). State-of-the-art
UoS learning methods have trouble reliably learning the underlying subspaces
whenever the subspaces are close to each other. Indeed, while the distance
between the two subspaces learned by the SSC algorithm
\cite{ElhamifarV.PAMI2013} from noisy images of the individuals in (b) is
$2.4103$, it is $2.4537$ for the case of ``similar-looking'' individuals in
(a).}
\label{fig:motivateexample}
\end{figure}

\begin{remark} The MC-UoS model and the learning problem
\eqref{eqn:linearproblem} can be further motivated as follows. Consider a set
of facial images of individuals under varying illumination conditions in the Extended Yale B dataset \cite{LeeHK.PAMI2005}, as in
Figs.~\ref{fig:Example1} and \ref{fig:Example2}. It is generally agreed that
all images of an individual in this case can be regarded as lying near a
$9$-dimensional subspace \cite{BasriJ.PAMI2003}, which can be computed in a
straightforward manner using singular value decomposition (SVD). The subspace
distance defined in \eqref{eqn:dmetric} can be used in this case to identify
similar-looking individuals. Given noisy training images of such ``similar''
individuals, traditional methods for UoS learning such as \emph{sparse
subspace clustering} (SSC) \cite{ElhamifarV.PAMI2013} that rely only on the
approximation error will be prone to errors. Fig.~\ref{fig:motivateexample} provides
a numerical validation of this claim, where it is shown that SSC has good
performance on noisy images of different-looking individuals
(cf.~Fig.~\ref{fig:Example2}), but its performance degrades in the case of
similar-looking individuals (cf.~Fig.~\ref{fig:Example1}). The MC-UoS
learning problem \eqref{eqn:linearproblem}, on the other hand, should be able
to handle both cases reliably because of the first term in
\eqref{eqn:linearproblem} that penalizes subspaces that do not cluster on the
Grassmann manifold. We refer the reader to Sec.~\ref{ssec:linearexperiment}
for detailed experiments that numerically validate this claim.
\end{remark}

In this paper, we study two variants of the MC-UoS learning problem described
by \eqref{eqn:linearproblem}. In the first variant, all $m$ dimensions of
each training sample in $\mathbf{Y}$ are observed and the geometry learning
problem is exactly given by \eqref{eqn:linearproblem}. In the second variant,
it is assumed that some of the $m$ dimensions of each training sample in
$\mathbf{Y}$ are unobserved (i.e., missing), which then requires a
recharacterization of \eqref{eqn:linearproblem} for the learning problem to
be well posed. We defer that recharacterization to Sec.~\ref{ssec:linearmiss}
of the paper. In order to quantify the performance of our learning
algorithms, we will resort to generation of noisy test data as
follows. Given noiseless (synthetic or real) data sample $\mathbf{x}$ with
$\|\mathbf{x}\|_2^2=1$, noisy test sample $\mathbf{z}$ is given by
$\mathbf{z} = \mathbf{x} + \boldsymbol{\xi}$ with the additive noise
$\boldsymbol{\xi} \sim \mathcal{N}(\boldsymbol{0},
(\sigma_{te}^2/m)\mathbf{I}_m)$. We will then report the metric of
\emph{average approximation error of noisy test data} using the learned
subspaces for synthetic and real data. Finally, in the case of synthetic
data drawn from an MC-UoS, we will also measure the performance of our
algorithms in terms of \emph{average normalized subspace distances} between
the learned and the true subspaces. We defer a formal description of
both these metrics to Sec.~\ref{sssec:syntheticlinear}, which describes in
detail the setup of our experiments.

\subsection{Geometry Learning for Highly Nonlinear Data}
\label{ssec:probkernel}

Our second geometry learning problem corresponds to the case of
high-dimensional data drawn from a mixture of nonlinear manifolds in the
ambient space $\R^m$. The basic premise of our model in this case is
that when data drawn from a mixture of nonlinear manifolds are mapped through
a nonlinear map $\phi: \R^{m} \to \cF$ to a higher-dimensional feature space
$\cF \subset \R^{\widetilde{m}}$ with $\widetilde{m} \ggg m$, then the
$\phi$-mapped ``images'' of these data can be modeled as lying near an MC-UoS
$\cM_L$ in the feature space. In order to learn this model, we once again
assume access to a collection of $N$ training samples, $\mathbf{Y}= [ \mathbf{y}_1, \dots, \mathbf{y}_N ] \in \R^{m \times N}$, with the
fundamental difference here being that the \emph{mapped training data}
$\phi(\mathbf{Y}) = [\phi(\mathbf{y}_1), \dots, \phi(\mathbf{y}_N)]$ are now
assumed to be drawn from an MC-UoS $\cM_L = \bigcup_{\ell=1}^{L} \cS_{\ell}
\subset \cG_{\widetilde{m},s} \subset \cF$. Here, we also make the simplified
assumption that $\rank(\phi(\mathbf{Y}))=N$, which is justified as long as
$\widetilde{m} \ggg N$ and no two training samples are identical. Our goal in
this setting is to learn the (feature space) MC-UoS $\cM_L$ using the
training data $\mathbf{Y}$, which in theory can still be achieved by solving
the following variant of \eqref{eqn:linearproblem}:
\begin{align}   \label{eqn:nonlinearproblem}
\{\cS_{\ell}\}_{\ell=1}^L & = \argmin_{\{\cS_{\ell}\} \subset \cG_{\widetilde{m},s}} \sum_{\substack{\ell,p=1\\\ell \neq p}}^{L} d_u^{2}(\cS_{\ell},\cS_p) \nonumber \\
&\qquad\qquad\qquad + \lambda \sum_{i=1}^{N} \| \phi(\mathbf{y}_i) - P_{\cS_{l_i}} \phi(\mathbf{y}_i) \|_{2}^{2},
\end{align}
where $l_i = \argmin_{\ell} \| \phi(\mathbf{y}_i) - P_{\cS_{\ell}} \phi(\mathbf{y}_i)
\|_{2}^{2}$ with $P_{\cS_{\ell}} \phi(\mathbf{y}_i)$ denoting the (orthogonal)
projection of $\phi(\mathbf{y}_i)$ onto the $s$-dimensional subspace $\cS_{\ell}$ in
$\R^{\widetilde{m}}$.

In practice, however, solving \eqref{eqn:nonlinearproblem} directly is likely
to be computationally intractable due to the extremely high dimensionality of
the feature space. Instead, we are interested in solving the problem of
MC-UoS learning in the feature space using the ``kernel trick''
\cite{ScholkopfSM.NC1998}, which involves transforming
\eqref{eqn:nonlinearproblem} into a learning problem that only requires
evaluations of inner products in $\cF$. Such a transformation can then be
followed with the use of a Mercer kernel $\kappa$, which is a positive
semidefinite function $\kappa: \R^{m} \times \R^{m} \to \R$ that satisfies
$\kappa(\mathbf{y},\mathbf{y'}) = \langle \phi(\mathbf{y}), \phi(\mathbf{y'})
\rangle$ for all $\mathbf{y},\mathbf{y'} \in \R^{m}$, to develop algorithms
that can learn an MC-UoS in the feature space without explicit mapping of the
training data to the feature space. We term the learning of an MC-UoS in the
feature space using the kernel trick as metric-constrained kernel
union-of-subspaces (MC-KUoS) learning. Similar to the case of MC-UoS
learning, we consider two scenarios in this paper for MC-KUoS learning. The
first one of these scenarios corresponds to the standard setup in which all
$m$ dimensions of each training sample in $\mathbf{Y}$ are observed, while
the second scenario corresponds to the case of ``missing data'' in which some
dimensions of each training sample in $\mathbf{Y}$ remain unobserved.
Finally, we will evaluate the proposed MC-KUoS learning algorithms
using ($i$) the metric of average approximation error of noisy test data, and
($ii$) their clustering performance on training data having either complete or
missing entries. We conclude here by pointing out that MC-KUoS learning
invariably also leads us to the problem of finding the ``pre-images'' of data
in the feature space induced by our chosen kernel (e.g., Gaussian or
polynomial kernel) \cite{MikaSSMSR.NIPS1998,KwokT.TNN2004}, which will also
be addressed in this paper.

\begin{remark}
It is worth noting here that \eqref{eqn:nonlinearproblem} requires knowledge
of the nonlinear map $\phi$. However, since we rely on the kernel trick for
our MC-KUoS learning framework, we only need access to an appropriate kernel
$\kappa$. It is assumed in this paper that such a kernel is readily available
to us. While learning the ``best'' kernel from training data is an
interesting extension of our work, it is beyond the scope of this paper.
\end{remark}

\section{MC-UoS Learning for Mildly Nonlinear Data}
\label{sec:linearsolver}

In this section, we describe our approach to the problem of MC-UoS learning
for mildly nonlinear data. We begin our discussion for the case when all $m$
dimensions of each training sample are available to us.

\subsection{MC-UoS Learning Using Complete Data}
\label{ssec:linearcomplete}

In order to reduce the effects of noisy training data, we begin with
a pre-processing step that centers the data matrix
$\mathbf{Y}$.\footnote{While such pre-processing is common in many
geometry learning algorithms, it is not central to our framework.} This
involves defining the mean of the samples in $\mathbf{Y}$ as
$\bar{\mathbf{y}} = \frac{1}{N} \sum_{i=1}^{N} \mathbf{y}_i$ and then
subtracting this mean from $\mathbf{Y}$ to obtain the centered data
$\widetilde{\mathbf{Y}} = [ \widetilde{\mathbf{y}}_1, \dots,
\widetilde{\mathbf{y}}_N ]$, where $\widetilde{\mathbf{y}}_i = \mathbf{y}_i -
\bar{\mathbf{y}}$, $i=1,\dots,N$. Next, we focus on simplification of the
optimization problem \eqref{eqn:linearproblem}. To this end, we first define
an $L \times N$ indicator matrix $\mathbf{W}$ that identifies memberships of
the $\widetilde{\mathbf{y}}_i$'s in different subspaces, where
$w_{\ell,i}=1$, $\ell=1,\dots,L$, $i=1,\dots,N$, if and only if
$\widetilde{\mathbf{y}}_i$ is ``closest'' to subspace $\cS_{\ell}$;
otherwise, $w_{\ell,i}=0$. Mathematically,
\begin{align}    \label{eqn:defW}
\mathbf{W} = \big[w_{\ell,i} \in \{0,1\}: \forall i=1,\dots,N, \sum_{\ell=1}^{L} w_{\ell,i}=1\big].
\end{align}
Further, notice that $\| \mathbf{y}_i - P_{\cS_{\ell}} \mathbf{y}_i \|_{2}^{2}$ in
\eqref{eqn:linearproblem} can be rewritten as
\begin{align}
\| \mathbf{y}_i - P_{\cS_{\ell}} \mathbf{y}_i \|_{2}^{2} = \| \widetilde{\mathbf{y}}_i - P_{\cS_{\ell}} \widetilde{\mathbf{y}}_i \|_{2}^{2} = \| \widetilde{\mathbf{y}}_i \|_2^2 -  \| \mathbf{D}_{\ell}^{T} \widetilde{\mathbf{y}}_i \|_2^2,
\end{align}
where $\mathbf{D}_{\ell} \in \R^{m \times s}$ denotes an (arbitrary) orthonormal basis
of $\cS_{\ell}$. Therefore, defining $\mathbf{D} = [\mathbf{D}_1, \dots, \mathbf{D}_L]$ to be a
collection of orthonormal bases of $\cS_{\ell}$'s, we can rewrite
\eqref{eqn:linearproblem} as $(\mathbf{D},\mathbf{W}) = \argmin_{\mathbf{D},\mathbf{W}} F_1(\mathbf{D},\mathbf{W})$ with
the objective function $F_1(\mathbf{D},\mathbf{W})$ given by\footnote{Note that the
minimization here is being carried out under the assumption of $\mathbf{D}_{\ell}$'s
being orthonormal and $\mathbf{W}$ being described by \eqref{eqn:defW}.}
\begin{align}  \label{eqn:linearcomprob}
& F_1(\mathbf{D},\mathbf{W}) = \sum_{\substack{\ell,p=1\\\ell \neq p}}^{L} \| \mathbf{D}_{\ell} - P_{\cS_p} \mathbf{D}_{\ell} \|_{F}^{2} \nonumber \\
&\qquad\qquad + \lambda \sum_{i=1}^{N} \sum_{\ell=1}^{L} w_{\ell,i} (\|\widetilde{\mathbf{y}}_i\|_2^2 -  \| \mathbf{D}_{\ell}^{T} \widetilde{\mathbf{y}}_i \|_2^2).
\end{align}

\algsetup{indent=0.5em}
\begin{algorithm}[t]
\caption{Metric-Constrained Union-of-Subspaces Learning (MiCUSaL)}
\label{algo:MiCUSaL}%
\textbf{Input:} Training data $\mathbf{Y} \in \R^{m \times N}$, number of subspaces $L$, dimension of subspaces $s$, and parameter $\lambda$.\\
\textbf{Initialize:} Random orthonormal bases $\{ \mathbf{D}_{\ell} \in \R^{m \times s} \}_{\ell=1}^{L}$.

\begin{algorithmic}[1]
\STATE $\bar{\mathbf{y}} \gets \frac{1}{N} \sum_{i=1}^{N} \mathbf{y}_i, \ \widetilde{\mathbf{y}}_i \gets \mathbf{y}_i - \bar{\mathbf{y}}, \ i=1,\dots,N$.
\WHILE {stopping rule}
\FOR{$i=1$ to $N$ (\emph{Subspace Assignment})}%
\STATE $l_i \gets \argmax_{\ell} \| \mathbf{D}_{\ell}^{T} \widetilde{\mathbf{y}}_i \|_2$.
\STATE $w_{l_i,i} \gets 1$ and $\forall \ell \neq l_i, \ w_{\ell,i} \gets 0$.
\ENDFOR
\FOR{$\ell=1$ to $L$ (\emph{Subspace Update})}%
\STATE $\mathbf{c}_{\ell} \gets \{i \in \{1,\dots,N\}: w_{\ell,i} = 1\}$.
\STATE $\widetilde{\mathbf{Y}}_{\ell} \gets [\widetilde{\mathbf{y}}_i: i \in \mathbf{c}_{\ell}]$.
\STATE $\mathbf{A}_{\ell} \gets \sum_{p \neq \ell} {\mathbf{D}_{p} \mathbf{D}_{p}^{T}}
+ \frac{\lambda}{2} \widetilde{\mathbf{Y}}_{\ell} \widetilde{\mathbf{Y}}_{\ell}^{T}$.
\STATE Eigen decomposition of $\mathbf{A}_{\ell} = \mathbf{U}_{\ell} \mathbf{\Sigma}_{\ell} \mathbf{U}_{\ell}^{T}$.
\STATE $\mathbf{D}_{\ell} \gets$ Columns of $\mathbf{U}_{\ell}$ corresponding to the \\
$s$-largest diagonal elements in $\mathbf{\Sigma}_{\ell}$.
\ENDFOR
\ENDWHILE
\end{algorithmic}
\textbf{Output:} Orthonormal bases $\{ \mathbf{D}_{\ell} \in \R^{m \times s} \}_{\ell=1}^{L}$.
\end{algorithm}

Minimizing \eqref{eqn:linearcomprob} simultaneously over $\mathbf{D}$ and $\mathbf{W}$ is
challenging and is likely to be computationally infeasible. Instead, we adopt
an alternate minimization approach \cite{Bertsekas.1999,BezdekH.NPSC2003}, which involves iteratively
solving \eqref{eqn:linearcomprob} by alternating between the following two
steps: ($i$) minimizing $F_1(\mathbf{D},\mathbf{W})$ over $\mathbf{W}$ for a fixed $\mathbf{D}$, which we term as
the \emph{subspace assignment} step; and ($ii$) minimizing $F_1(\mathbf{D},\mathbf{W})$ over
$\mathbf{D}$ for a fixed $\mathbf{W}$, which we term as the \emph{subspace update} stage. To
begin this alternate minimization, we start with an initial $\mathbf{D}$ in which each
block $\mathbf{D}_{\ell} \in \R^{m \times s}$ is a random orthonormal basis. Next, we
fix this $\mathbf{D}$ and carry out subspace assignment, which now amounts to solving for each $i=1,\dots,N$,
\begin{align}   \label{eqn:linearcomsubassign}
l_i = \argmin_{\ell=1,\dots,L} \| \widetilde{\mathbf{y}}_i - P_{\cS_{\ell}} \widetilde{\mathbf{y}}_i \|_2^2 = \argmax_{\ell=1,\dots,L} {\| \mathbf{D}_{\ell}^{T} \widetilde{\mathbf{y}}_i \|_2^2},
\end{align}
and then setting $w_{l_i,i}=1$ and $w_{\ell,i}=0~\forall \ell \neq l_i$. In order
to move to the subspace update step, we fix the matrix $\mathbf{W}$ and focus on
optimizing $F_1(\mathbf{D},\mathbf{W})$ over $\mathbf{D}$. However, this step requires more attention
since minimizing over the entire $\mathbf{D}$ at once will also lead to a large-scale
optimization problem. We address this problem by once again resorting to
block-coordinate descent (BCD) \cite{Bertsekas.1999} and updating only one
$\mathbf{D}_{\ell}$ at a time while keeping the other $\mathbf{D}_{p}$'s ($p \neq \ell$) fixed
in \eqref{eqn:linearcomprob}. In this regard, suppose we are in the process
of updating $\mathbf{D}_{\ell}$ for a fixed $\ell$ during the subspace update step.
Define $\mathbf{c}_{\ell} = \{i \in \{1,\dots,N\}: w_{\ell,i} = 1\}$ to be the set
containing the indices of all $\widetilde{\mathbf{y}}_i$'s that are assigned to
$\cS_{\ell}$ (equivalently, $\mathbf{D}_{\ell}$) and let $\widetilde{\mathbf{Y}}_{\ell} =
[ \widetilde{\mathbf{y}}_i: i \in \mathbf{c}_{\ell} ]$ be the corresponding $m \times
|\mathbf{c}_{\ell}|$ matrix. Then it can be shown after some manipulations of
\eqref{eqn:linearcomprob} that updating $\mathbf{D}_{\ell}$ is equivalent to solving the
following problem:
\begin{align}      \label{eqn:linearcomblockupdate}
\mathbf{D}_{\ell} & = \argmin_{ \mathbf{Q} \in \cV_{m,s} }  \sum_{p \neq \ell} \| \mathbf{Q} - P_{\cS_p} \mathbf{Q} \|_{F}^{2} + \frac{\lambda}{2}  ( \| \widetilde{\mathbf{Y}}_{\ell} \|_{F}^{2} - \| \mathbf{Q}^{T} \widetilde{\mathbf{Y}}_{\ell} \|_{F}^{2} ) \nonumber \\
& = \argmax_{ \mathbf{Q} \in \cV_{m,s} } \tr \Big( \mathbf{Q}^{T} ( \sum_{p \neq \ell} {\mathbf{D}_{p} \mathbf{D}_{p}^{T}} + \frac{\lambda}{2} \widetilde{\mathbf{Y}}_{\ell} \widetilde{\mathbf{Y}}_{\ell}^{T} ) \mathbf{Q} \Big),
\end{align}
where $\cV_{m,s}$ denotes the Stiefel manifold, defined as the collection of
all $m \times s$ orthonormal matrices. Note that
\eqref{eqn:linearcomblockupdate} has an intuitive interpretation. When
$\lambda = 0$, \eqref{eqn:linearcomblockupdate} reduces to the problem of
finding a subspace that is closest to the remaining $L-1$ subspaces in our
collection. When $\lambda = \infty$, \eqref{eqn:linearcomblockupdate} reduces
to the PCA problem, in which case the learning problem
\eqref{eqn:linearproblem} reduces to the subspace clustering problem studied
in \cite{HoYLLK.CVPR2003}. By selecting an appropriate $\lambda \in
(0,\infty)$ in \eqref{eqn:linearcomblockupdate}, we straddle the two extremes
of \emph{subspace closeness} and \emph{data approximation}. In order to solve
\eqref{eqn:linearcomblockupdate}, we define an $m \times m$ symmetric matrix
$\mathbf{A}_{\ell} = \sum_{p \neq \ell} {\mathbf{D}_{p} \mathbf{D}_{p}^{T}}
+ \frac{\lambda}{2} \widetilde{\mathbf{Y}}_{\ell} \widetilde{\mathbf{Y}}_{\ell}^{T}$. It then follows from
\cite{KokiopoulouCS.NLAA2010} that \eqref{eqn:linearcomblockupdate} has a closed-form
solution that involves eigen decomposition of $\mathbf{A}_{\ell}$.
Specifically, $\mathbf{D}_{\ell} = \argmax \tr(\mathbf{D}_{\ell}^{T}\mathbf{A}_{\ell}\mathbf{D}_{\ell})$ is given by the
first $s$ eigenvectors of $\mathbf{A}_{\ell}$ associated with its $s$-largest
eigenvalues.

This completes our description of the subspace update step. We can now
combine the subspace assignment and subspace update steps to fully describe
our algorithm for MC-UoS learning. This algorithm, which we term
\emph{metric-constrained union-of-subspaces learning} (MiCUSaL), is given by
Algorithm~\ref{algo:MiCUSaL}. In terms of the complexity of this
algorithm in each iteration, notice that the subspace assignment step
requires $\mathcal{O}(mLsN)$ operations. In addition, the total number of operations
needed to compute the $\mathbf{A}_{\ell}$'s in each iteration is
$\mathcal{O}(m^2(Ls+N))$. Finally, each iteration also requires $L$ eigen
decompositions of $m \times m$ matrices, each one of which has
$\mathcal{O}(m^3)$ complexity. Therefore, the computational complexity of MiCUSaL in each iteration is
given by $\mathcal{O}(m^3L+m^2N+m^2Ls+mLsN)$. We conclude this discussion by pointing
out that we cannot guarantee convergence of MiCUSaL to a global optimal
solution. However, since the objective function $F_1$ in
\eqref{eqn:linearcomprob} is bounded below by zero and MiCUSaL ensures that
$F_1$ does not increase after each iteration, it follows that MiCUSaL
iterates do indeed converge (possibly to one of the local optimal solutions).
This local convergence, of course, will be a function of the
initialization of MiCUSaL. In this paper, we advocate the use of random
subspaces for initialization, while we study the impact of such random
initialization in Sec.~\ref{ssec:linearexperiment}.

\subsection{Practical Considerations}
\label{ssec:linearcompleteprac}

The MiCUSaL algorithm described in Sec.~\ref{ssec:linearcomplete} requires
knowledge of the number of subspaces $L$ and the dimension of subspaces $s$.
In practice, however, one cannot assume knowledge of these parameters a
priori. Instead, one must estimate both the number and the dimension of
subspaces from the training data themselves. In this section, we describe a
generalization of the MiCUSaL algorithm that achieves this objective. Our
algorithm, which we term \emph{adaptive MC-UoS learning} (aMiCUSaL), requires
only knowledge of loose upper bounds on $L$ and $s$,
which we denote by $L_{max}$ and $s_{max}$, respectively.

The aMiCUSaL algorithm initializes with a collection of random orthonormal bases $\mathbf{D}
= [\mathbf{D}_1, \dots, \mathbf{D}_{L_{max}}]$, where each basis $\mathbf{D}_{\ell}$ is a point on the
Stiefel manifold $\cV_{m,s_{max}}$. Similar to the case of MiCUSaL, it then
carries out the subspace assignment and subspace update steps in an iterative
fashion. Unlike MiCUSaL, however, we also greedily remove redundant subspaces
from our collection of subspaces $\{ \cS_{\ell} \}_{\ell=1}^{L_{max}}$ after
each subspace assignment step. This involves removal of $\mathbf{D}_{\ell}$ from $\mathbf{D}$
if no signals in our training data get assigned to the subspace $\cS_{\ell}$.
This step of \emph{greedy subspace pruning} ensures that only ``active''
subspaces survive before the subspace update step.

Once the aMiCUSaL algorithm finishes iterating between subspace assignment,
subspace pruning, and subspace update, we move onto the step of \emph{greedy
subspace merging}, which involves merging of pairs of subspaces that are
so close to each other that even a single subspace of the same
dimension can be used to well approximate the data represented by the two
subspaces individually.\footnote{Note that the step of subspace
merging is needed due to our lack of knowledge of the true number of
subspaces in the underlying model. In particular, the assumption here is that
the merging threshold $\epsilon_{min}$ in Algorithm~\ref{algo:aMiCUSaL} satisfies
$\epsilon_{min} \ll \frac{\epsilon}{\sqrt{s}}$.} In this step, we greedily merge pairs of
closest subspaces as long as their normalized subspace distance is below a
predefined threshold $\epsilon_{min} \in [0,1)$. Mathematically, the
subspace merging step involves first finding the pair of subspaces
$(\cS_{\ell^*}, \cS_{p^*})$ that satisfies
\begin{align}
(\ell^*,p^*) = \argmin_{\ell \neq p} d_u(\cS_{\ell},\cS_p) \ \text{s.t.} \ \frac{d_u(\cS_{\ell^*},\cS_{p^*})}{\sqrt{s_{max}}} \leq \epsilon_{min}.
\end{align}
We then merge $\cS_{\ell^*}$ and $\cS_{p^*}$ by setting $\mathbf{c}_{\ell^*} = \mathbf{c}_{\ell^*} \cup \mathbf{c}_{p^*}$ and $\widetilde{\mathbf{Y}}_{\ell^*} = [ \widetilde{\mathbf{y}}_i: i \in \mathbf{c}_{\ell^*} ]$, where $\mathbf{c}_{\ell^*}, \mathbf{c}_{p^*}$ are as defined in Algorithm~\ref{algo:MiCUSaL}. By defining an $m \times m$ symmetric matrix $\mathbf{A}_{\ell^*} = \sum_{\ell \neq \ell^*, p^*} {\mathbf{D}_{\ell}\mathbf{D}_{\ell}^{T}} + \frac{\lambda}{2} \widetilde{\mathbf{Y}}_{\ell^*} \widetilde{\mathbf{Y}}_{\ell^*}^{T}$, $\mathbf{D}_{\ell^*}$ is then set equal to the first $s_{max}$ eigenvectors of $\mathbf{A}_{\ell^*}$ associated with its $s_{max}$-largest eigenvalues. Finally, we remove $\mathbf{D}_{p^*}$ from $\mathbf{D}$. This process of finding the closest pair of subspaces and merging them is repeated until the normalized subspace distance between every pair of subspaces becomes greater than $\epsilon_{min}$. We assume without loss of generality that $\widehat{L}$ subspaces are left after this greedy subspace merging, where each $\cS_{\ell}$ ($\ell = 1,\dots,\widehat{L}$) is a subspace in $\R^m$ of dimension $s_{max}$.

\algsetup{indent=0.5em}
\begin{algorithm}[t]
\caption{Adaptive MC-UoS Learning (aMiCUSaL)}
\label{algo:aMiCUSaL}%
\textbf{Input:} Training data $\mathbf{Y} \in \R^{m \times N}$, loose upper bounds $L_{max}$ and $s_{max}$, and parameters $\lambda$, $k_1$, $k_2$, $\epsilon_{min}$.\\
\textbf{Initialize:} Random orthonormal bases $\{ \mathbf{D}_{\ell} \in \R^{m \times s_{max}} \}_{\ell=1}^{L_{max}}$, and set $\widehat{L} \gets L_{max}$.

\begin{algorithmic}[1]
\STATE $\bar{\mathbf{y}} \gets \frac{1}{N} \sum_{i=1}^{N} \mathbf{y}_i, \ \widetilde{\mathbf{y}}_i \gets \mathbf{y}_i - \bar{\mathbf{y}}, \ i=1,\dots,N$.
\WHILE {stopping rule}
\STATE Fix $\mathbf{D}$ and update $\mathbf{W}$ using \eqref{eqn:linearcomsubassign}. Also, set $\boldsymbol{\mathcal{T}} \gets \emptyset$ and $L_1 \gets 0$.
\FOR{$\ell=1$ to $\widehat{L}$ (\emph{Subspace Pruning})}
\STATE $\mathbf{c}_{\ell} \gets \{i \in \{1,\dots,N\}: w_{\ell,i} = 1\}$.
\STATE If $|\mathbf{c}_{\ell}| \neq 0$ then $\mathbf{c}_{L_1+1} \gets \mathbf{c}_{\ell}, \ \widetilde{\mathbf{Y}}_{L_1+1} \gets [ \widetilde{\mathbf{y}}_i: i \in \mathbf{c}_{\ell} ], \ L_1 \gets L_1+1$ and $\boldsymbol{\mathcal{T}} \gets \boldsymbol{\mathcal{T}} \cup \{ \ell \}$.
\ENDFOR
\STATE $\mathbf{D} \gets [ \mathbf{D}_{\boldsymbol{\mathcal{T}}_{(1)}}, \dots, \mathbf{D}_{\boldsymbol{\mathcal{T}}_{(L_1)}} ]$ and $\widehat{L} \gets L_1$.
\STATE Update each $\mathbf{D}_{\ell} \ (\ell = 1, \dots, \widehat{L})$ in $\mathbf{D}$ using \eqref{eqn:linearcomblockupdate}.
\ENDWHILE
\STATE $(\ell^*,p^*) = \argmin_{\ell \neq p, \ell, p = 1, \dots, \widehat{L}} d_u(\cS_{\ell},\cS_p)$.
\WHILE {$\frac{d_u(\cS_{\ell^*},\cS_{p^*})}{\sqrt{s_{max}}} \leq \epsilon_{min}$ (\emph{Subspace Merging})}
\STATE Merge $\cS_{\ell^*}$ and $\cS_{p^*}$, and update $\mathbf{Y}_{\ell^*}$ and $\mathbf{D}_{\ell^*}$.
\STATE $\mathbf{D} \gets [ \mathbf{D}_1, \dots, \mathbf{D}_{\ell^*}, \dots, \mathbf{D}_{p^{*}-1}, \mathbf{D}_{p^{*}+1}, \dots, \mathbf{D}_{\widehat{L}} ]$ and $\widehat{L} \gets \widehat{L}-1$.
\STATE $(\ell^*,p^*) = \argmin_{\ell \neq p, \ell, p = 1, \dots, \widehat{L}} d_u(\cS_{\ell},\cS_p)$.
\ENDWHILE
\STATE Fix $\mathbf{D}$, update $\mathbf{W} \in \R^{\widehat{L} \times N}$ using \eqref{eqn:linearcomsubassign}, and update $\{ \mathbf{c}_{\ell} \}_{\ell=1}^{\widehat{L}}$.
\FOR{$\ell=1$ to $\widehat{L}$}%
\STATE $\widetilde{\mathbf{Y}}_{\ell} \gets [\widetilde{\mathbf{y}}_i: i \in \mathbf{c}_{\ell}]$ and $\widehat{\mathbf{Y}}_{\ell} \gets \mathbf{D}_{\ell}\mathbf{D}_{\ell}^{T} \widetilde{\mathbf{Y}}_{\ell}$.
\STATE Calculate $\widehat{s}_{\ell}$ using \eqref{eqn:sellcalculate} and \eqref{eqn:sestimate}.
\ENDFOR
\STATE $s \gets \max \{\widehat{s}_1, \dots, \widehat{s}_{\widehat{L}} \}$.
\STATE $\widehat{\mathbf{D}}_{\ell} \gets $ First $s$ columns of $\mathbf{D}_{\ell}, \ \ell = 1, \dots, \widehat{L}$.
\STATE Initialize Algorithm~\ref{algo:MiCUSaL} with $\{ \widehat{\mathbf{D}}_{\ell} \}_{\ell=1}^{\widehat{L}}$ and update $\{ \widehat{\mathbf{D}}_{\ell} \}_{\ell=1}^{\widehat{L}}$ using Algorithm~\ref{algo:MiCUSaL}.
\end{algorithmic}
\textbf{Output:} Orthonormal bases $\{ \widehat{\mathbf{D}}_{\ell} \in \R^{m \times s} \}_{\ell=1}^{\widehat{L}}$.
\end{algorithm}

After subspace merging, we move onto the step of estimation of the dimension,
$s$, of the subspaces. To this end, we first estimate the dimension of each
subspace $\cS_{\ell}$, denoted by $s_{\ell}$, and then $s$ is selected as the
maximum of these $s_{\ell}$'s. There have been many efforts in the literature
to estimate the dimension of a subspace; see, e.g.,
\cite{LevinaB.NIPS2004,Bioucas-DiasN.TGRS2008,KritchmanN.CILS2008,PerryW.JSTSP2010}
for an incomplete list. In this paper, we focus on the method given in
\cite{LevinaB.NIPS2004}, which formulates the maximum likelihood estimator
(MLE) of $s_{\ell}$. This is because: ($i$) the noise level is unknown in our
problem, and ($ii$) the MLE in \cite{LevinaB.NIPS2004} has a simple form.
However, the MLE of \cite{LevinaB.NIPS2004} is sensitive to noise. We
therefore first apply a ``smoothing'' process before using that estimator.
This involves first updating $\mathbf{W}$ (i.e.,
$\mathbf{c}_{\ell}$'s) using the updated $\mathbf{D}$ and ``denoising'' our
data by projecting $\widetilde{\mathbf{Y}}_{\ell}$ onto $\cS_{\ell}$, given
by $\widehat{\mathbf{Y}}_{\ell} = \mathbf{D}_{\ell}\mathbf{D}_{\ell}^T
\widetilde{\mathbf{Y}}_{\ell}$, and then using $\widehat{\mathbf{Y}}_{\ell}$
to estimate $s_{\ell}$. For a given column $\widehat{\mathbf{y}}$ in
$\widehat{\mathbf{Y}}_{\ell}$ and a fixed number of nearest neighbors $k_0$, the
unbiased MLE of $s_{\ell}$ with respect to $\widehat{\mathbf{y}}$ is given by
\cite{LevinaB.NIPS2004}
\begin{align}   \label{eqn:sellcalculate}
\widehat{s}_{\ell}^{k_0}(\widehat{\mathbf{y}}) = \Big[ \frac{1}{k_0-2} \sum_{a=1}^{k_0-1} \log \frac{\Gamma_{k_0}(\widehat{\mathbf{y}})}{\Gamma_{a}(\widehat{\mathbf{y}})} \Big]^{-1},
\end{align}
where $\Gamma_{a}(\widehat{\mathbf{y}})$ is the $\ell_2$ distance between $\widehat{\mathbf{y}}$ and its $a$-th nearest neighbor in $\widehat{\mathbf{Y}}_{\ell}$. An estimate of $s_{\ell}$ can now be written as the average of all estimates with respect to every signal in $\widehat{\mathbf{Y}}_{\ell}$, i.e., $ \widehat{s}_{\ell}^{k_0} = \frac{1}{|\mathbf{c}_{\ell}|} \sum_{i \in \mathbf{c}_{\ell}} \widehat{s}_{\ell}^{k_0}(\widehat{\mathbf{y}}_{i}) $. In fact, as suggested in \cite{LevinaB.NIPS2004}, we estimate $s_{\ell}$ by averaging over a range of $k_0 =  k_1, \dots, k_2$, i.e.,
\begin{align}   \label{eqn:sestimate}
\widehat{s}_{\ell} = \frac{1}{k_2-k_1+1} \sum_{k_0=k_1}^{k_2} \widehat{s}_{\ell}^{k_0}.
\end{align}
Once we get an estimate $s = \max_{\ell} \widehat{s}_{\ell}$, we trim each orthonormal basis by keeping only the first $s$ columns of each (ordered) orthonormal basis $\mathbf{D}_{\ell}$ in our collection, which is denoted by $\widehat{\mathbf{D}}_{\ell}$.\footnote{Recall that the columns of $\mathbf{D}_{\ell}$ correspond to the eigenvectors of $\mathbf{A}_{\ell}$. Here, we are assuming that the order of these eigenvectors within $\mathbf{D}_{\ell}$ corresponds to the nonincreasing order of the eigenvalues of $\mathbf{A}_{\ell}$. Therefore, $\widehat{\mathbf{D}}_{\ell}$ comprises the eigenvectors of $\mathbf{A}_{\ell}$ associated with its $s$-largest eigenvalues.} Given the bases $\{ \widehat{\mathbf{D}}_{\ell} \in \R^{m \times s} \}_{\ell=1}^{\widehat{L}}$, we finally run MiCUSaL again that is initialized using these $\widehat{\mathbf{D}}_{\ell}$'s until it converges. Combining all the steps mentioned above, we can now formally describe \emph{adaptive MC-UoS learning} (aMiCUSaL) in Algorithm~\ref{algo:aMiCUSaL}.

\subsection{MC-UoS Learning Using Missing Data}
\label{ssec:linearmiss}

In this section, we study MC-UoS learning for the case of training data having missing entries. To be specific, for each $\mathbf{y}_i$ in $\mathbf{Y}$, we assume to only observe its entries at locations given by the set $\mathbf{\Omega}_i \subset \{1,\dots,m\}$ with $|\mathbf{\Omega}_i| > s$, which is denoted by $[\mathbf{y}_i]_{\mathbf{\Omega}_i} \in \R^{|\mathbf{\Omega}_i|}$. Since we do not have access to the complete data, it is impossible to compute the quantities $\| \mathbf{y}_i - P_{\cS_{\ell}} \mathbf{y}_i \|_{2}^{2}$ in \eqref{eqn:linearproblem} explicitly. But, it is shown in \cite{BalzanoRN.ISIT2010} that $ \| [\mathbf{y}_i]_{\mathbf{\Omega}_i} - P_{\cS_{\ell\mathbf{\Omega}_i}} [\mathbf{y}_i]_{\mathbf{\Omega}_i} \|_2^2 $ for uniformly random $\mathbf{\Omega}_i$ is very close to $\frac{|\mathbf{\Omega}_i|}{m} \| \mathbf{y}_i - P_{\cS_{\ell}} \mathbf{y}_i\|_{2}^{2}$ with very high probability as long as $|\mathbf{\Omega}_i|$ is slightly greater than $s$. Here, $P_{\cS_{\ell\mathbf{\Omega}_i}}$ is defined as $P_{\cS_{\ell\mathbf{\Omega}_i}} = [\mathbf{D}_{\ell}]_{\mathbf{\Omega}_i,:} {([\mathbf{D}_{\ell}]_{\mathbf{\Omega}_i,:})}^{\dagger}$ with ${([\mathbf{D}_{\ell}]_{\mathbf{\Omega}_i,:})}^{\dagger} = \big( [\mathbf{D}_{\ell}]_{\mathbf{\Omega}_i,:}^{T} [\mathbf{D}_{\ell}]_{\mathbf{\Omega}_i,:} \big)^{-1} [\mathbf{D}_{\ell}]_{\mathbf{\Omega}_i,:}^{T}$. Motivated by this, we replace $\| \mathbf{y}_i - P_{\cS_{\ell}} \mathbf{y}_i\|_{2}^{2}$ by $ \frac{m}{|\mathbf{\Omega}_i|} \| [\mathbf{y}_i]_{\mathbf{\Omega}_i} - P_{\cS_{\ell\mathbf{\Omega}_i}} [\mathbf{y}_i]_{\mathbf{\Omega}_i} \|_2^2 $ in \eqref{eqn:linearproblem} and reformulate the MC-UoS learning problem as $(\mathbf{D},\mathbf{W}) = \argmin_{\mathbf{D},\mathbf{W}} F_2(\mathbf{D},\mathbf{W})$ with the objective function $F_2(\mathbf{D},\mathbf{W})$ given by
\begin{align}   \label{eqn:linearmissprob}
& F_2(\mathbf{D},\mathbf{W}) = \sum_{\substack{\ell,p=1\\\ell \neq p}}^{L} \| \mathbf{D}_{\ell} - P_{\cS_p} \mathbf{D}_{\ell} \|_{F}^{2} \nonumber \\
&\qquad + \lambda \sum_{i=1}^{N} \sum_{\ell=1}^{L} w_{\ell,i} \frac{m}{|\mathbf{\Omega}_i|} \big{\Vert} [\mathbf{y}_i]_{\mathbf{\Omega}_i} - P_{\cS_{\ell\mathbf{\Omega}_i}} [\mathbf{y}_i]_{\mathbf{\Omega}_i} \big{\Vert}_2^2.
\end{align}

As in Sec.~\ref{ssec:linearcomplete}, we propose to solve this problem by making use of alternating minimization that comprises subspace assignment and subspace update steps. To this end, we again initialize $\mathbf{D}$ such that each block $\mathbf{D}_{\ell} \in \R^{m \times s}$ is a random orthonormal basis. Next, when $\mathbf{D}$ is fixed, subspace assignment corresponds to solving for each $i=1,\dots,N$,
\begin{align}
l_i = \argmin_{\ell=1,\dots,L} \| [\mathbf{y}_i]_{\mathbf{\Omega}_i} - P_{\cS_{\ell\mathbf{\Omega}_i}} [\mathbf{y}_i]_{\mathbf{\Omega}_i} \|_2^2,
\end{align}
and then setting $w_{l_i,i}=1$ and $w_{\ell,i}=0~\forall \ell \neq l_i$. When $\mathbf{W}$ is fixed, we carry out subspace update using BCD again, in which case $\min_{\mathbf{D}} F_2(\mathbf{D},\mathbf{W})$ for a fixed $\mathbf{W}$ can be decoupled into $L$ distinct problems of the form $\mathbf{D}_{\ell} = \argmin_{\mathbf{D}_{\ell} \in \cV_{m,s}} f_2(\mathbf{D}_{\ell})$, $\ell = 1, \dots, L$, with
\begin{align*}
f_2(\mathbf{D}_{\ell}) & =  - \tr(\mathbf{D}_{\ell}^{T}\mathbf{A}_{\ell}\mathbf{D}_{\ell})  \\
&\qquad\qquad + \frac{\lambda}{2} \sum_{i \in \mathbf{c}_{\ell}} \frac{m}{|\mathbf{\Omega}_i|} \big{\Vert} [\mathbf{y}_i]_{\mathbf{\Omega}_i} - P_{\cS_{\ell\mathbf{\Omega}_i}} [\mathbf{y}_i]_{\mathbf{\Omega}_i} \big{\Vert}_2^2.
\end{align*}
Here, $\mathbf{c}_{\ell}$ is as defined in Sec.~\ref{ssec:linearcomplete} and $\mathbf{A}_{\ell} = \sum_{p \neq \ell} \mathbf{D}_{p}\mathbf{D}_{p}^{T}$. It is also easy to verify that $f_2(\mathbf{D}_{\ell})$ is invariant to the choice of the orthonormal basis of $\cS_{\ell}$; hence, we can treat $\min_{\mathbf{D}_{\ell} \in \cV_{m,s}} f_2(\mathbf{D}_{\ell})$ as an optimization problem on the Grassmann manifold \cite{EdelmanAS.SIAM1998}. Note that we can rewrite $f_2(\mathbf{D}_{\ell})$ as $f_2(\mathbf{D}_{\ell}) = \sum_{q=0}^{|\mathbf{c}_{\ell}|} f_2^{(q)}(\mathbf{D}_{\ell}) $, where $f_2^{(0)}(\mathbf{D}_{\ell}) = -\tr(\mathbf{D}_{\ell}^{T}\mathbf{A}_{\ell}\mathbf{D}_{\ell})$ and $f_2^{(q)}(\mathbf{D}_{\ell}) =  \frac{\lambda m}{2|\mathbf{\Omega}_{{\mathbf{c}_{\ell}}_{(q)}}|} \| [\mathbf{y}_{{\mathbf{c}_{\ell}}_{(q)}}]_{\mathbf{\Omega}_{{\mathbf{c}_{\ell}}_{(q)}}} - P_{\cS_{\ell \mathbf{\Omega}_{{\mathbf{c}_{\ell}}_{(q)}}}} [\mathbf{y}_{{\mathbf{c}_{\ell}}_{(q)}}]_{\mathbf{\Omega}_{{\mathbf{c}_{\ell}}_{(q)}}} \|_2^2$ for $q = 1, \dots, |\mathbf{c}_{\ell}|$. In here, ${\mathbf{c}_{\ell}}_{(q)}$ denotes the $q$-th element in $\mathbf{c}_{\ell}$. In order to minimize $f_2(\mathbf{D}_{\ell})$, we employ incremental gradient descent procedure on Grassmann manifold \cite{NedicB.SIAM2001}, which performs the update with respect to a single component of $f_2(\mathbf{D}_{\ell})$ in each step. To be specific, we first compute the gradient of a single cost function $f_2^{(q)}(\mathbf{D}_{\ell})$ in $f_2(\mathbf{D}_{\ell})$, and move along a short geodesic curve in the gradient direction. For instance, the gradient of $f_2^{(0)}(\mathbf{D}_{\ell})$ is
\begin{align*}
\nabla f_2^{(0)} = (\mathbf{I}_{m} - \mathbf{D}_{\ell}\mathbf{D}_{\ell}^{T}) \frac{df_2^{(0)}}{d\mathbf{D}_{\ell}} = -2(\mathbf{I}_{m} - \mathbf{D}_{\ell}\mathbf{D}_{\ell}^{T})\mathbf{A}_{\ell}\mathbf{D}_{\ell}.
\end{align*}
Then the geodesic equation emanating from $\mathbf{D}_{\ell}$ in the direction $- \nabla f_2^{(0)}$ with a step length $\eta$ is given by \cite{EdelmanAS.SIAM1998}
\begin{align}
\mathbf{D}_{\ell}(\eta) = \mathbf{D}_{\ell}\mathbf{V}_{\ell} \cos(\mathbf{\Sigma}_{\ell} \eta) \mathbf{V}_{\ell}^{T} + \mathbf{U}_{\ell} \sin(\mathbf{\Sigma}_{\ell} \eta) \mathbf{V}_{\ell}^{T},
\end{align}
where $\mathbf{U}_{\ell} \mathbf{\Sigma}_{\ell} \mathbf{V}_{\ell}^{T}$ is the
SVD decomposition of $- \nabla f_2^{(0)}$. The update of $\mathbf{D}_{\ell}$
with respect to $f_2^{(q)}(\mathbf{D}_{\ell})$ $(q = 1, \dots,
|\mathbf{c}_{\ell}|)$ can be performed as in the GROUSE algorithm
\cite{BalzanoNR.Allerton2010} \emph{but} with a step size $\frac{\lambda m}{2
|\mathbf{\Omega}_{{\mathbf{c}_{\ell}}_{(q)}}|}\eta$. In order for $f_2$ to
converge, we also reduce the step size after each iteration
\cite{BalzanoNR.Allerton2010}. We complete this discussion by
presenting our learning algorithm for missing data in Algorithm~\ref{algo:rMiCUSaL}, termed \emph{robust MC-UoS learning} (rMiCUSaL).
By defining $T = \max_i |\mathbf{\Omega}_i|$, the subspace assignment step of rMiCUSaL requires $\mathcal{O}(TLs^2N)$ flops in each iteration \cite{BalzanoNR.Allerton2010}. Computing the $\mathbf{A}_{\ell}$'s in each iteration requires
$\mathcal{O}(m^2Ls)$ operations. Next, the cost of updating each $\mathbf{D}_{\ell}$ with respect to
$f_2^{(0)}(\mathbf{D}_{\ell})$ is $\mathcal{O}(m^3)$, while it is
$\mathcal{O}(ms+Ts^2)$ with respect to $f_2^{(q)}(\mathbf{D}_{\ell})$ for $q \neq 0$
\cite{BalzanoNR.Allerton2010}. It therefore follows that the computational
complexity of rMiCUSaL in each iteration is $\mathcal{O}(m^3L+m^2Ls+msN+TLs^2N)$.
We refer the reader to Sec.~\ref{ssec:linearexperiment} for exact running
time of rMiCUSaL on training data.

\algsetup{indent=0.5em}
\begin{algorithm}[t]
\caption{Robust MC-UoS learning (rMiCUSaL)}
\label{algo:rMiCUSaL}%
\textbf{Input:} Training data $\{ [\mathbf{y}_i]_{\mathbf{\Omega}_i} \}_{i=1}^{N}$, number of subspaces $L$, dimension of subspaces $s$, and parameters $\lambda$ and $\eta$.\\
\textbf{Initialize:} Random orthonormal bases $\{ \mathbf{D}_{\ell} \in \R^{m \times s} \}_{\ell=1}^{L}$.

\begin{algorithmic}[1]
\WHILE {stopping rule}
\FOR{$i=1$ to $N$ (\emph{Subspace Assignment})}%
\STATE $l_i \gets \argmin_{\ell} \| [\mathbf{y}_i]_{\mathbf{\Omega}_i} - P_{\cS_{\ell\mathbf{\Omega}_i}} [\mathbf{y}_i]_{\mathbf{\Omega}_i} \|_2^2$.
\STATE $w_{l_i,i} \gets 1$ and $\forall \ell \neq l_i, \ w_{\ell,i} \gets 0$.
\ENDFOR
\FOR{$\ell=1$ to $L$ (\emph{Subspace Update})}%
\STATE $\mathbf{c}_{\ell} \gets \{i \in \{1,\dots,N\}: w_{\ell,i} = 1\}, \ t \gets 0$.
\WHILE {stopping rule}
\STATE $t \gets t+1, \ \eta_t \gets \frac{\eta}{t}$.
\STATE $\mathbf{A}_{\ell} \gets \sum_{p \neq \ell} \mathbf{D}_{p}\mathbf{D}_{p}^{T}, \ \mathbf{\Delta}_{\ell} \gets 2(\mathbf{I}_{m} - \mathbf{D}_{\ell}\mathbf{D}_{\ell}^{T})\mathbf{A}_{\ell}\mathbf{D}_{\ell}$.
\STATE $\mathbf{D}_{\ell} \gets \mathbf{D}_{\ell}\mathbf{V}_{\ell} \cos(\mathbf{\Sigma}_{\ell} \eta_t) \mathbf{V}_{\ell}^{T} + \mathbf{U}_{\ell} \sin(\mathbf{\Sigma}_{\ell} \eta_t) \mathbf{V}_{\ell}^{T}$, where $\mathbf{U}_{\ell} \mathbf{\Sigma}_{\ell} \mathbf{V}_{\ell}^{T}$ is the compact SVD of $\mathbf{\Delta}_{\ell}$.
\FOR{$q=1$ to $|\mathbf{c}_{\ell}|$}
\STATE $\boldsymbol{\theta} \gets { ( [\mathbf{D}_{\ell}]_{\mathbf{\Omega}_{{\mathbf{c}_{\ell}}_{(q)}},:} ) }^{\dagger} [\mathbf{y}_{{\mathbf{c}_{\ell}}_{(q)}}]_{\mathbf{\Omega}_{{\mathbf{c}_{\ell}}_{(q)}}}, \ \boldsymbol{\omega} \gets \mathbf{D}_{\ell} \boldsymbol{\theta}$.
\STATE $\mathbf{r} \gets \boldsymbol{0}_m, \ [\mathbf{r}]_{\mathbf{\Omega}_{{\mathbf{c}_{\ell}}_{(q)}}} \gets [\mathbf{y}_{{\mathbf{c}_{\ell}}_{(q)}}]_{\mathbf{\Omega}_{{\mathbf{c}_{\ell}}_{(q)}}} - [\boldsymbol{\omega}]_{\mathbf{\Omega}_{{\mathbf{c}_{\ell}}_{(q)}}}$.
\STATE $\mathbf{D}_{\ell} \gets \mathbf{D}_{\ell} + \Big( (\cos(\mu \frac{\lambda m}{|\mathbf{\Omega}_{{\mathbf{c}_{\ell}}_{(q)}}|} \eta_t) -1) \frac{\boldsymbol{\omega}}{\| \boldsymbol{\omega} \|_2} + \sin(\mu \frac{\lambda m}{|\mathbf{\Omega}_{{\mathbf{c}_{\ell}}_{(q)}}|} \eta_t) \frac{\mathbf{r}}{\| \mathbf{r} \|_2} \Big) \frac{\boldsymbol{\theta}^{T}}{ \| \boldsymbol{\theta} \|_2}$, where $\mu = \| \mathbf{r} \|_2 \| \boldsymbol{\omega} \|_2$.
\ENDFOR
\ENDWHILE
\ENDFOR
\ENDWHILE
\end{algorithmic}
\textbf{Output:} Orthonormal bases $\{ \mathbf{D}_{\ell} \in \R^{m \times s} \}_{\ell=1}^{L}$.
\end{algorithm}

\section{MC-KUoS Learning for Highly Nonlinear Data}
\label{sec:kernelsolver}

In this section, we present algorithms to solve the problem of MC-KUoS learning from $\mathbf{Y} \in \R^{m \times N}$ for highly nonlinear data. We first generalize the MiCUSaL algorithm using the kernel trick \cite{ScholkopfSM.NC1998} to learn an MC-KUoS from complete data. To deal with the case of ``missing data,'' we propose ``kernel function value estimators'' to solve \eqref{eqn:nonlinearproblem}. Finally, we discuss the problem of finding the ``pre-images'' of data in the feature space based on the MC-KUoS model in Sec.~\ref{ssec:preimage}.

\subsection{MC-KUoS Learning Using Complete Data}
\label{ssec:kernelcomplete}

To begin our discussion, we define the kernel matrix on the training data $\mathbf{Y}$ to be $\mathbf{G} = \phi(\mathbf{Y})^{T}\phi(\mathbf{Y}) \in \R^{N \times N}$, with its individual entries $g_{i,j} = \kappa(\mathbf{y}_i,\mathbf{y}_j)$ for a pre-specified kernel $\kappa: \R^{m} \times \R^{m} \to \R$. Under the assumption that $\rank(\phi(\mathbf{Y}))=N$, the matrix $\mathbf{G}$ is positive definite. Similar to Algorithm~\ref{algo:MiCUSaL}, we begin with centering the $\phi$-mapped data in the feature space $\cF$ as a pre-processing stage.\footnote{This step is only for the purpose of derivation of our algorithm. In particular, explicit centering of data in the feature space is never required in the following.} We denote the mean of the $\phi$-mapped ``images'' of $\mathbf{Y}$ by $\overline{\boldsymbol{\phi}} = \frac{1}{N} \sum_{i=1}^{N} \phi(\mathbf{y}_i)$ and write the $N$ centered ``mapped training data'' as $\widetilde{\phi}(\mathbf{Y}) = [ \widetilde{\phi}(\mathbf{y}_1), \dots, \widetilde{\phi}(\mathbf{y}_N) ]$, where $\widetilde{\phi}(\mathbf{y}_i) = \phi(\mathbf{y}_i) - \overline{\boldsymbol{\phi}}$, $i = 1,\dots,N$. The centered kernel matrix $\widetilde{\mathbf{G}} = \widetilde{\phi}(\mathbf{Y})^{T} \widetilde{\phi}(\mathbf{Y})$ can be calculated from $\mathbf{G}$ by \cite{ScholkopfSM.NC1998}
\begin{align}
\widetilde{\mathbf{G}} = \mathbf{G} - \mathbf{H}_{N}\mathbf{G} - \mathbf{G}\mathbf{H}_{N}  + \mathbf{H}_{N}\mathbf{G}\mathbf{H}_{N},
\end{align}
where $\mathbf{H}_{N}$ is an $N \times N$ matrix with all elements $\frac{1}{N}$. Then for any $\mathbf{y}, \mathbf{y'} \in \R^m$, we have \cite{KwokT.TNN2004}
\begin{align*}
\widetilde{\kappa}(\mathbf{y},\mathbf{y'}) & = \widetilde{\phi}(\mathbf{y})^{T} \widetilde{\phi}(\mathbf{y'}) \\
& = \kappa(\mathbf{y},\mathbf{y'}) - \frac{1}{N}\boldsymbol{1}_{N}^{T} \boldsymbol{k}_{\mathbf{y}} - \frac{1}{N}\boldsymbol{1}_{N}^{T} \boldsymbol{k}_{\mathbf{y'}} + \frac{1}{N^2}\boldsymbol{1}_{N}^{T} \mathbf{G} \boldsymbol{1}_{N},
\end{align*}
where $\boldsymbol{1}_{N} = [1,1,\dots,1]^{T}$ is an $N$-dimensional vector and $\boldsymbol{k}_{\mathbf{y}} = [\kappa(\mathbf{y},\mathbf{y}_1), \dots, \kappa(\mathbf{y},\mathbf{y}_N)]^{T} \in \R^N$. To write the expression in \eqref{eqn:nonlinearproblem} in terms of inner products, we again use $\mathbf{W}$ to denote the membership indicator matrix as in \eqref{eqn:defW}, where $w_{\ell,i}=1$, $\ell=1,\dots,L$, $i=1,\dots,N$, if and only if $\widetilde{\phi}(\mathbf{y}_i)$ is assigned to subspace $\cS_{\ell}$. Let $\mathbf{D} = [ \mathbf{D}_1, \dots, \mathbf{D}_L ]$, where $\mathbf{D}_{\ell}$ is an (arbitrary) orthonormal basis of $\cS_{\ell}$. Then for any $i = 1,\dots,N$, we have the following
\begin{align}
\| \phi(\mathbf{y}_i) - P_{\cS_{\ell}} \phi(\mathbf{y}_i) \|_{2}^{2} & =  \| \widetilde{\phi}(\mathbf{y}_i) - P_{\cS_{\ell}} \widetilde{\phi}(\mathbf{y}_i) \|_{2}^{2}   \nonumber \\
& = \| \widetilde{\phi}(\mathbf{y}_i) \|_2^2 -  \| \mathbf{D}_{\ell}^{T} \widetilde{\phi}(\mathbf{y}_i) \|_2^2.
\end{align}
Therefore, \eqref{eqn:nonlinearproblem} can be written as $(\mathbf{D},\mathbf{W}) = \argmin_{\mathbf{D},\mathbf{W}} F_3(\mathbf{D},\mathbf{W})$ with the objective function $F_3(\mathbf{D},\mathbf{W})$ given by
\begin{align}  \label{eqn:nonlinearcomprob}
& F_3(\mathbf{D},\mathbf{W}) = \sum_{\substack{\ell,p=1\\\ell \neq p}}^{L} \| \mathbf{D}_{\ell} - P_{\cS_p} \mathbf{D}_{\ell} \|_{F}^{2} \nonumber \\
&\qquad\qquad + \lambda \sum_{i=1}^{N} \sum_{\ell=1}^{L} w_{\ell,i} (\| \widetilde{\phi}(\mathbf{y}_i) \|_2^2 -  \| \mathbf{D}_{\ell}^{T} \widetilde{\phi}(\mathbf{y}_i) \|_2^2).
\end{align}

Before discussing our algorithm to minimize \eqref{eqn:nonlinearcomprob} using the kernel trick, we further simplify the terms in \eqref{eqn:nonlinearcomprob}. We again define $\mathbf{c}_{\ell} = \{i \in \{1,\dots,N\}: w_{\ell,i} = 1\}$ to be the set containing the indices of all $\widetilde{\phi}(\mathbf{y}_i)$'s that are assigned to $\cS_{\ell}$, and let $\mathbf{Y}_{\ell} = [\mathbf{y}_i: i \in \mathbf{c}_{\ell}]$ be the corresponding $m \times |\mathbf{c}_{\ell}|$ matrix. Then the centered data that are assigned to subspace $\cS_{\ell}$ can be denoted by $\widetilde{\phi}(\mathbf{Y}_{\ell}) = [\widetilde{\phi}(\mathbf{y}_i): i \in \mathbf{c}_{\ell} ]$. Since $\cS_{\ell}$ is spanned by the columns of $\widetilde{\phi}(\mathbf{Y}_{\ell})$, we can write $\mathbf{D}_{\ell} = \widetilde{\phi}(\mathbf{Y}_{\ell}) \mathbf{E}_{\ell}$, where $\mathbf{E}_{\ell} \in \R^{N_{\ell} \times s}$ is some basis representation matrix with $N_{\ell} = |\mathbf{c}_{\ell}|$ such that $\mathbf{D}_{\ell}$ is an orthonormal matrix. Also, it is easy to verify that $\mathbf{E}_{\ell}$ has to satisfy $\mathbf{E}_{\ell}^{T} [\widetilde{\mathbf{G}}]_{\mathbf{c}_{\ell},\mathbf{c}_{\ell}}  \mathbf{E}_{\ell} = \mathbf{I}_{s}$, where $[\widetilde{\mathbf{G}}]_{\mathbf{c}_{\ell},\mathbf{c}_{\ell}} = \widetilde{\phi}(\mathbf{Y}_{\ell})^{T} \widetilde{\phi}(\mathbf{Y}_{\ell})$ denotes the centered kernel matrix for subspace $\cS_{\ell}$. Now instead of using $\mathbf{D}_{\ell}$ explicitly for computations, it suffices to use $\mathbf{c}_{\ell}$ and $\mathbf{E}_{\ell}$ for MC-KUoS learning and all the computations involving $\mathbf{D}_{\ell}$ can be carried out using $\mathbf{c}_{\ell}$, $\mathbf{E}_{\ell}$ and the kernel trick. Specifically, notice that for any $i = 1, \dots, N$, we can write
\begin{align*}
\| \widetilde{\phi}(\mathbf{y}_i) \|_2^2 -  \| \mathbf{D}_{\ell}^{T} \widetilde{\phi}(\mathbf{y}_i) \|_2^2 = \widetilde{\kappa}(\mathbf{y}_i,\mathbf{y}_i) - \|  \mathbf{E}_{\ell}^{T} \widetilde{\phi}(\mathbf{Y}_{\ell})^{T} \widetilde{\phi}(\mathbf{y}_i)\|_2^2,
\end{align*}
where $\widetilde{\kappa}(\mathbf{y}_i,\mathbf{y}_i) = \kappa(\mathbf{y}_i,\mathbf{y}_i) - \frac{2}{N} \boldsymbol{1}_{N}^{T} \boldsymbol{k}_{\mathbf{y}_i}  + \frac{1}{N^2} \boldsymbol{1}_{N}^{T} \mathbf{G} \boldsymbol{1}_{N}$. To compute $\widetilde{\phi}(\mathbf{Y}_{\ell})^{T} \widetilde{\phi}(\mathbf{y}_i)$, we define $\phi(\mathbf{Y}_{\ell}) = [\phi(\mathbf{y}_i): i \in \mathbf{c}_{\ell}]$ and let $\boldsymbol{\psi}_{\ell}(\mathbf{y}_i) = [\kappa(\mathbf{y}_{{\mathbf{c}_{\ell}}_{(1)}},\mathbf{y}_i), \dots, \kappa(\mathbf{y}_{{\mathbf{c}_{\ell}}_{(N_{\ell})}},\mathbf{y}_i)]^{T} \in \R^{N_{\ell}}$ be a vector with elements given by the inner products between $\phi(\mathbf{y}_i)$ and columns of $\phi(\mathbf{Y}_{\ell})$. Then $\boldsymbol{\widetilde{\psi}}_{\ell}(\mathbf{y}_i) = \widetilde{\phi}(\mathbf{Y}_{\ell})^{T} \widetilde{\phi}(\mathbf{y}_i) = \boldsymbol{\psi}_{\ell}(\mathbf{y}_i) - \frac{1}{N} \boldsymbol{1}_{N_{\ell}} \boldsymbol{1}_{N}^{T} \boldsymbol{k}_{\mathbf{y}_i} - \frac{1}{N} [\mathbf{G}]_{\mathbf{c}_{\ell},:} \boldsymbol{1}_{N} + \frac{1}{N^2} \boldsymbol{1}_{N_{\ell}} \boldsymbol{1}_{N}^{T} \mathbf{G} \boldsymbol{1}_{N}$. Therefore, we can write $\| \phi(\mathbf{y}_i) - P_{\cS_{\ell}} \phi(\mathbf{y}_i) \|_2^2 = \widetilde{\kappa}(\mathbf{y}_i,\mathbf{y}_i) - \|  \mathbf{E}_{\ell}^{T} \boldsymbol{\widetilde{\psi}}_{\ell}(\mathbf{y}_i)\|_2^2$. Also, we have
\begin{align}
& d_u^2(\cS_{\ell},\cS_p) \nonumber \\
& = \| \mathbf{D}_{\ell} - P_{\cS_p} \mathbf{D}_{\ell} \|_{F}^{2} = s - \tr(\mathbf{D}_{\ell}^{T} \mathbf{D}_{p} \mathbf{D}_{p}^{T} \mathbf{D}_{\ell})   \nonumber \\
& = s - \tr \Big[ (\widetilde{\phi}(\mathbf{Y}_{\ell}) \mathbf{E}_{\ell})^{T} \widetilde{\phi}(\mathbf{Y}_p) \mathbf{E}_p (\widetilde{\phi}(\mathbf{Y}_p) \mathbf{E}_p)^{T} \widetilde{\phi}(\mathbf{Y}_{\ell}) \mathbf{E}_{\ell} \Big] \nonumber \\
& = s - \tr \Big[ \mathbf{E}_{\ell}^{T} [\widetilde{\mathbf{G}}]_{\mathbf{c}_{\ell}, \mathbf{c}_p} \mathbf{E}_p \mathbf{E}_p^{T} [\widetilde{\mathbf{G}}]_{\mathbf{c}_p, \mathbf{c}_{\ell}} \mathbf{E}_{\ell} \Big],
\end{align}
where $[\widetilde{\mathbf{G}}]_{\mathbf{c}_{\ell},\mathbf{c}_p} = \widetilde{\phi}(\mathbf{Y}_{\ell})^{T} \widetilde{\phi}(\mathbf{Y}_p)$ denotes the centered inter-subspace kernel matrix between $\cS_{\ell}$ and $\cS_p$.

\algsetup{indent=0.5em}
\begin{algorithm}[t]
\caption{Initialization for $\cS_{\ell}$'s in $\cF$ (GKIOP)}
\label{algo:GKIOP}
\textbf{Input:} Centered kernel matrix $\widetilde{\mathbf{G}} \in \R^{N \times N}$, number of subspaces $L$, and dimension of subspaces $s$. \\
\textbf{Initialize:} $\boldsymbol{\mathcal{I}}_N \gets \{1,\dots,N\}$ and $\{ \mathbf{c}_\ell \gets \emptyset \}_{\ell=1}^L$.

\begin{algorithmic}[1]
\FOR{$\ell=1$ to $L$}
\STATE Choose an arbitrary element in $\boldsymbol{\mathcal{I}}_N$, include that element in $\mathbf{c}_{\ell}$, and set $\boldsymbol{\mathcal{I}}_N \gets \boldsymbol{\mathcal{I}}_N \setminus \mathbf{c}_{\ell}$.
\FOR{$q=2$ to $s$}
\STATE $i^* \gets \argmax_{i \in \boldsymbol{\mathcal{I}}_N} \sum_{j \in \mathbf{c}_{\ell}}  \widetilde{g}_{i,j}$.
\STATE Set $\mathbf{c}_{\ell} \gets \mathbf{c}_{\ell} \cup \{i^*\}$ and $\boldsymbol{\mathcal{I}}_N \gets \boldsymbol{\mathcal{I}}_N \setminus \{i^*\}$.
\ENDFOR
\STATE Eigen decomposition of $[\widetilde{\mathbf{G}}]_{\mathbf{c}_{\ell},\mathbf{c}_{\ell}} = \mathbf{U}_{\ell} \mathbf{\Sigma}_{\ell} \mathbf{U}_{\ell}^{T}$.
\STATE $\mathbf{E}_{\ell} \gets \mathbf{U}_{\ell} \mathbf{\Sigma}_{\ell}^{-\frac{1}{2}}$.
\ENDFOR
\end{algorithmic}
\textbf{Output:} Initial $\{ \mathbf{c}_{\ell} \}_{\ell=1}^{L}$ and $\{ \mathbf{E}_{\ell} \in \R^{s \times s} \}_{\ell=1}^{L}$.
\end{algorithm}

Now we are ready to describe our algorithm in detail. Similar to MiCUSaL, we
minimize \eqref{eqn:nonlinearcomprob} by alternating between ($i$) minimizing
$F_3(\mathbf{D},\mathbf{W})$ over $\mathbf{W}$ for a fixed $\mathbf{D}$ (the
\emph{kernel subspace assignment} step) and ($ii$) minimizing
$F_3(\mathbf{D},\mathbf{W})$ over $\mathbf{D}$ for a fixed $\mathbf{W}$ (the
\emph{kernel subspace update} step). To begin this alternate optimization
strategy, we start by initializing the orthonormal basis of each subspace. As
discussed earlier, the orthonormal basis $\mathbf{D}_{\ell}$ of $\cS_{\ell}$
can be represented as $\mathbf{D}_{\ell} =
\widetilde{\phi}(\mathbf{Y}_{\ell}) \mathbf{E}_{\ell}$ and we can compute
$\mathbf{E}_{\ell}$ explicitly by using
$[\widetilde{\mathbf{G}}]_{\mathbf{c}_{\ell},\mathbf{c}_{\ell}}$. Therefore
the initialization of $\mathbf{D}_{\ell}$ is equivalent to initializing
$\mathbf{c}_{\ell}$. Note that any $s$ linear independent vectors define an
$s$-dimensional subspace. Therefore, to initialize $\mathbf{c}_{\ell}$, we
need to choose $s$ samples in the training set such that the $\phi$-mapped
``images'' of these training samples are linearly independent in the feature
space. Our selection of $s$ samples is based on
the intuition that the inner products between samples that lie in the same
subspace in the feature space will be typically large
\cite{HeckelB.arxiv2013}. Our initialization procedure then involves greedily
selecting a new sample $\mathbf{y}_{i^*}$ from the training data in each step
such that the sum of the inner products between
$\widetilde{\phi}(\mathbf{y}_{i^*})$ and the data points already in
$\widetilde{\phi}(\mathbf{Y}_{\ell})$ is the largest, and then setting
$\mathbf{Y}_{\ell} = \mathbf{Y}_{\ell} \cup \mathbf{y}_{i^*}$. We list our
initialization method in Algorithm~\ref{algo:GKIOP}, referred to as \emph{greedy kernel initial-orthogonalization procedure} (GKIOP). Based on the assumption that all the $\phi(\mathbf{y}_i)$'s are linearly independent, it is guaranteed that $\widetilde{\phi}(\mathbf{Y}_{\ell})$ can define an $s$-dimensional subspace $\cS_{\ell}$. Note that $\bigcap_{{\ell}=1}^{L} \mathbf{c}_{\ell} = \emptyset$ and we
compute $\mathbf{E}_{\ell}$ by $\mathbf{E}_{\ell} = \mathbf{U}_{\ell}
\mathbf{\Sigma}_{\ell}^{-\frac{1}{2}}$, where $[\widetilde{\mathbf{G}}]_{\mathbf{c}_{\ell},\mathbf{c}_{\ell}} = \mathbf{U}_{\ell} \mathbf{\Sigma}_{\ell} \mathbf{U}_{\ell}^{T}$. Since $\mathbf{D}_{\ell} =
\widetilde{\phi}(\mathbf{Y}_{\ell})\mathbf{E}_{\ell}$, it is easy to convince
oneself that $\mathbf{D}_{\ell}^{T} \mathbf{D}_{\ell} = \mathbf{I}_{s}$ in
this case.

We now move onto the kernel subspace assignment step. When $\mathbf{D}$ (equivalently, $\mathbf{c}_{\ell}$'s and $\mathbf{E}_{\ell}$'s) is fixed, kernel subspace assignment corresponds to first solving $\forall i=1,\dots,N$,
\begin{align}   \label{eqn:kernelsubassign}
l_i & = \argmin_{\ell=1,\dots,L} \| \widetilde{\phi}(\mathbf{y}_i) - P_{\cS_{\ell}} \widetilde{\phi}(\mathbf{y}_i) \|_{2}^{2}   \nonumber \\
& = \argmin_{\ell=1,\dots,L} \widetilde{\kappa}(\mathbf{y}_i,\mathbf{y}_i) - \|  \mathbf{E}_{\ell}^{T} \boldsymbol{\widetilde{\psi}}_{\ell}(\mathbf{y}_i)\|_2^2,
\end{align}
and then setting $w_{l_i,i}=1$ and $w_{\ell,i}=0~\forall \ell \neq l_i$. Next, for the kernel subspace update stage, since $\mathbf{W}$ is fixed, all the $\mathbf{c}_{\ell}$'s and $\mathbf{Y}_{\ell}$'s are fixed. By writing $\mathbf{D}_{\ell} = \widetilde{\phi}(\mathbf{Y}_{\ell})\mathbf{E}_{\ell}$, minimization of \eqref{eqn:nonlinearcomprob} for a fixed $\mathbf{W}$ can be written as a function of $\mathbf{E}_{\ell}$'s as follows:
\begin{align}   \label{eqn:kernelblockupdate}
& \min_{ \{ \mathbf{E}_{\ell} \} } f_3(\mathbf{E}_1,\dots,\mathbf{E}_L) = \sum_{\substack{\ell,p=1\\\ell \neq p}}^{L} \| \widetilde{\phi}(\mathbf{Y}_{\ell})\mathbf{E}_{\ell} - P_{\cS_p} (\widetilde{\phi}(\mathbf{Y}_{\ell})\mathbf{E}_{\ell}) \|_{F}^{2}  \nonumber \\
&\qquad\qquad\qquad + \lambda \sum_{\ell=1}^{L} \big( \| \widetilde{\phi}(\mathbf{Y}_{\ell}) \|_{F}^{2} -  \| \mathbf{E}_{\ell}^{T} \widetilde{\phi}(\mathbf{Y}_{\ell})^{T} \widetilde{\phi}(\mathbf{Y}_{\ell}) \|_{F}^{2} \big) \nonumber \\
&\quad \mbox{s.t.} \quad \mathbf{E}_{\ell}^{T} [\widetilde{\mathbf{G}}]_{\mathbf{c}_{\ell},\mathbf{c}_{\ell}} \mathbf{E}_{\ell} = \mathbf{I}_{s}, \ \ell = 1,2,\dots,L.
\end{align}

\algsetup{indent=0.5em}
\begin{algorithm}[t]
\caption{Metric-Constrained Kernel Union-of-Subspaces Learning (MC-KUSaL)}
\label{algo:MCKUSaL}%
\textbf{Input:} Training data $\mathbf{Y} \in \R^{m \times N}$, number and dimension of subspaces $L$ and $s$, kernel function $\kappa$ and parameter $\lambda$.\\

\begin{algorithmic}[1]
\STATE Compute kernel matrix $\mathbf{G}$: $g_{i,j} \gets \kappa(\mathbf{y}_i,\mathbf{y}_j)$.
\STATE $\widetilde{\mathbf{G}} \gets \mathbf{G} - \mathbf{H}_{N}\mathbf{G} - \mathbf{G}\mathbf{H}_{N}  + \mathbf{H}_{N}\mathbf{G}\mathbf{H}_{N}$.
\STATE Initialize $\{ \mathbf{c}_{\ell} \}_{\ell=1}^{L}$ and $\{ \mathbf{E}_{\ell} \}_{\ell=1}^{L} $ using GKIOP (Algorithm~\ref{algo:GKIOP}).
\WHILE {stopping rule}
\FOR{$i=1$ to $N$ (\emph{Kernel Subspace Assignment})}%
\STATE $l_i \gets \argmin_{\ell} \widetilde{\kappa}(\mathbf{y}_i,\mathbf{y}_i) - \|  \mathbf{E}_{\ell}^{T} \boldsymbol{\widetilde{\psi}}_{\ell}(\mathbf{y}_i) \|_2^2$.
\STATE $w_{l_i,i} \gets 1$ and $\forall \ell \neq l_i, \ w_{\ell,i} \gets 0$.
\ENDFOR
\FOR{$\ell=1$ to $L$ (\emph{Kernel Bases Initialization})}%
\STATE $\mathbf{c}_{\ell} \gets \{i \in \{1,\dots,N\}: w_{\ell,i} = 1\}$ and $N_{\ell} \gets |\mathbf{c}_{\ell}|$.
\STATE Eigen decomposition of $[\widetilde{\mathbf{G}}]_{\mathbf{c}_{\ell},\mathbf{c}_{\ell}} = \mathbf{U}_{\ell} \mathbf{\Sigma}_{\ell} \mathbf{U}_{\ell}^{T}$, with the diagonal elements of $\mathbf{\Sigma}_{\ell}$ in nonincreasing order.
\STATE $\mathbf{E}_{\ell} \gets  [\mathbf{U}_{\ell}]_{:, \boldsymbol{\mathcal{I}}_s} [\mathbf{\Sigma}_{\ell}]_{\boldsymbol{\mathcal{I}}_s,\boldsymbol{\mathcal{I}}_s}^{-\frac{1}{2}}$.
\ENDFOR
\WHILE {stopping rule}
\FOR{$\ell=1$ to $L$ (\emph{Kernel Subspace Update})}%
\STATE $\mathbf{A}_{\ell} \gets \sum_{p \neq \ell} { [\widetilde{\mathbf{G}}]_{\mathbf{c}_{\ell}, \mathbf{c}_p} \mathbf{E}_p \mathbf{E}_p^{T} [\widetilde{\mathbf{G}}]_{\mathbf{c}_p, \mathbf{c}_{\ell}} } + \frac{\lambda}{2} [\widetilde{\mathbf{G}}]_{\mathbf{c}_{\ell},\mathbf{c}_{\ell}}^{2}$.
\STATE $\mathbf{E}_{\ell} \gets$ Eigenvectors corresponding to the $s$-largest eigenvalues for the generalized problem $\mathbf{A}_{\ell} \mathbf{b} = \zeta [\widetilde{\mathbf{G}}]_{\mathbf{c}_{\ell},\mathbf{c}_{\ell}} \mathbf{b}$ such that $\mathbf{E}_{\ell}^{T} [\widetilde{\mathbf{G}}]_{\mathbf{c}_{\ell},\mathbf{c}_{\ell}} \mathbf{E}_{\ell} = \mathbf{I}_{s}$.
\ENDFOR
\ENDWHILE
\ENDWHILE
\end{algorithmic}
\textbf{Output:} $\{ N_{\ell} \in \mathbb{N} \}_{\ell=1}^{L}$, $\{ \mathbf{c}_{\ell} \}_{\ell=1}^{L}$ and $\{ \mathbf{E}_{\ell} \in \R^{ N_{\ell} \times s} \}_{\ell=1}^{L}$.
\end{algorithm}

Instead of updating all the $\mathbf{E}_{\ell}$'s simultaneously, which is again a difficult optimization problem, we use BCD to minimize $f_3$ and update $\mathbf{E}_{\ell}$'s sequentially. Unlike MC-UoS learning, however, we have to be careful here since the number of samples in $\mathbf{Y}$ that belong to $\mathbf{Y}_{\ell}$ (i.e., $N_{\ell}$) may change after each subspace assignment step. In particular, we first need to initialize all the $\mathbf{E}_{\ell}$'s such that $\mathbf{E}_{\ell} \in \R^{N_{\ell} \times s}$ and $\mathbf{E}_{\ell}^{T} [\widetilde{\mathbf{G}}]_{\mathbf{c}_{\ell},\mathbf{c}_{\ell}} \mathbf{E}_{\ell} = \mathbf{I}_{s}$. To do so, we again apply eigen decomposition of $[\widetilde{\mathbf{G}}]_{\mathbf{c}_{\ell},\mathbf{c}_{\ell}} = \mathbf{U}_{\ell} \mathbf{\Sigma}_{\ell} \mathbf{U}_{\ell}^{T}$ with the diagonal entries of $\mathbf{\Sigma}_{\ell}$ in nonincreasing order. Then we define $\boldsymbol{\mathcal{I}}_s = \{1,\dots,s\}$ and $\mathbf{E}_{\ell} =  [\mathbf{U}_{\ell}]_{:, \boldsymbol{\mathcal{I}}_s} [\mathbf{\Sigma}_{\ell}]_{\boldsymbol{\mathcal{I}}_s,\boldsymbol{\mathcal{I}}_s}^{-\frac{1}{2}}$. After this bases initialization step, we are ready to update $\mathbf{E}_{\ell}$'s sequentially and after some manipulations, each BCD subproblem of \eqref{eqn:kernelblockupdate} can be expressed as
\begin{align}   \label{eqn:kernelcomblockupdate}
\mathbf{E}_{\ell} & = \argmin_{ \mathbf{Q}: \mathbf{Q}^{T} [\widetilde{\mathbf{G}}]_{\mathbf{c}_{\ell},\mathbf{c}_{\ell}}  \mathbf{Q} = \mathbf{I}_{s} }  \sum_{p \neq \ell} \| \widetilde{\phi}(\mathbf{Y}_{\ell}) \mathbf{Q} - P_{\cS_p} (\widetilde{\phi}(\mathbf{Y}_{\ell}) \mathbf{Q}) \|_{F}^{2} \nonumber \\
&\qquad\qquad + \frac{\lambda}{2}  ( \| \widetilde{\phi}(\mathbf{Y}_{\ell}) \|_{F}^{2} - \| \mathbf{Q}^{T} \widetilde{\phi}(\mathbf{Y}_{\ell})^{T} \widetilde{\phi}(\mathbf{Y}_{\ell}) \|_{F}^{2} )  \nonumber  \\
& = \argmax_{ \mathbf{Q}: \mathbf{Q}^{T} [\widetilde{\mathbf{G}}]_{\mathbf{c}_{\ell},\mathbf{c}_{\ell}} \mathbf{Q} = \mathbf{I}_{s} }  \tr (\mathbf{Q}^{T} \mathbf{A}_{\ell} \mathbf{Q}),
\end{align}
where $\mathbf{A}_{\ell} = \sum_{p \neq \ell} { [\widetilde{\mathbf{G}}]_{\mathbf{c}_{\ell}, \mathbf{c}_p} \mathbf{E}_p \mathbf{E}_p^{T} [\widetilde{\mathbf{G}}]_{\mathbf{c}_p, \mathbf{c}_{\ell}} }
+ \frac{\lambda}{2} [\widetilde{\mathbf{G}}]_{\mathbf{c}_{\ell},\mathbf{c}_{\ell}}^{2}$ is a
symmetric matrix of dimension $N_{\ell} \times N_{\ell}$. Note that
\eqref{eqn:kernelcomblockupdate} has a similar intuitive interpretation as
\eqref{eqn:linearcomblockupdate}. When $\lambda = 0$,
\eqref{eqn:kernelcomblockupdate} reduces to the problem of finding a subspace
that is closest to the remaining $L-1$ subspaces in the feature space. When
$\lambda = \infty$, \eqref{eqn:kernelcomblockupdate} reduces to the kernel
PCA problem \cite{ScholkopfSM.Advance1999}. Since
$[\widetilde{\mathbf{G}}]_{\mathbf{c}_{\ell},\mathbf{c}_{\ell}}$ is a
positive definite matrix, it again follows from \cite{KokiopoulouCS.NLAA2010} that
the trace of $\mathbf{E}_{\ell}^{T}\mathbf{A}_{\ell}\mathbf{E}_{\ell}$ is
maximized when $\mathbf{E}_{\ell} = [\mathbf{b}_1, \dots, \mathbf{b}_s]$ is
the set of eigenvectors associated with the $s$-largest eigenvalues for the
generalized problem $\mathbf{A}_{\ell} \mathbf{b} = \zeta
[\widetilde{\mathbf{G}}]_{\mathbf{c}_{\ell},\mathbf{c}_{\ell}} \mathbf{b}$,
with $\mathbf{E}_{\ell}^{T}
[\widetilde{\mathbf{G}}]_{\mathbf{c}_{\ell},\mathbf{c}_{\ell}}
\mathbf{E}_{\ell} = \mathbf{I}_{s}$. The whole algorithm can be detailed in
Algorithm~\ref{algo:MCKUSaL}, which we refer to as
\emph{metric-constrained kernel union-of-subspaces learning} (MC-KUSaL).
An important thing to notice here is that the complexity of MC-KUSaL
does not scale with the dimensionality of the feature space $\cF$ owing to
our use of the kernel trick.

\subsection{MC-KUoS Learning Using Missing Data}
\label{ssec:kernelmiss}

In this section, we focus on MC-KUoS learning for the case of training data having missing entries in the input space. Our setup is similar to the one in Sec.~\ref{ssec:linearmiss}. That is, for $i = 1, \dots, N$, we observe $\mathbf{y}_i$ only at locations $\mathbf{\Omega}_i \subset \{1,\dots,m\}$. In the following, the resulting observed vector of $\mathbf{y}_i$ is denoted by $[\mathbf{y}_i]_{\mathbf{\Omega}_i} \in \R^{|\mathbf{\Omega}_i|}$. Also, we assume that the observed indices of each signal, $\mathbf{\Omega}_i$, are drawn uniformly at random with replacement from $\{1,\dots,m\}$. Note that the results derived in here can also be translated to the case of sampling $\mathbf{\Omega}_i$ without replacement (we refer the reader to \cite[Lemma 1]{ErikssonBN.AIStats2012} as an example). Given the missing data aspect of this setup and the kernel trick, it is clear that we cannot apply the method in Sec.~\ref{ssec:linearmiss} for MC-KUoS learning. However, as described in Sec.~\ref{ssec:kernelcomplete}, the solution to the MC-KUoS learning problem using complete data only requires computations of the inner products in $\cF$. In this regard, we propose an estimate of the kernel function value $\kappa(\mathbf{y}_i,\mathbf{y}_j)$ using incomplete data $[\mathbf{y}_i]_{\mathbf{\Omega}_i}$ and $[\mathbf{y}_j]_{\mathbf{\Omega}_j}$. Mathematically, our goal is to find a proxy function $h(\cdot,\cdot)$ such that $h([\mathbf{y}_i]_{\mathbf{\Omega}_i},[\mathbf{y}_j]_{\mathbf{\Omega}_j}) \approx \kappa(\mathbf{y}_i,\mathbf{y}_j)$. To derive this proxy function, we start by considering the relationship between $[\mathbf{y}_i]_{\mathbf{\Omega}_i}, [\mathbf{y}_j]_{\mathbf{\Omega}_j}$ and $\mathbf{y}_i, \mathbf{y}_j$ in the context of different kernel functions.

We first consider isotropic kernels of the form $\kappa(\mathbf{y}_i,\mathbf{y}_j) = k(\| \mathbf{y}_i-\mathbf{y}_j \|_2^2)$ for our analysis. To begin, we define $\mathbf{z}_{ij}^{-} = \mathbf{y}_i - \mathbf{y}_j$ and $\mathbf{\Omega}_{ij} = \mathbf{\Omega}_{i} \cap \mathbf{\Omega}_{j}$, resulting in $[\mathbf{z}_{ij}^{-}]_{\mathbf{\Omega}_{ij}} = [\mathbf{y}_i]_{\mathbf{\Omega}_{ij}} - [\mathbf{y}_j]_{\mathbf{\Omega}_{ij}} \in \R^{|\mathbf{\Omega}_{ij}|}$. For any vector $\mathbf{z}_{ij}^{-}$, the authors in \cite{BalzanoRN.ISIT2010} define the coherence of a subspace spanned by a vector $\mathbf{z}_{ij}^{-}$ to be $\mu(\mathbf{z}_{ij}^{-}) = \frac{m\| \mathbf{z}_{ij}^{-} \|_{\infty}^2}{\| \mathbf{z}_{ij}^{-} \|_2^2}$ and show that $\|[\mathbf{z}_{ij}^{-}]_{\mathbf{\Omega}_{ij}}\|_2^2$ is close to $\frac{|\mathbf{\Omega}_{ij}|}{m} \| \mathbf{z}_{ij}^{-} \|_2^2$ with high probability. Leveraging this result, we can give the following corollary that is essentially due to \cite[Lemma 1]{BalzanoRN.ISIT2010} by plugging in the definition of $\mathbf{z}_{ij}^{-}$ and $[\mathbf{z}_{ij}^{-}]_{\mathbf{\Omega}_{ij}}$.
\begin{corollary}  \label{cor:l2norm}
Let $\delta > 0$, $\mathbf{\Omega}_{ij} = \mathbf{\Omega}_{i} \cap \mathbf{\Omega}_{j}$ and $\alpha = \sqrt{\frac{2 \mu(\mathbf{y}_i-\mathbf{y}_j)^2}{|\mathbf{\Omega}_{ij}|} \log(\frac{1}{\delta})}$. Then with probability at least $1-2\delta$,
\begin{align*}
(1-\alpha) \|\mathbf{y}_i - \mathbf{y}_j\|_2^2 \leq \frac{m}{|\mathbf{\Omega}_{ij}|} \|[\mathbf{z}_{ij}^{-}]_{\mathbf{\Omega}_{ij}}\|_2^2 \leq (1+\alpha) \|\mathbf{y}_i - \mathbf{y}_j\|_2^2.
\end{align*}
\end{corollary}

Using this simple relationship in Corollary~\ref{cor:l2norm}, we can replace the distance term $\| \mathbf{y}_i - \mathbf{y}_j \|_2^2$ in any isotropic kernel function by $\frac{m}{|\mathbf{\Omega}_{ij}|} \|[\mathbf{y}_i]_{\mathbf{\Omega}_{ij}} - [\mathbf{y}_j]_{\mathbf{\Omega}_{ij}}\|_2^2$ and provide an estimate of its true value $\kappa(\mathbf{y}_i,\mathbf{y}_j)$ using entries of $\mathbf{y}_i$ and $\mathbf{y}_j$ that correspond to $\mathbf{\Omega}_{ij}$ only. For example, for the Gaussian kernel $\kappa(\mathbf{y}_i,\mathbf{y}_j) = \exp( - \frac{ \|\mathbf{y}_i-\mathbf{y}_j\|_2^2}{c} )$ with $c>0$, we can replace $\kappa(\mathbf{y}_i,\mathbf{y}_j)$ with $h([\mathbf{y}_i]_{\mathbf{\Omega}_i},[\mathbf{y}_j]_{\mathbf{\Omega}_j}) = \exp( - \frac{m \|[\mathbf{y}_i]_{\mathbf{\Omega}_{ij}} - [\mathbf{y}_j]_{\mathbf{\Omega}_{ij}}\|_2^2}{|\mathbf{\Omega}_{ij}|c} )$ in our algorithms. In this case, the following result provides bounds for estimation of the Gaussian kernel value.
\begin{theorem}  \label{thm:gaussiankernel}
Let $\delta > 0$, $\mathbf{\Omega}_{ij} = \mathbf{\Omega}_{i} \cap \mathbf{\Omega}_{j}$ and $\alpha = \sqrt{\frac{2 \mu(\mathbf{y}_i-\mathbf{y}_j)^2}{|\mathbf{\Omega}_{ij}|} \log(\frac{1}{\delta})}$. Then for a Gaussian kernel $\kappa(\mathbf{y}_i,\mathbf{y}_j)$, with probability at least $1-2\delta$, we have
\begin{align*}
h([\mathbf{y}_i]_{\mathbf{\Omega}_{i}},[\mathbf{y}_j]_{\mathbf{\Omega}_{j}})^{\frac{1}{1 - \alpha}} \leq \kappa(\mathbf{y}_i,\mathbf{y}_j) \leq h([\mathbf{y}_i]_{\mathbf{\Omega}_{i}},[\mathbf{y}_j]_{\mathbf{\Omega}_{j}})^{\frac{1}{1 + \alpha}}.
\end{align*}
\end{theorem}
\noindent We skip the proof of this theorem since it is elementary. We also note here that $h([\mathbf{y}_i]_{\mathbf{\Omega}_{i}},[\mathbf{y}_i]_{\mathbf{\Omega}_{i}}) = \kappa(\mathbf{y}_i,\mathbf{y}_i) = 1$ as a special case for Gaussian kernels.

Next, we consider dot product kernels of the form $\kappa(\mathbf{y}_i,\mathbf{y}_j) = k(\langle \mathbf{y}_i, \mathbf{y}_j \rangle)$, where we again need to estimate $\langle \mathbf{y}_i, \mathbf{y}_j \rangle$ using entries of $\mathbf{y}_i$ and $\mathbf{y}_j$ corresponding to $\mathbf{\Omega}_{ij}$ only. In order to find an estimator of $\langle \mathbf{y}_i, \mathbf{y}_j \rangle$, we define $\mathbf{z}_{ij}^{*} = \mathbf{y}_i \circ \mathbf{y}_j \in \R^m$ to be the coordinate-wise product of $\mathbf{y}_i$ and $\mathbf{y}_j$. This means that $\langle \mathbf{y}_i, \mathbf{y}_j \rangle$ and $\langle [\mathbf{y}_i]_{\mathbf{\Omega}_{ij}}, [\mathbf{y}_j]_{\mathbf{\Omega}_{ij}} \rangle$ equal the sum of all the entries of $\mathbf{z}_{ij}^{*}$ and $[\mathbf{z}_{ij}^{*}]_{\mathbf{\Omega}_{ij}} \in \R^{|\mathbf{\Omega}_{ij}|}$, respectively. We now have the following lemma that describes deviation of the estimated inner product between $\mathbf{y}_i$ and $\mathbf{y}_j$.
\begin{lemma}  \label{lemma:innerproduct}
Let $\delta > 0$, $\mathbf{\Omega}_{ij} = \mathbf{\Omega}_{i} \cap \mathbf{\Omega}_{j}$ and $\beta = \sqrt{\frac{2 m^2 \| \mathbf{y}_i \circ \mathbf{y}_j \|_{\infty}^2}{|\mathbf{\Omega}_{ij}|} \log(\frac{1}{\delta})}$. Then with probability at least $1-2\delta$,
\begin{align}  \label{eqn:dotprod}
\langle \mathbf{y}_i, \mathbf{y}_j \rangle - \beta \leq \frac{m}{|\mathbf{\Omega}_{ij}|} \langle [\mathbf{y}_i]_{\mathbf{\Omega}_{ij}}, [\mathbf{y}_j]_{\mathbf{\Omega}_{ij}} \rangle \leq \langle \mathbf{y}_i, \mathbf{y}_j \rangle + \beta.
\end{align}
\end{lemma}
\begin{IEEEproof}
See Appendix \ref{append:proof}.
\end{IEEEproof}

The above lemma establishes that $\langle [\mathbf{y}_i]_{\mathbf{\Omega}_{ij}}, [\mathbf{y}_j]_{\mathbf{\Omega}_{ij}} \rangle$ is close to $\frac{|\mathbf{\Omega}_{ij}|}{m} \langle \mathbf{y}_i, \mathbf{y}_j \rangle$ with high probability. We once again use this relationship and give an estimate of the corresponding kernel function value. For example, for the polynomial kernel $\kappa(\mathbf{y}_i, \mathbf{y}_j) = (\langle \mathbf{y}_i, \mathbf{y}_j \rangle +c)^d$ with $d>0$ and $c \geq 0$, we have $h([\mathbf{y}_i]_{\mathbf{\Omega}_i},[\mathbf{y}_j]_{\mathbf{\Omega}_j}) = ( \frac{m}{|\mathbf{\Omega}_{ij}|} \langle [\mathbf{y}_i]_{\mathbf{\Omega}_{ij}}, [\mathbf{y}_j]_{\mathbf{\Omega}_{ij}} \rangle +c)^d$. To analyze the bounds on estimated kernel function value in this case, notice that if \eqref{eqn:dotprod} holds and $d$ is odd, we will have
\begin{align*}
(\langle \mathbf{y}_i, \mathbf{y}_j \rangle - \beta + c)^d & \leq (\frac{m}{|\mathbf{\Omega}_{ij}|} \langle [\mathbf{y}_i]_{\mathbf{\Omega}_{ij}}, [\mathbf{y}_j]_{\mathbf{\Omega}_{ij}} \rangle + c)^d \\
& \leq (\langle \mathbf{y}_i, \mathbf{y}_j \rangle + \beta + c)^d.
\end{align*}
But the above inequalities cannot be guaranteed to hold when $d$ is even. Using this, we trivially obtain the theorem below, as a counterpart of Theorem~\ref{thm:gaussiankernel}, for polynomial kernels.
\begin{theorem}  \label{thm:polykernel}
Let $\delta > 0$, $\mathbf{\Omega}_{ij} = \mathbf{\Omega}_{i} \cap \mathbf{\Omega}_{j}$ and $\beta = \sqrt{\frac{2 m^2 \| \mathbf{y}_i \circ \mathbf{y}_j \|_{\infty}^2}{|\mathbf{\Omega}_{ij}|} \log(\frac{1}{\delta})}$. Then for a polynomial kernel $\kappa(\mathbf{y}_i, \mathbf{y}_j)$ with an odd degree $d$, with probability at least $1-2\delta$, we have
\begin{align*}
&(h([\mathbf{y}_i]_{\mathbf{\Omega}_{i}}, [\mathbf{y}_j]_{\mathbf{\Omega}_{j}})^{\frac{1}{d}} - \beta)^d  \leq \kappa(\mathbf{y}_i, \mathbf{y}_j) \\
\text{and}\quad & \kappa(\mathbf{y}_i, \mathbf{y}_j) \leq (h([\mathbf{y}_i]_{\mathbf{\Omega}_{i}}, [\mathbf{y}_j]_{\mathbf{\Omega}_{j}})^{\frac{1}{d}} + \beta)^d.
\end{align*}
\end{theorem}

Based on the discussion above, we can estimate the kernel function value $\kappa(\mathbf{y}_i, \mathbf{y}_j)$ using the associated proxy function $h(\cdot, \cdot)$ that utilizes entries of $\mathbf{y}_i$ and $\mathbf{y}_j$ belonging to $\mathbf{\Omega}_{ij}$ only. Thus, we can compute the estimated kernel matrix $\mathbf{G} \in \R^{N \times N}$ as $g_{i,j} = h([\mathbf{y}_i]_{\mathbf{\Omega}_i},[\mathbf{y}_j]_{\mathbf{\Omega}_j})$ in the case of missing data. But the positive definiteness of $\mathbf{G}$ is not guaranteed in this case. We therefore first need to find a positive definite matrix $\widehat{\mathbf{G}} \approx \mathbf{G}$ before we can carry on with MC-KUoS learning in this setting. To deal with this issue, we begin with eigen decomposition of $\mathbf{G} = \mathbf{U} \mathbf{\Lambda} \mathbf{U}^{T}$, where $\mathbf{\Lambda} = \diag \{\lambda_{\mathbf{G}}^{(1)}, \dots, \lambda_{\mathbf{G}}^{(N)} \}$ contains eigenvalues of $\mathbf{G}$. The resulting approximated kernel matrix $\widehat{\mathbf{G}}$ that is ``closest'' to $\mathbf{G}$ can then be calculated by $\widehat{\mathbf{G}} = \mathbf{U} \widehat{\mathbf{\Lambda}} \mathbf{U}^{T}$, where $\widehat{\mathbf{\Lambda}} = \diag \{ \lambda_{\widehat{\mathbf{G}}}^{(1)}, \dots, \lambda_{\widehat{\mathbf{G}}}^{(N)} \}$ and each $\lambda_{\widehat{\mathbf{G}}}^{(i)}$, $i = 1, \dots, N$, is defined as
\begin{displaymath}
\lambda_{\widehat{\mathbf{G}}}^{(i)} = \left\{ \begin{array}{ll}
\lambda_{\mathbf{G}}^{(i)}, & \textrm{$\lambda_{\mathbf{G}}^{(i)} > 0$} \\
\delta_{min}, & \textrm{$\lambda_{\mathbf{G}}^{(i)} = 0$} \\
-\lambda_{\mathbf{G}}^{(i)}, & \textrm{$\lambda_{\mathbf{G}}^{(i)} < 0$}.
\end{array} \right.
\end{displaymath}
Here, $\delta_{min}>0$ is a predefined (small) parameter. Using the above procedure, one can obtain a positive definite matrix $\widehat{\mathbf{G}}$ such that $\widehat{g}_{i,j} \approx \kappa(\mathbf{y}_i, \mathbf{y}_j)$ and use it for MC-UoS learning in the feature space. Effectively, MC-KUoS learning in the presence of missing data also relies on Algorithm~\ref{algo:MCKUSaL}, with the difference being that we use $\widehat{g}_{i,j}$, obtained from $h([\mathbf{y}_i]_{\mathbf{\Omega}_i},[\mathbf{y}_j]_{\mathbf{\Omega}_j})$, in lieu of $\kappa(\mathbf{y}_i, \mathbf{y}_j)$ in the overall learning process, which includes both kernel subspace assignment and kernel subspace update stages. We dub this approach \emph{robust MC-KUoS learning} (rMC-KUSaL). We conclude this section by noting that we can also robustify classical kernel PCA by using $\widehat{\mathbf{G}}$ as a means of performing kernel PCA with missing data, which we call rKPCA in our experiments.

\subsection{Pre-Image Reconstruction}
\label{ssec:preimage}

Thus far in this section, we have discussed MC-UoS learning in the kernel space with complete and missing data using the kernel trick. Now suppose we are given a new noisy (test) sample $\mathbf{z} = \mathbf{x} + \boldsymbol{\xi} \in \R^m$, where $\boldsymbol{\xi}$ is a noise term and $\widetilde{\phi}(\mathbf{x}) = \phi(\mathbf{x}) - \overline{\boldsymbol{\phi}}$ belongs to one of the subspaces in $\cM_L$ (i.e., $\widetilde{\phi}(\mathbf{x}) \in \cS_{\tau}, \tau \in \{1, \dots, L\}$). In most information processing tasks, one needs to first find a representation of this sample $\mathbf{z}$ in terms of the learned MC-KUoS, which is akin to ``denoising'' $\mathbf{z}$. The ``denoised sample'' in the feature space is the projection of $\phi(\mathbf{z})$ onto $\cS_{\tau}$, which is given by $P_{\cS_{\tau}} \phi(\mathbf{z}) = \mathbf{D}_{\tau}\mathbf{D}_{\tau}^{T} \widetilde{\phi}(\mathbf{z}) + \overline{\boldsymbol{\phi}}$ with $\widetilde{\phi}(\mathbf{z}) = \phi(\mathbf{z}) - \overline{\boldsymbol{\phi}}$. However, in order to visualize the ``denoised'' sample in the ambient space, we often need to project $P_{\cS_{\tau}} \phi(\mathbf{z})$ onto the input space in many applications \cite{Burges.ICML1996,MikaSSMSR.NIPS1998}, which is termed \emph{pre-image reconstruction}. In this section, we consider the problem of pre-image reconstruction based on the MC-KUoS model.

Mathematically, the problem of pre-image reconstruction can be stated as follows. We are given $\mathbf{z} \in \R^m$ and we are interested in finding $\widehat{\mathbf{z}} \in \R^m$ whose mapping to the feature space is closest to the projection of $\phi(\mathbf{z})$ onto the learned MC-UoS in $\cF$. This involves first finding the index $\tau$ such that $\tau = \argmin_{\ell} \| \widetilde{\phi}(\mathbf{z}) - P_{\cS_{\ell}} \widetilde{\phi}(\mathbf{z}) \|_2^2$, which can be easily done using the kernel subspace assignment step described in \eqref{eqn:kernelsubassign}. Next, we need to solve $\widehat{\mathbf{z}} = \argmin_{\boldsymbol{\varrho} \in \R^m} \|\phi(\boldsymbol{\varrho}) - P_{\cS_{\tau}} \phi(\mathbf{z}) \|_2^2$. To solve this problem, we leverage the ideas in \cite{KwokT.TNN2004,RathiDT.SPIE2006} that only use feature-space distances to find $\widehat{\mathbf{z}}$ (equivalently, to find the pre-image of $P_{\cS_{\tau}} \phi(\mathbf{z})$). We first study this problem when the training samples $\mathbf{Y}$ are complete.

\subsubsection{Pre-Image Reconstruction Using Complete Data}
\label{sssec:preimagecomp}

We first calculate the squared ``feature distance'' between $P_{\cS_{\tau}} \phi(\mathbf{z})$ and any $\phi(\mathbf{y}_i)$, $i = 1 \dots, N$, defined as \cite{KwokT.TNN2004}
\begin{align}   \label{eqn:featuredist}
& d_{\mathcal{F}}^2(\phi(\mathbf{y}_i),P_{\cS_{\tau}} \phi(\mathbf{z}))   \nonumber \\
& = \| P_{\cS_{\tau}} \phi(\mathbf{z}) \|_2^2 + \| \phi(\mathbf{y}_i) \|_2^2 - 2 (P_{\cS_{\tau}} \phi(\mathbf{z}))^{T} \phi(\mathbf{y}_i).
\end{align}
Notice that $\| P_{\cS_{\tau}} \phi(\mathbf{z}) \|_2^2$ and $(P_{\cS_{\tau}} \phi(\mathbf{z}))^{T} \phi(\mathbf{y}_i)$ can be calculated in terms of kernel representation as follows:
\begin{align*}
& \| P_{\cS_{\tau}} \phi(\mathbf{z}) \|_2^2 \nonumber \\
& = \widetilde{\phi}(\mathbf{z})^{T}\mathbf{D}_{\tau} \mathbf{D}_{\tau}^{T}\widetilde{\phi}(\mathbf{z}) + \overline{\boldsymbol{\phi}}^{T}\overline{\boldsymbol{\phi}} + 2 \widetilde{\phi}(\mathbf{z})^{T}\mathbf{D}_{\tau} \mathbf{D}_{\tau}^{T}\overline{\boldsymbol{\phi}}  \nonumber \\
& = \widetilde{\phi}(\mathbf{z})^{T} \widetilde{\phi}(\mathbf{Y}_{\tau}) \mathbf{E}_{\tau}\mathbf{E}_{\tau}^{T} \widetilde{\phi}(\mathbf{Y}_{\tau})^{T} \widetilde{\phi}(\mathbf{z}) + \frac{1}{N^2} \boldsymbol{1}_{N}^{T}\mathbf{G}\boldsymbol{1}_{N}   \nonumber \\
&\qquad + \frac{2}{N} \widetilde{\phi}(\mathbf{z})^{T} \widetilde{\phi}(\mathbf{Y}_{\tau}) \mathbf{E}_{\tau} \mathbf{E}_{\tau}^{T} \widetilde{\phi}(\mathbf{Y}_{\tau})^{T}\phi(\mathbf{Y})\boldsymbol{1}_{N}   \nonumber \\
& = \boldsymbol{\widetilde{\psi}}_{\tau}(\mathbf{z})^{T} \mathbf{E}_{\tau}\mathbf{E}_{\tau}^{T} \Big( \boldsymbol{\widetilde{\psi}}_{\tau}(\mathbf{z}) + \frac{2}{N} [\mathbf{G}]_{\mathbf{c}_{\tau},:}\boldsymbol{1}_{N} - \frac{2}{N^2}\boldsymbol{1}_{N_{\tau}} \boldsymbol{1}_{N}^{T} \mathbf{G} \boldsymbol{1}_{N} \Big)  \nonumber \\
&\qquad + \frac{1}{N^2} \boldsymbol{1}_{N}^{T}\mathbf{G}\boldsymbol{1}_{N},
\end{align*}
and
\begin{align*}
& (P_{\cS_{\tau}} \phi(\mathbf{z}))^{T} \phi(\mathbf{y}_i)  \\
& = \boldsymbol{\widetilde{\psi}}_{\tau}(\mathbf{z})^{T} \mathbf{E}_{\tau}\mathbf{E}_{\tau}^{T} \Big( \boldsymbol{\psi}_{\tau}(\mathbf{y}_i) - \frac{1}{N}\boldsymbol{1}_{N_{\tau}} \boldsymbol{1}_{N}^{T}\boldsymbol{k}_{\mathbf{y}_i} \Big) +  \frac{1}{N}\boldsymbol{1}_{N}^{T}\boldsymbol{k}_{\mathbf{y}_i}.
\end{align*}
Therefore, \eqref{eqn:featuredist} becomes
\begin{align*}
& d_{\mathcal{F}}^2(\phi(\mathbf{y}_i),P_{\cS_{\tau}} \phi(\mathbf{z}))  \\
& = \boldsymbol{\widetilde{\psi}}_{\tau}(\mathbf{z})^{T} \mathbf{E}_{\tau}\mathbf{E}_{\tau}^{T} \Big( \boldsymbol{\widetilde{\psi}}_{\tau}(\mathbf{z}) + \frac{2}{N} [\mathbf{G}]_{\mathbf{c}_{\tau},:}\boldsymbol{1}_{N} - 2\boldsymbol{\psi}_{\tau}(\mathbf{y}_i) \\
&\qquad - \frac{2}{N^2}\boldsymbol{1}_{N_{\tau}} \boldsymbol{1}_{N}^{T} \mathbf{G} \boldsymbol{1}_{N} + \frac{2}{N}\boldsymbol{1}_{N_{\tau}} \boldsymbol{1}_{N}^{T}\boldsymbol{k}_{\mathbf{y}_i} \Big) + g_{i,i} \\
&\qquad + \frac{1}{N^2} \boldsymbol{1}_{N}^{T}\mathbf{G}\boldsymbol{1}_{N} - \frac{2}{N}\boldsymbol{1}_{N}^{T}\boldsymbol{k}_{\mathbf{y}_i}
\end{align*}
with $g_{i,i} = \kappa(\mathbf{y}_i,\mathbf{y}_i)$.

We now describe our method for pre-image reconstruction using the Gaussian kernel $\kappa(\mathbf{y}_i,\mathbf{y}_j) = \exp( - \frac{ \|\mathbf{y}_i-\mathbf{y}_j\|_2^2}{c} )$ first. In this case, the problem of minimizing $\|\phi(\widehat{\mathbf{z}}) - P_{\cS_{\tau}} \phi(\mathbf{z}) \|_2^2$ is equivalent to maximizing the function $\rho(\widehat{\mathbf{z}}) = (P_{\cS_{\tau}} \phi(\mathbf{z}))^{T}\phi(\widehat{\mathbf{z}})$ \cite{MikaSSMSR.NIPS1998}, whose extremum can be obtained by setting $\nabla_{\widehat{\mathbf{z}}} \rho = 0$, where $\nabla_{\widehat{\mathbf{z}}} \rho$ denotes the gradient of $\rho$ with respect to $\widehat{\mathbf{z}}$. To do so, we express $\rho(\widehat{\mathbf{z}})$ as
\begin{align}
& \rho(\widehat{\mathbf{z}}) \nonumber \\
& = (  \mathbf{D}_{\tau}\mathbf{D}_{\tau}^{T} \widetilde{\phi}(\mathbf{z}) + \overline{\boldsymbol{\phi}} )^{T} \phi(\widehat{\mathbf{z}}) \nonumber \\
& = \widetilde{\phi}(\mathbf{z})^{T} \widetilde{\phi}(\mathbf{Y}_{\tau}) \mathbf{E}_{\tau} \mathbf{E}_{\tau}^{T} \widetilde{\phi}(\mathbf{Y}_{\tau})^{T} \phi(\widehat{\mathbf{z}}) + \frac{1}{N} \boldsymbol{1}_{N}^{T} \phi(\mathbf{Y})^{T} \phi(\widehat{\mathbf{z}})  \nonumber \\
& = \boldsymbol{\widetilde{\psi}}_{\tau}(\mathbf{z})^{T} \mathbf{E}_{\tau}\mathbf{E}_{\tau}^{T} (\boldsymbol{\psi}_{\tau}(\widehat{\mathbf{z}}) - \frac{1}{N} \boldsymbol{1}_{N_{\tau}} \boldsymbol{1}_{N}^{T} \boldsymbol{k}_{\widehat{\mathbf{z}}} ) + \frac{1}{N} \boldsymbol{1}_{N}^{T} \boldsymbol{k}_{\widehat{\mathbf{z}}}  \nonumber \\
& = \boldsymbol{\zeta}_{\tau}(\mathbf{z})^{T} (\boldsymbol{\psi}_{\tau}(\widehat{\mathbf{z}}) - \frac{1}{N} \boldsymbol{1}_{N_{\tau}} \boldsymbol{1}_{N}^{T} \boldsymbol{k}_{\widehat{\mathbf{z}}} ) + \frac{1}{N} \boldsymbol{1}_{N}^{T} \boldsymbol{k}_{\widehat{\mathbf{z}}},
\end{align}
where $\boldsymbol{\zeta}_{\tau}(\mathbf{z}) = \mathbf{E}_{\tau}\mathbf{E}_{\tau}^{T} \boldsymbol{\widetilde{\psi}}_{\tau}(\mathbf{z}) \in \R^{|N_{\tau}|}$. Next, we define $\boldsymbol{\chi} = \frac{1}{N} (1-\boldsymbol{\zeta}_{\tau}(\mathbf{z})^{T} \boldsymbol{1}_{N_{\tau}}) \boldsymbol{1}_{N} \in \R^{N}$ and let $\widehat{\boldsymbol{\chi}}$ be an $N$-dimensional vector such that $[\widehat{\boldsymbol{\chi}}]_{\mathbf{c}_{\tau}} = [\boldsymbol{\chi}]_{\mathbf{c}_{\tau}} + \boldsymbol{\zeta}_{\tau}(\mathbf{z})$ and $[\widehat{\boldsymbol{\chi}}]_{\boldsymbol{\mathcal{I}}_N \setminus \mathbf{c}_{\tau}} = [\boldsymbol{\chi}]_{\boldsymbol{\mathcal{I}}_N \setminus \mathbf{c}_{\tau}}$ (recall that $\boldsymbol{\mathcal{I}}_N = \{1,\dots,N\}$ and $\mathbf{c}_{\tau}$ contains all the indices of $\widetilde{\phi}(\mathbf{y}_i)$'s that are assigned to $\cS_{\tau}$), which means $\rho(\widehat{\mathbf{z}}) = \widehat{\boldsymbol{\chi}}^{T} \boldsymbol{k}_{\widehat{\mathbf{z}}} = \sum_{i=1}^{N} \widehat{\boldsymbol{\chi}}_{(i)} \kappa(\widehat{\mathbf{z}},\mathbf{y}_i)$. By setting $\nabla_{\widehat{\mathbf{z}}} \rho = 0$, we get
\begin{align*}
\widehat{\mathbf{z}} = \frac{ \sum_{i=1}^{N} \widehat{\boldsymbol{\chi}}_{(i)} \exp( - \|\widehat{\mathbf{z}}-\mathbf{y}_i\|_2^2/c ) \mathbf{y}_i }{ \sum_{i=1}^{N} \widehat{\boldsymbol{\chi}}_{(i)} \exp( - \|\widehat{\mathbf{z}}-\mathbf{y}_i\|_2^2/c ) }.
\end{align*}
By using the approximation $P_{\cS_{\tau}} \phi(\mathbf{z}) \approx \phi(\widehat{\mathbf{z}})$ and the relation $\|\widehat{\mathbf{z}}-\mathbf{y}_i\|_2^2 = -c\log ( \frac{1}{2}(2- d_{\mathcal{F}}^2(\phi(\mathbf{y}_i), \phi(\widehat{\mathbf{z}}) )) )$ \cite{KwokT.TNN2004}, a unique pre-image can now be obtained by the following formula:
\begin{align}   \label{eqn:gaussianpreimage}
\widehat{\mathbf{z}} = \frac{ \sum_{i=1}^{N} \widehat{\boldsymbol{\chi}}_{(i)} \Big( \frac{1}{2} \big(2 - d_{\mathcal{F}}^2(  P_{\cS_{\tau}} \phi(\mathbf{z}), \phi(\mathbf{y}_i)  ) \big) \Big) \mathbf{y}_i }{ \sum_{i=1}^{N} \widehat{\boldsymbol{\chi}}_{(i)} \Big( \frac{1}{2} \big(2 - d_{\mathcal{F}}^2(  P_{\cS_{\tau}} \phi(\mathbf{z}), \phi(\mathbf{y}_i)  ) \big) \Big) }.
\end{align}

Next, for the polynomial kernel $\kappa(\mathbf{y}_i, \mathbf{y}_j) = (\langle \mathbf{y}_i, \mathbf{y}_j \rangle +c)^d$ with an odd degree $d$, we can follow a similar procedure and have the following expression for an approximate solution for pre-image reconstruction:
\begin{align}   \label{eqn:polypreimage}
\widehat{\mathbf{z}} = \sum_{i=1}^{N} \widehat{\boldsymbol{\chi}}_{(i)} \Big( \frac{ (P_{\cS_{\tau}} \phi(\mathbf{z}))^{T} \phi(\mathbf{y}_i) }{ \| P_{\cS_{\tau}} \phi(\mathbf{z}) \|_2^2} \Big)^{\frac{d-1}{d}} \mathbf{y}_i.
\end{align}

\subsubsection{Pre-Image Reconstruction Using Missing Data}
\label{sssec:preimagemiss}

We next consider the problem of reconstructing the pre-image of $P_{\cS_{\tau}} \phi(\mathbf{z})$ when the training samples have missing entries. As can be easily seen from \eqref{eqn:gaussianpreimage}, the solution of a pre-image for the Gaussian kernel can be written as $\widehat{\mathbf{z}} = \frac{ \sum_{i=1}^{N} e_i \mathbf{y}_i }{ \sum_{i=1}^{N} e_i }$, where $e_i = \widehat{\boldsymbol{\chi}}_{(i)} \big( \frac{1}{2} (2 - d_{\mathcal{F}}^2(  P_{\cS_{\tau}} \phi(\mathbf{z}), \phi(\mathbf{y}_i)  ) )  \big)$. Similarly, from \eqref{eqn:polypreimage}, we can also write the solution of $\widehat{\mathbf{z}}$ to be $\widehat{\mathbf{z}} = \sum_{i=1}^{N} e_i \mathbf{y}_i$ for the polynomial kernel, where $e_i = \widehat{\boldsymbol{\chi}}_{(i)} \big( \frac{ (P_{\cS_{\tau}} \phi(\mathbf{z}))^{T} \phi(\mathbf{y}_i) }{ \| P_{\cS_{\tau}} \phi(\mathbf{z}) \|_2^2} \big)^{\frac{d-1}{d}}$ in this case. In words, the pre-image solution is a linear combination of the training data, where the weights $e_i$'s can be explicitly computed using the respective kernel functions. In this regard, as described in Sec.~\ref{ssec:kernelmiss}, for each $i = 1,\dots,N$, we can estimate $\kappa(\mathbf{z},\mathbf{y}_i)$ using entries of $\mathbf{z}$ belonging to $\mathbf{\Omega}_i$ (i.e., $[\mathbf{z}]_{\mathbf{\Omega}_i}$) and $[\mathbf{y}_i]_{\mathbf{\Omega}_i}$, where the estimated kernel function value is denoted by $h(\mathbf{z},[\mathbf{y}_i]_{\mathbf{\Omega}_i})$.

Based on the estimated kernel function values $h(\mathbf{z},[\mathbf{y}_i]_{\mathbf{\Omega}_i})$'s, we can then find the solution of $\tau$ such that $\tau = \argmin_{\ell} \| \widetilde{\phi}(\mathbf{z}) - P_{\cS_{\ell}} \widetilde{\phi}(\mathbf{z}) \|_2^2$, and calculate the weights $e_i$'s ($i = 1, \dots, N$). Note that unlike the complete data case, we do need to compute the entries of $\widehat{\mathbf{z}}$ separately in this case. To be specific, for the $u$-th entry of $\widehat{\mathbf{z}}$, $u = 1, \dots, m$, we define $\mathbf{r}_u$ to be the set containing the indices of the samples $\mathbf{y}_i$'s whose $u$-th entry are observed. Then $\widehat{\mathbf{z}}_{(u)} = \frac{ \sum_{i \in \mathbf{r}_u} e_i {\mathbf{y}_i}_{(u)} }{ (\sum_{i=1}^{N} e_i) |\mathbf{r}_u| /N }$ for the Gaussian kernel and $\widehat{\mathbf{z}}_{(u)} = \sum_{i \in \mathbf{r}_u} e_i {\mathbf{y}_i}_{(u)}$ for the polynomial kernel. We conclude this section by noting that the methods described in here can also be applied to the case when the test sample $\mathbf{z}$ has missing entries.

\section{Experimental Results}
\label{sec:experiment}

In this section, we present several experimental results demonstrating the
effectiveness of our proposed methods for data representation. In particular,
we are interested in learning an MC-UoS from complete/missing noisy training
data, followed by denoising of complete noisy test samples using the learned
geometric structures. In the case of MC-KUoS learning, we evaluate the
performance of our algorithms by focusing on ($i$) denoising of complete
noisy test samples, and ($ii$) clustering of complete/missing training data.

\subsection{Experiments for MC-UoS Learning}
\label{ssec:linearexperiment}

In this section, we examine the effectiveness of MC-UoS learning using
Algorithms~\ref{algo:MiCUSaL}--\ref{algo:rMiCUSaL}. For the complete data
experiments, we compare MiCUSaL/aMiCUSaL with several state-of-the-art UoS
learning algorithms such as Block-Sparse Dictionary Design (SAC+BK-SVD)
\cite{Zelnik-ManorRE.TSP2012}, $K$-subspace clustering ($K$-sub)
\cite{HoYLLK.CVPR2003}, Sparse Subspace Clustering (SSC)
\cite{ElhamifarV.PAMI2013}, Robust Sparse Subspace Clustering (RSSC)
\cite{SoltanolkotabiEC.AS2014}, Robust Subspace Clustering via Thresholding
(TSC) \cite{HeckelB.arxiv2013}, as well as with Principal Component Analysis
(PCA) \cite{Hotelling.JEP1933}. In the case of the UoS learning algorithms,
we use codes provided by their respective authors. In the case of SSC, we use
the noisy variant of the optimization program in \cite{ElhamifarV.PAMI2013}
and set $\lambda_{z} = \alpha_z/\mu_z$ in all experiments, where
$\lambda_z$ and $\mu_z$ are as defined in \cite[(13) \&
(14)]{ElhamifarV.PAMI2013}, while the parameter $\alpha_z$ varies in
different experiments. In the case of RSSC, we set $\lambda = 1/ \sqrt{s}$ as
per \cite{SoltanolkotabiEC.AS2014}, while the tuning parameter in TSC is set
$q = \max (3, \lceil N/(L \times 20) \rceil)$ when $L$ is provided. For the case of training
data having missing entries, we compare the results of rMiCUSaL with $k$-GROUSE
\cite{BalzanoSRN.SSP2012} and GROUSE
\cite{BalzanoNR.Allerton2010}.\footnote{As discussed in
\cite{WuB.ICASSP2014}, we omit the results for SSC with missing data in this
paper because it fills in the missing entries with random values, resulting
in relatively poor performance for problems with missing data.}

In order to generate noisy training and test data in these experiments, we
start with sets of ``clean'' training and test samples, denoted by
$\mathbf{X}$ and $\mathbf{X}^{te}$, respectively. We then add white Gaussian
noise to these samples to generate noisy training and test samples
$\mathbf{Y}$ and $\mathbf{Z}$, respectively. In the following, we use
$\sigma_{tr}^2$ and $\sigma_{te}^2$ to denote variance of noise added to
training and test samples, respectively. In the missing data experiments, for
every fixed noise variance $\sigma_{tr}^2$, we create training (but not test)
data with different percentages of missing values, where the number of
missing entries is set to be $10\%$, $30\%$ and $50\%$ of the signal
dimension. Our reported results are based on random initializations
of MiCUSaL and aMiCUSaL algorithms. In this regard, we adopt the following
simple approach to mitigate any stability issues that might arise due to
random initializations. We perform multiple random initializations for every
fixed $\mathbf{Y}$ and $\lambda$, and then retain the learned MC-UoS
structure that results in the smallest value of the objective function in
\eqref{eqn:linearproblem}. We also use a similar approach for selecting the
final structures returned by $K$-subspace clustering and Block-Sparse
Dictionary Design, with the only difference being that
\eqref{eqn:linearproblem} in this case is replaced by the approximation error
of training data.

\subsubsection{Experiments on Synthetic Data}
\label{sssec:syntheticlinear}

In the first set of synthetic experiments, we consider $L=5$ subspaces of the same dimension $s=13$ in an $m=180$-dimensional ambient space. The five subspaces $\cS_{\ell}$'s of $\R^{180}$ are defined by their orthonormal bases $\{ \mathbf{T}_{\ell} \in \R^{m \times s} \}_{\ell=1}^{5}$ as follows. We start with a random orthonormal basis $\mathbf{T}_1 \in \R^{m \times s}$ and for every $\ell \geq 2$, we set $ \mathbf{T}_{\ell} = \orth( \mathbf{T}_{\ell-1} + t_s \mathbf{W}_{\ell}) $ where every entry in $\mathbf{W}_{\ell} \in \R^{m \times s}$ is a uniformly distributed random number between $0$ and $1$, and $\orth(\cdot)$ denotes the orthogonalization process. The parameter $t_s$ controls the distance between subspaces, and we set $t_s = 0.04$ in these experiments.

After generating the subspaces, we generate a set of $n_{\ell}$ points from
$\cS_{\ell}$ as $\mathbf{X}_{\ell} = \mathbf{T}_{\ell} \mathbf{C}_{\ell}$,
where $\mathbf{C}_{\ell} \in \R^{s \times n_{\ell}}$ is a matrix
whose elements are drawn independently and identically from $\mathcal{N}(0,1)$ distribution.
In here, we set $n_1 = n_3 = n_5 = 150$, and $n_2 = n_4 = 100$; hence, $N =
650$. We then stack all the data into a matrix $\mathbf{X} =
[\mathbf{X}_1,\dots, \mathbf{X}_5] =  \{\mathbf{x}_i\}_{i=1}^{N}$ and
normalize all the samples to unit $\ell_2$ norms. Test data $\mathbf{X}^{te}
\in \R^{m \times N}$ are produced using the same foregoing strategy. Then we
add white Gaussian noise with different expected noise power to both
$\mathbf{X}$ and $\mathbf{X}^{te}$. Specifically, we set $\sigma_{tr}^2$
to be $0.1$, while $\sigma_{te}^2$ ranges from $0.1$ to $0.5$. We generate $\mathbf{X}$ and $\mathbf{X}^{te}$ $10$ times, while
Monte Carlo simulations for noisy data are repeated $20$ times for every
fixed $\mathbf{X}$ and $\mathbf{X}^{te}$. Therefore, the results reported in
here correspond to an average of $200$ random trials.

\begin{table*}[htbp]
\centering \caption{$d_{avg}$ of different UoS learning algorithms for synthetic data}
\begin{tabular}{c|c|c|c|c|c|c}
\hline
$d_{avg}$ & \multicolumn{6}{c}{Algorithms}  \\
\hline
\multirow{2}{*}{Complete} & MiCUSaL($\lambda=2$) & SAC+BK-SVD & $K$-sub & SSC & RSSC & TSC \\
\cline{2-7}
   & \textbf{0.1331} & 0.2187 & 0.1612 & 0.1386 & 0.2215 & 0.2275 \\
\hline
\multirow{2}{*}{Missing} & rMiCUSaL-$10\%$ & kGROUSE-$10\%$ & rMiCUSaL-$30\%$ & kGROUSE-$30\%$ & rMiCUSaL-$50\%$ & kGROUSE-$50\%$ \\
\cline{2-7}
   & \textbf{0.1661} & 0.3836 & \textbf{0.1788} & 0.4168 & \textbf{0.2047} & 0.4649  \\
\hline
\end{tabular}
\label{tab:davg}
\end{table*}

\begin{figure}[htbp]
\centering
\subfigure[]{\includegraphics[width=1.6in]{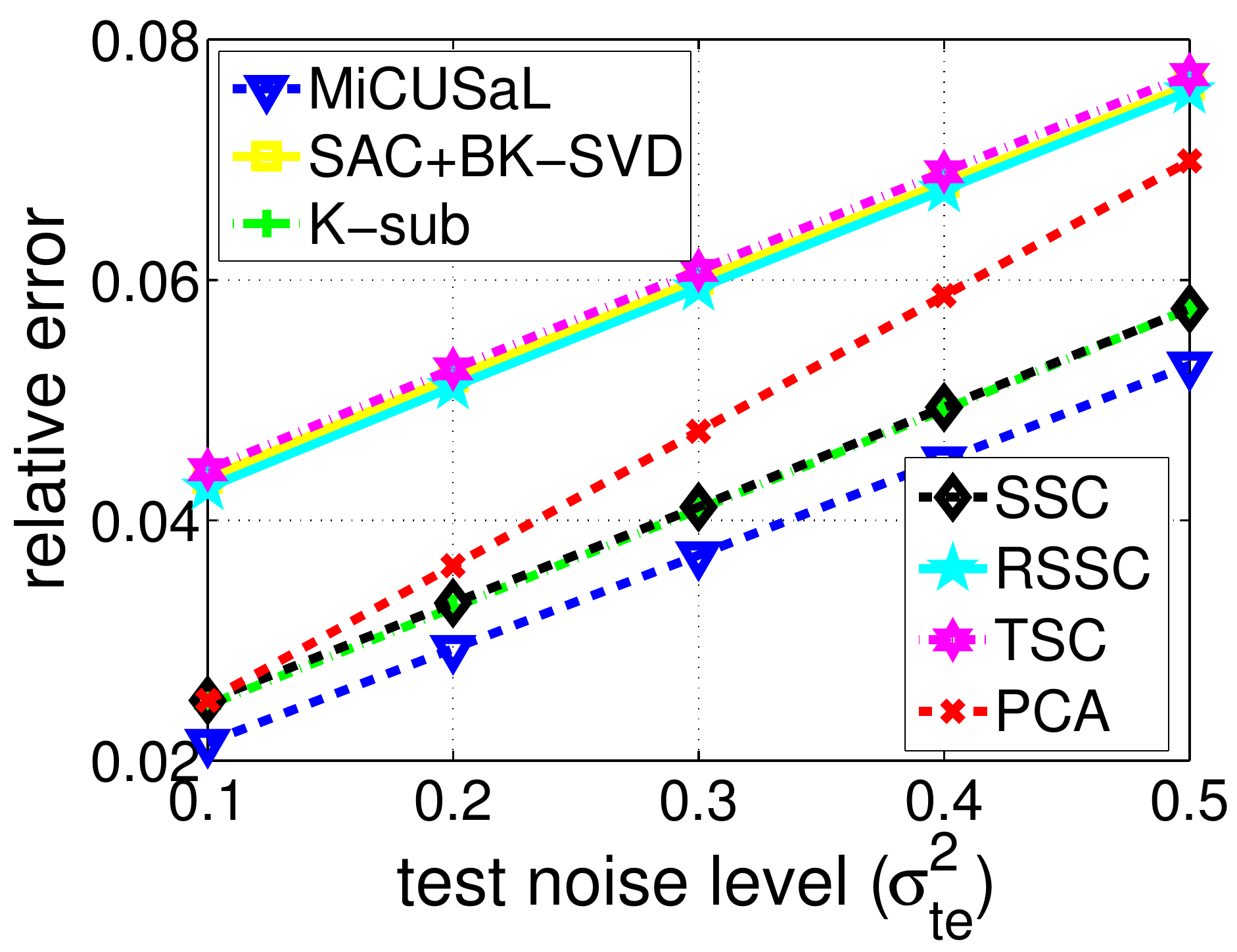} \label{fig:SyntheticCompleteDenoise10}}
\subfigure[]{\includegraphics[width=1.6in]{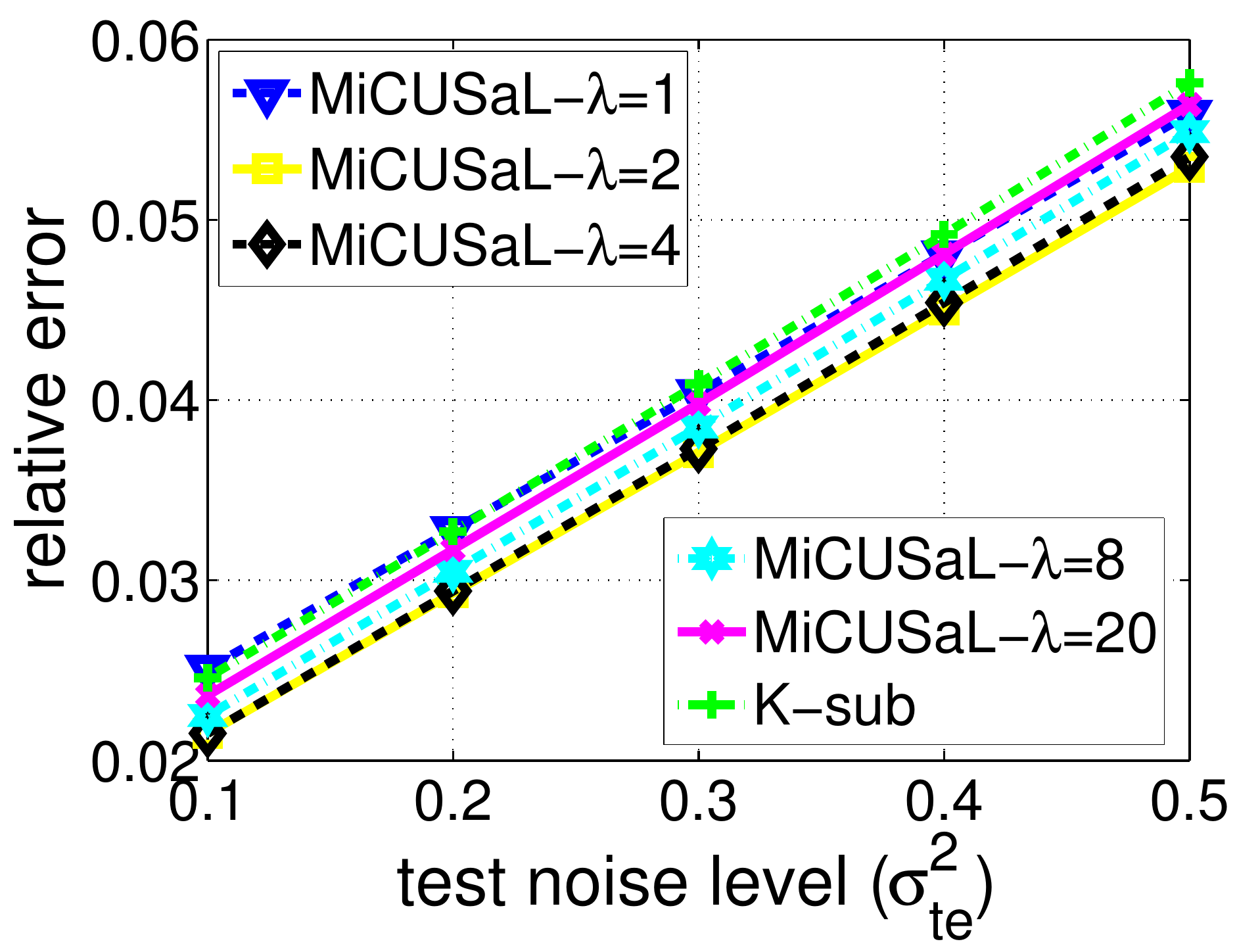} \label{fig:SyntheticCompleteTrace10}}
\subfigure[]{\includegraphics[width=1.6in]{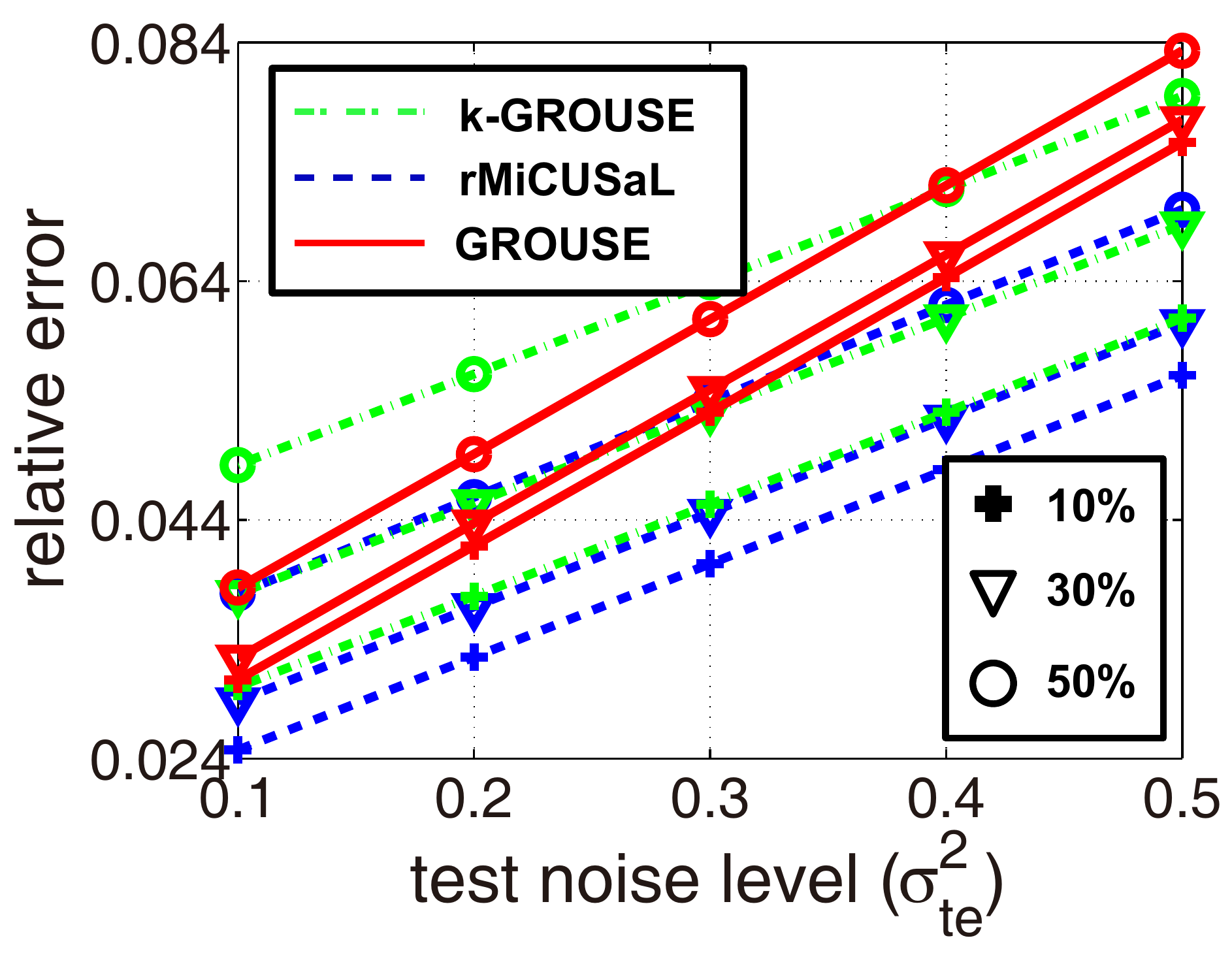} \label{fig:SyntheticMissDenoise10}}
\subfigure[]{\includegraphics[width=1.6in]{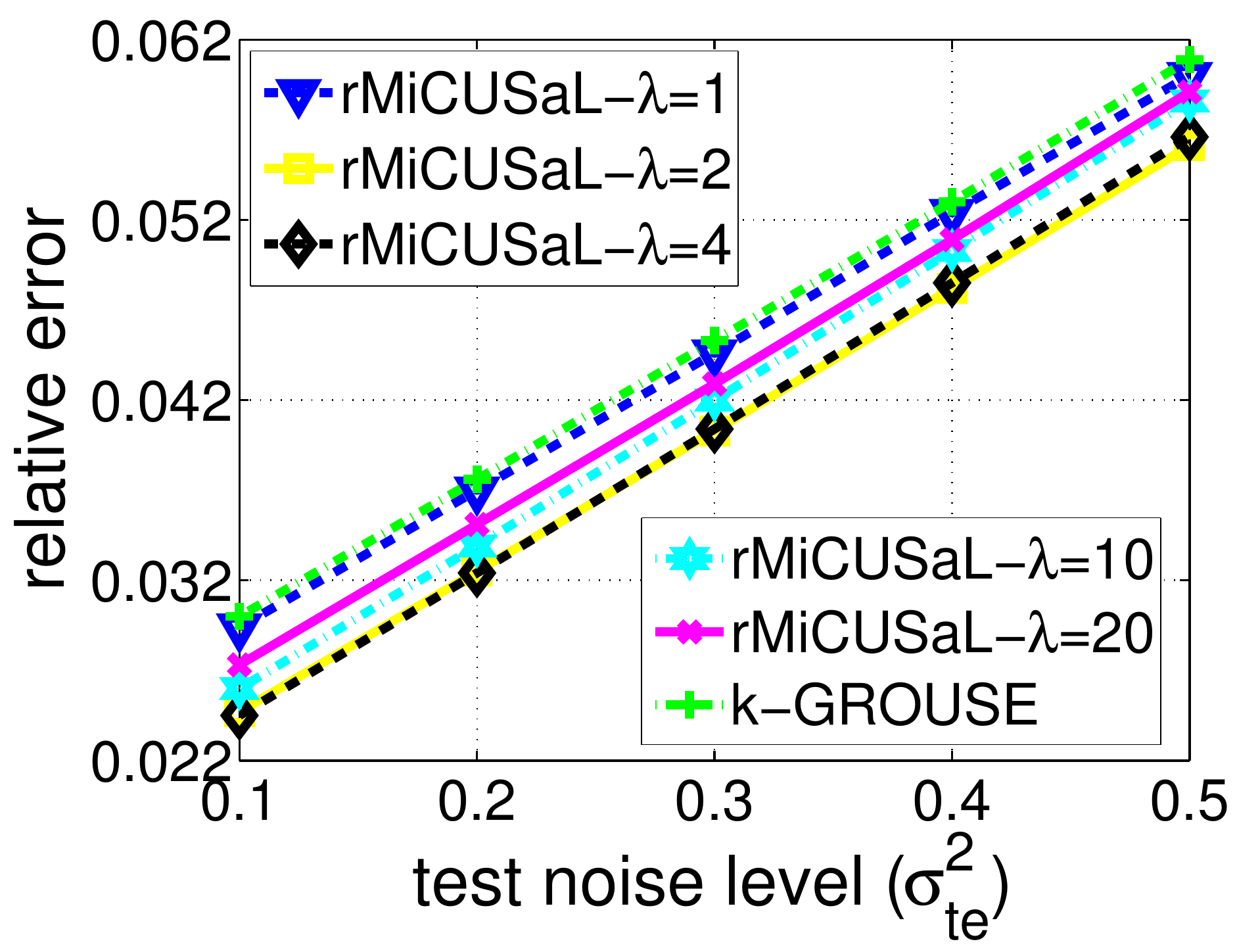} \label{fig:SyntheticMissTrace10}}
\caption{Comparison of MC-UoS learning performance on synthetic data. (a) and (c) show relative errors of test signals for complete and missing data experiments, where $\lambda=2$ for both MiCUSaL and rMiCUSaL. The numbers in the legend of (c) indicate the percentages of missing entries within the training data. (b) and (d) show relative errors of test signals for MiCUSaL and rMiCUSaL (with $10\%$ missing entries) using different $\lambda$'s.}
\label{fig:synthetic}
\end{figure}

We make use of the collection of noisy samples, $\mathbf{Y}$, to learn
a union of $L$ subspaces of dimension $s$ and stack the learned orthonormal
bases $\{\mathbf{D}_{\ell}\}_{\ell=1}^{L}$ into $\mathbf{D}$. In this set of
experiments, we use MiCUSaL and rMiCUSaL for complete and missing data
experiments, respectively. The number of random initializations used
to select the final geometric structure in these experiments is $8$ for every
fixed $\mathbf{Y}$ and $\lambda$. We use the following metrics for performance analysis of MC-UoS learning. Since
we have knowledge of the ground truth $\cS_{\ell}$'s, represented by their
ground truth orthonormal bases $\mathbf{T}_{\ell}$'s, we first find the pairs
of estimated and true subspaces that are the best match, i.e.,
$\mathbf{D}_{\ell}$ is matched to $\mathbf{T}_{\widehat{\ell}}$ using
$\widehat{\ell} = \argmax_p \| \mathbf{D}_{\ell}^{T} \mathbf{T}_{p}
\|_{F}$. We also ensure that no two $\mathbf{D}_{\ell}$'s are matched to the
same $\mathbf{T}_{\widehat{\ell}}$. Then we define $d_{avg}$ to be the
\emph{average normalized subspace distances} between these pairs, i.e., $
d_{avg} = \frac{1}{L} \sum_{\ell=1}^{L} \sqrt{ \frac{s - \tr(\mathbf{D}_{\ell}^{T} \mathbf{T}_{\widehat{\ell}}
\mathbf{T}_{\widehat{\ell}}^{T} \mathbf{D}_{\ell}) }{s} }$. A smaller
$d_{avg}$ indicates better performance of MC-UoS learning. Also, if the
learned subspaces are close to the ground truth, they are expected to have
good representation performance on test data. A good measure in this regard
is the \emph{mean of relative reconstruction errors of the test
samples} using learned subspaces. To be specific, if the training data are
complete, we first represent every test signal $\mathbf{z}$ such that
$\mathbf{z} \approx \mathbf{D}_{\tau} \boldsymbol{\alpha}^{te} +
\bar{\mathbf{y}}$ where $\tau = \argmax_{\ell} \|\mathbf{D}_{\ell}^T
(\mathbf{z} - \bar{\mathbf{y}})\|_2^2$ (recall that $\bar{\mathbf{y}} = \frac{1}{N} \sum_{i=1}^{N} \mathbf{y}_i$) and $\boldsymbol{\alpha}^{te} =
\mathbf{D}_{\tau}^T (\mathbf{z} - \bar{\mathbf{y}})$. The relative
reconstruction error with respect to its noiseless part, $\mathbf{x}$, is
then defined as $\frac{\|\mathbf{x} - (\mathbf{D}_{\tau}
\boldsymbol{\alpha}^{te} + \bar{\mathbf{y}})\|_2^2}{\| \mathbf{x} \|_2^2}$.
On the other hand, if the training data have missing entries then for a test
signal $\mathbf{z}$, the reconstruction error with respect to $\mathbf{x}$ is
simply calculated by $\frac{\|\mathbf{x} - \mathbf{D}_{\tau}
\mathbf{D}_{\tau}^T \mathbf{z}\|_2^2}{\| \mathbf{x} \|_2^2}$, where $\tau =
\argmax_{\ell} \|\mathbf{D}_{\ell}^T \mathbf{z}\|_2^2$.

To compare with other UoS learning methods, we choose $\lambda=2$ for both
MiCUSaL and rMiCUSaL. In the complete data experiments, we perform SSC with
$\alpha_z=60$. We set the subspace dimension for PCA to be the (unrealizable,
ideal) one that yields the best denoising result on \emph{training} samples.
We also use the same subspace dimension (again, unrealizable and
ideal) for GROUSE in the corresponding missing data experiments. Table~\ref{tab:davg} summarizes the $d_{avg}$'s of different UoS learning
algorithms for both complete and missing data experiments. As can be seen,
MiCUSaL produces smaller $d_{avg}$'s, which in turn leads to smaller relative
errors of test data; see Fig.~\ref{fig:SyntheticCompleteDenoise10} for a
validation of this claim. For MC-UoS learning with missing data, rMiCUSaL also learns a better MC-UoS
in that: ($i$) for a fixed percentage of the number of missing observations
in the training data, the $d_{avg}$ for rMiCUSaL is much smaller than the one
for $k$-GROUSE (see Table~\ref{tab:davg}); and ($ii$) rMiCUSaL outperforms
$k$-GROUSE and GROUSE in terms of smaller reconstruction errors of test data.
Moreover, we can infer from Fig.~\ref{fig:SyntheticMissDenoise10} that for a
fixed $\sigma_{te}$, when the number of missing entries increases, the
performance of rMiCUSaL degrades less compared to $k$-GROUSE. We also test
the UoS learning performance with complete data when the
subspaces are not close to each other (e.g., $t_s=0.2$). In this case, all
the UoS learning algorithms, including MiCUSaL, learn the subspaces
successfully. We omit these plots because of space constraints.

\begin{table}[htbp]
\centering \caption{$d_{avg}$ of MiCUSaL and rMiCUSaL for different $\lambda$'s using synthetic data}
\begin{tabular}{c|c|c|c|c|c}
\hline
$d_{avg}$ & \multicolumn{5}{c}{$\lambda$}  \\
\hline
\multirow{2}{*}{MiCUSaL} & $\lambda=1$ & $\lambda=2$ & $\lambda=4$ & $\lambda=8$ & $\lambda=20$ \\
\cline{2-6}
   & 0.1552 & 0.1331 & \textbf{0.1321} & 0.1378 & 0.1493  \\
\hline
rMiCUSaL & $\lambda=1$ & $\lambda=2$ & $\lambda=4$ & $\lambda=10$ & $\lambda=20$ \\
\cline{2-6}
 ($10\%$ missing)  & 0.2096 & \textbf{0.1661} & 0.1725 & 0.2065 & 0.2591  \\
\hline
\end{tabular}
\label{tab:davglambda}
\end{table}

For both MiCUSaL and rMiCUSaL, we also analyze the effect of the key
parameter, $\lambda$, on the UoS learning performance. We implement MiCUSaL
with $\lambda \in \{1,2,4,8,20\}$ in the complete data experiments and
select $\lambda \in \{1,2,4,10,20\}$ for rMiCUSaL in the missing data
experiments, where the number of missing entries in the training data is
$10\%$ of the signal dimension. The results are shown in Fig.~\ref{fig:SyntheticCompleteTrace10},
Fig.~\ref{fig:SyntheticMissTrace10} and Table~\ref{tab:davglambda}. We can see when $\lambda=1$, both the $d_{avg}$'s and
reconstruction errors of the test data are large for MiCUSaL and
rMiCUSaL. This is because the learned subspaces are too close to each other,
which results in poor data representation capability of the learned
$\mathbf{D}_{\ell}$'s. When $\lambda=2$ or $4$, both these algorithms achieve good
performance in terms of small $d_{avg}$'s and relative errors of test data.
As $\lambda$ increases further, both $d_{avg}$ and relative errors of test
data also increase. Furthermore, as $\lambda$ grows, the curves of
relative errors of test data for MiCUSaL and rMiCUSaL get closer to the ones
for $K$-sub and $k$-GROUSE, respectively. This phenomenon coincides with our
discussion in Sec.~\ref{sec:linearsolver}. Finally, we note that both MiCUSaL
and rMiCUSaL achieve their best performance when $\lambda \in [2,4]$, and
deviations of the representation errors of test data are very small when
$\lambda$ falls in this range.

Next, we study the effect of random initialization of subspaces on
MiCUSaL performance by calculating the standard deviation of the mean of the
reconstruction errors of the test data for the $8$ random
initializations. The mean of these $200$ standard deviations ends up being
only $0.003$ for all $\sigma_{te}$'s when $\lambda=2$. In addition, as
$\lambda$ gets larger, the variation of the results increases only slightly
(the mean of the standard deviations is $0.0034$ for $\lambda=8$). On the
other hand, the mean of the standard deviations for $K$-sub is $0.0039$.
Furthermore, the performance gaps between MiCUSaL and all other
methods are larger than $0.003$. Finally, the learned MC-UoS structure that
results in the smallest value of the objective function
\eqref{eqn:linearproblem} always results in the best denoising performance. This
suggests that MiCUSaL always generates the best results and it is mostly
insensitive to the choice of initial subspaces during the random
initialization.

\begin{table*}[htbp]
\centering \caption{Running time comparison (in sec) for rMiCUSaL and $k$-GROUSE}
\begin{tabular}{c|c|c|c|c|c|c}
\hline
Data & \multicolumn{3}{c|}{rMiCUSaL} & \multicolumn{3}{c}{$k$-GROUSE} \\
\hline
\multirow{3}{*}{Synthetic, $m=180$, $N=650$, $L=5$, $s=13$}  &  \multicolumn{3}{c|}{Missing entries ($\%$)} & \multicolumn{3}{c}{Missing entries ($\%$)} \\
\cline{2-7}
  & $10\%$ & $30\%$ & $50\%$ & $10\%$ & $30\%$ & $50\%$ \\
\cline{2-7}
  & 8.02 & 7.41 & 6.62 & 7.46 & 6.95 & 6.11 \\
\hline
\multirow{3}{*}{San Francisco, $m=600$, $N=722$, $L=5$, $s=12$}   &  \multicolumn{3}{c|}{Missing entries ($\%$)} & \multicolumn{3}{c}{Missing entries ($\%$)} \\
\cline{2-7}
  & $10\%$ & $30\%$ & $50\%$ & $10\%$ & $30\%$ & $50\%$ \\
\cline{2-7}
  & 23.19 & 22.46 & 20.31 & 16.56 & 14.75 & 12.53 \\
\hline
\end{tabular}
\label{tab:timecomp}
\end{table*}

We also examine the running times of rMiCUSaL and $k$-GROUSE per iteration,
which include both the subspace assignment and subspace update stages. For
each subspace $\cS_{\ell}$, we implement the optimization over
$\mathbf{D}_{\ell}$ (i.e., Steps $8$ to $17$ in Algorithm~\ref{algo:rMiCUSaL})
for $100$ iterations. All experiments are carried out using Matlab R2013a on
an Intel i7-2600 3.4GHz CPU with 16 GB RAM. From the fourth row of
Table~\ref{tab:timecomp}, we observe rMiCUSaL takes slightly more time
compared to $k$-GROUSE because rMiCUSaL needs two more steps for updating
$\mathbf{D}_{\ell}$ (Steps $10$ and $11$ in Algorithm~\ref{algo:rMiCUSaL}).
However, the advantage of rMiCUSaL over $k$-GROUSE in learning a better UoS
significantly outweighs this slight increase in computational complexity. We
can also see that as the number of missing entries increases, both algorithms
become faster. The reason for this is that when $|\mathbf{\Omega}_i|$
decreases for all $i$'s, less time is needed during the subspace assignment
step and for computing $\boldsymbol{\theta}$ and $\mathbf{r}$ in
Algorithm~\ref{algo:rMiCUSaL}.

\subsubsection{Experiments on City Scene Data}
\label{sssec:realcity}

\begin{figure}[t]
\centering
\subfigure[]{\includegraphics[width=1.6in]{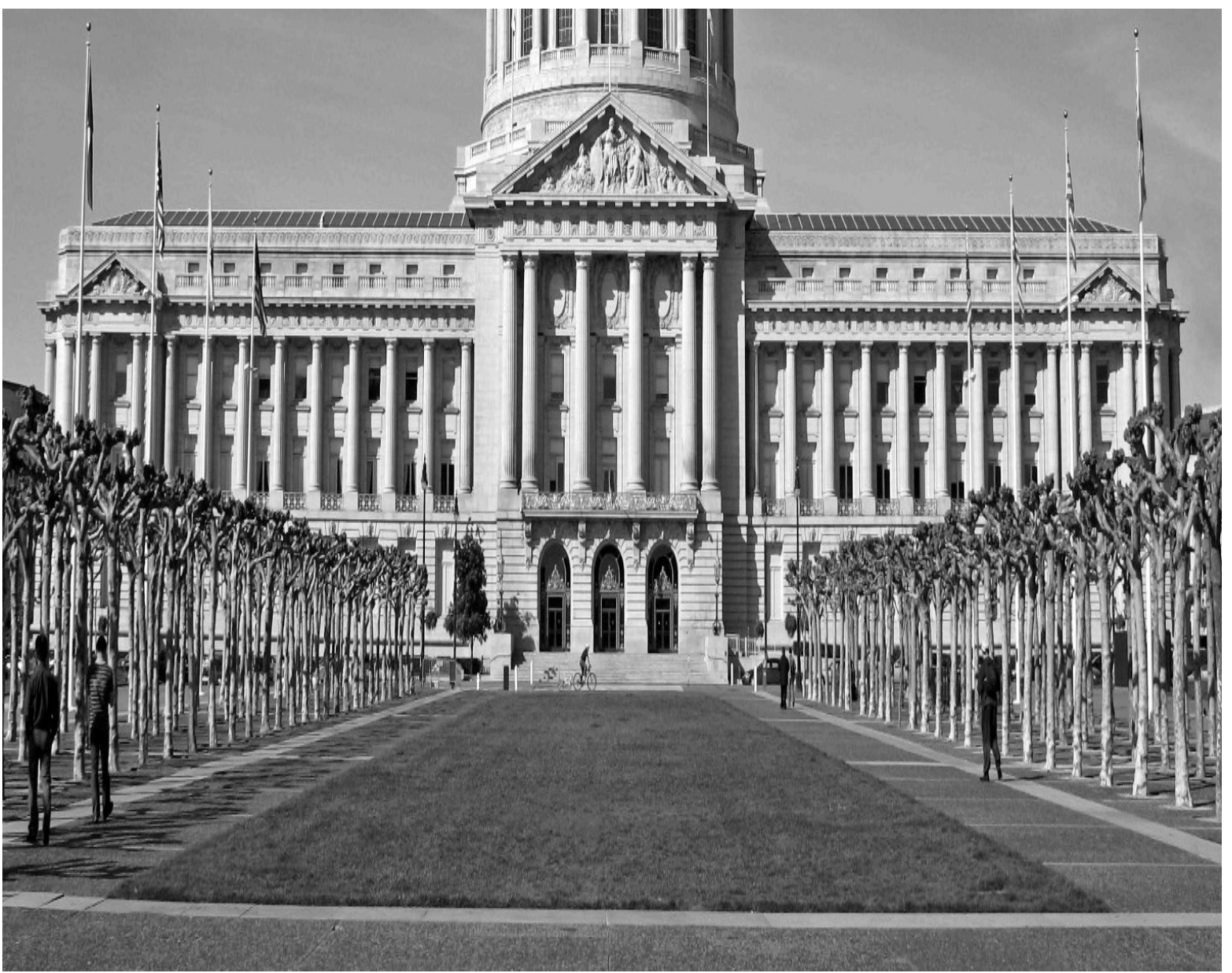} \label{fig:Sanfran}}
\quad
\subfigure[]{\includegraphics[width=1.6in]{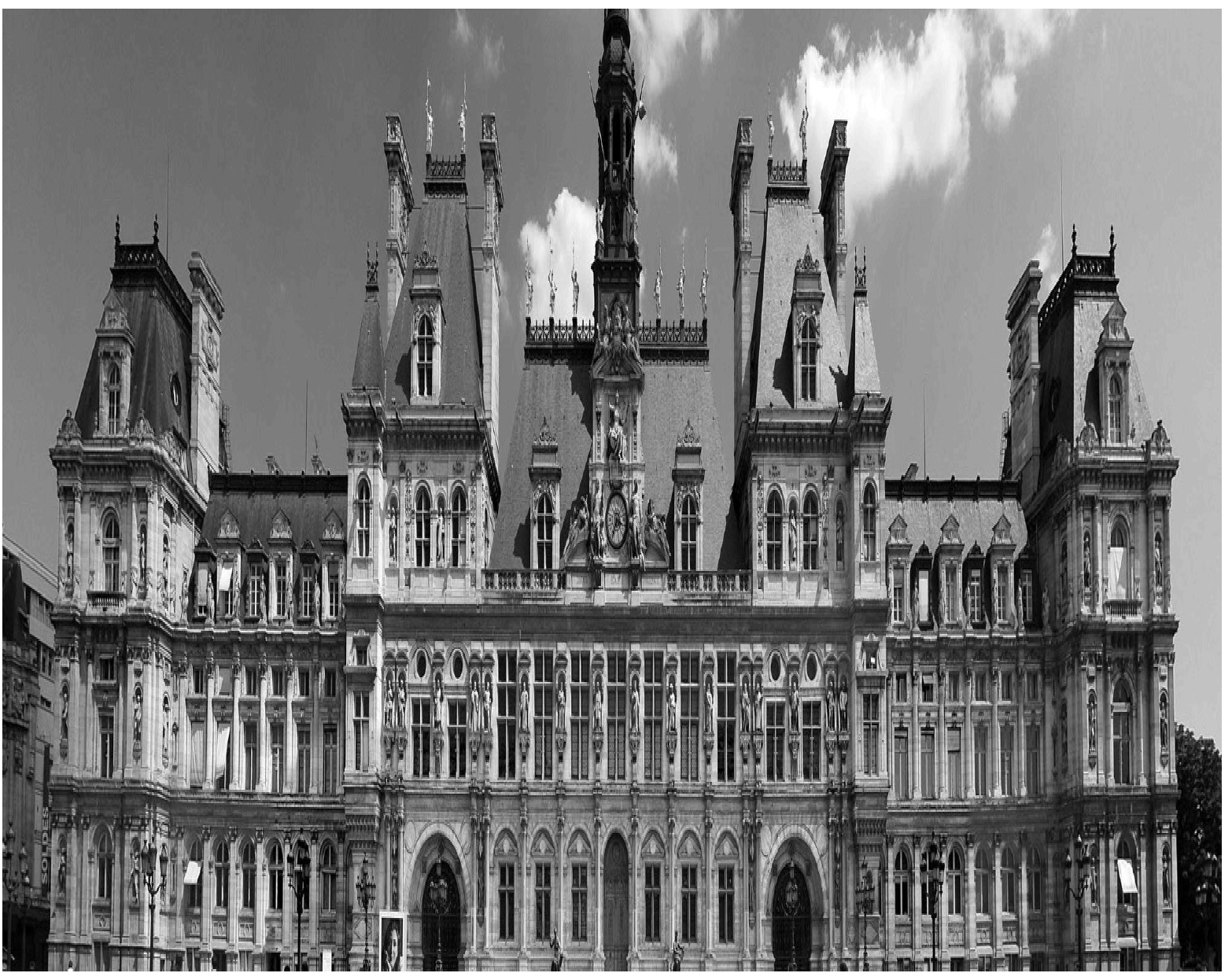} \label{fig:Paris}}
\caption{(a) San Francisco City Hall image. (b) Paris City Hall image.}
\label{fig:realdata}
\end{figure}

\begin{figure*}[t]
\centering
\subfigure[$\sigma_{tr}^2 = 0.02$]{\includegraphics[width=1.6in]{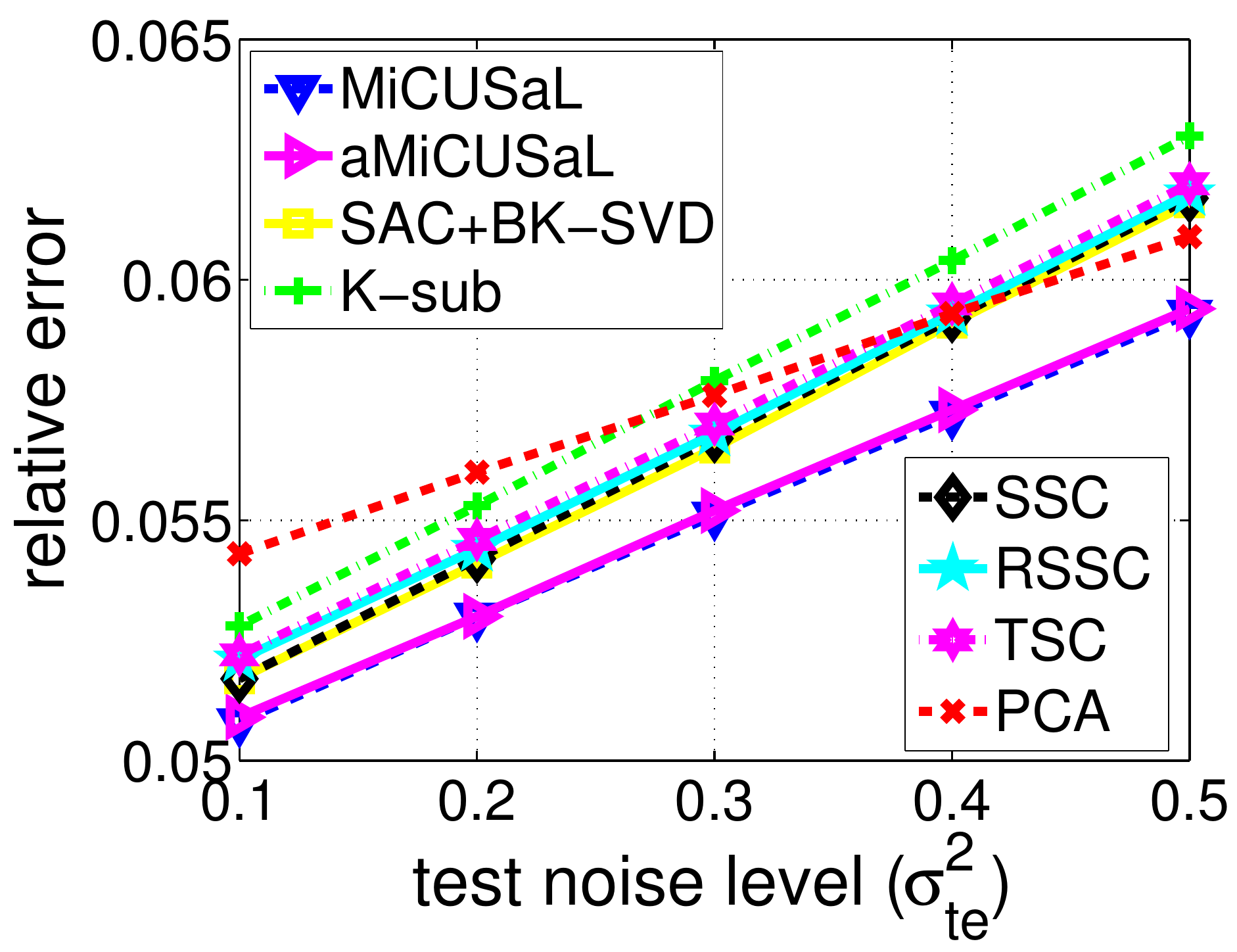} \label{fig:SanfranCompleteDenoise2}}
\quad
\subfigure[$\sigma_{tr}^2 = 0.02$]{\includegraphics[width=1.6in]{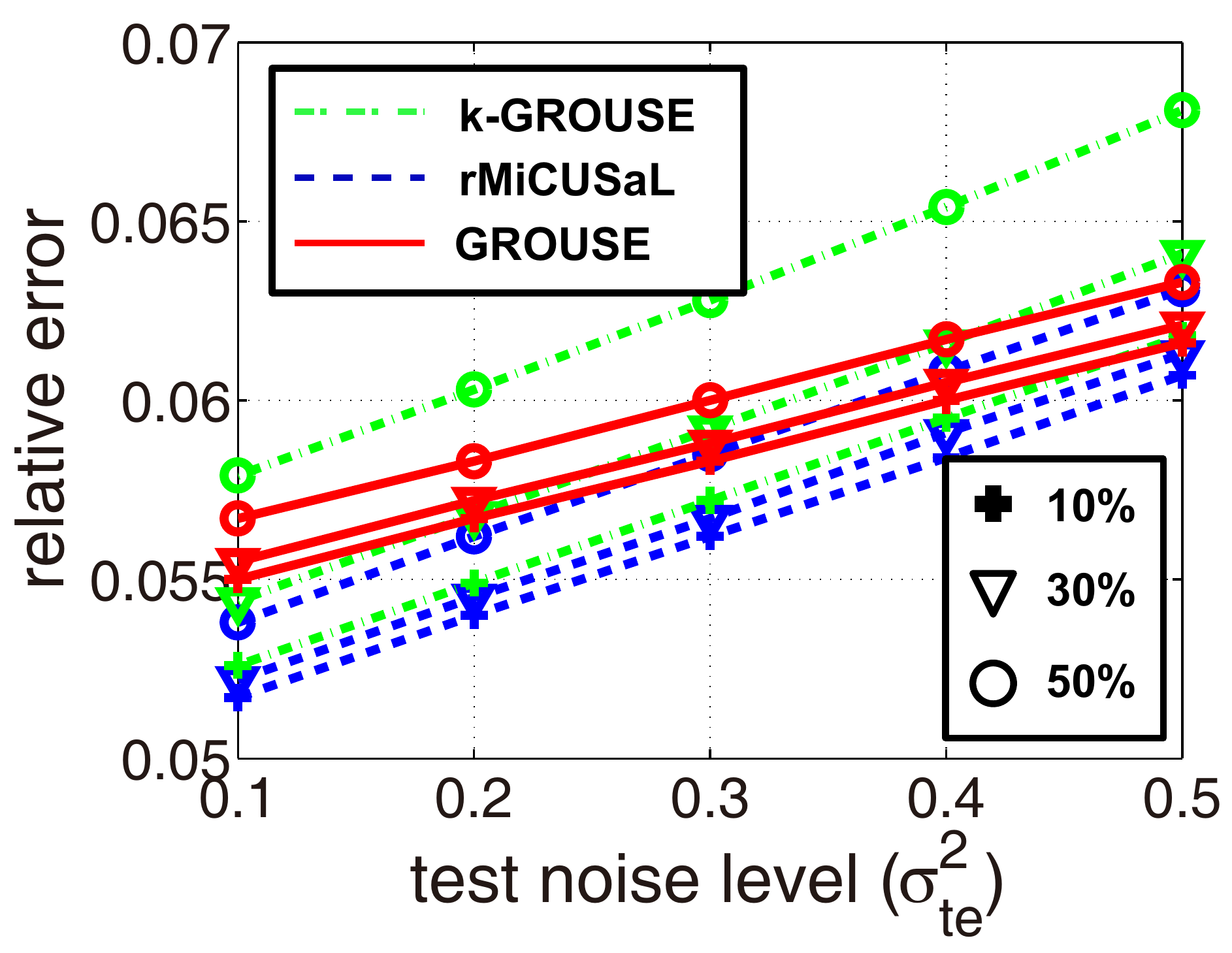} \label{fig:SanfranMissDenoise2}}
\quad
\subfigure[$\sigma_{tr}^2 = 0.02$]{\includegraphics[width=1.6in]{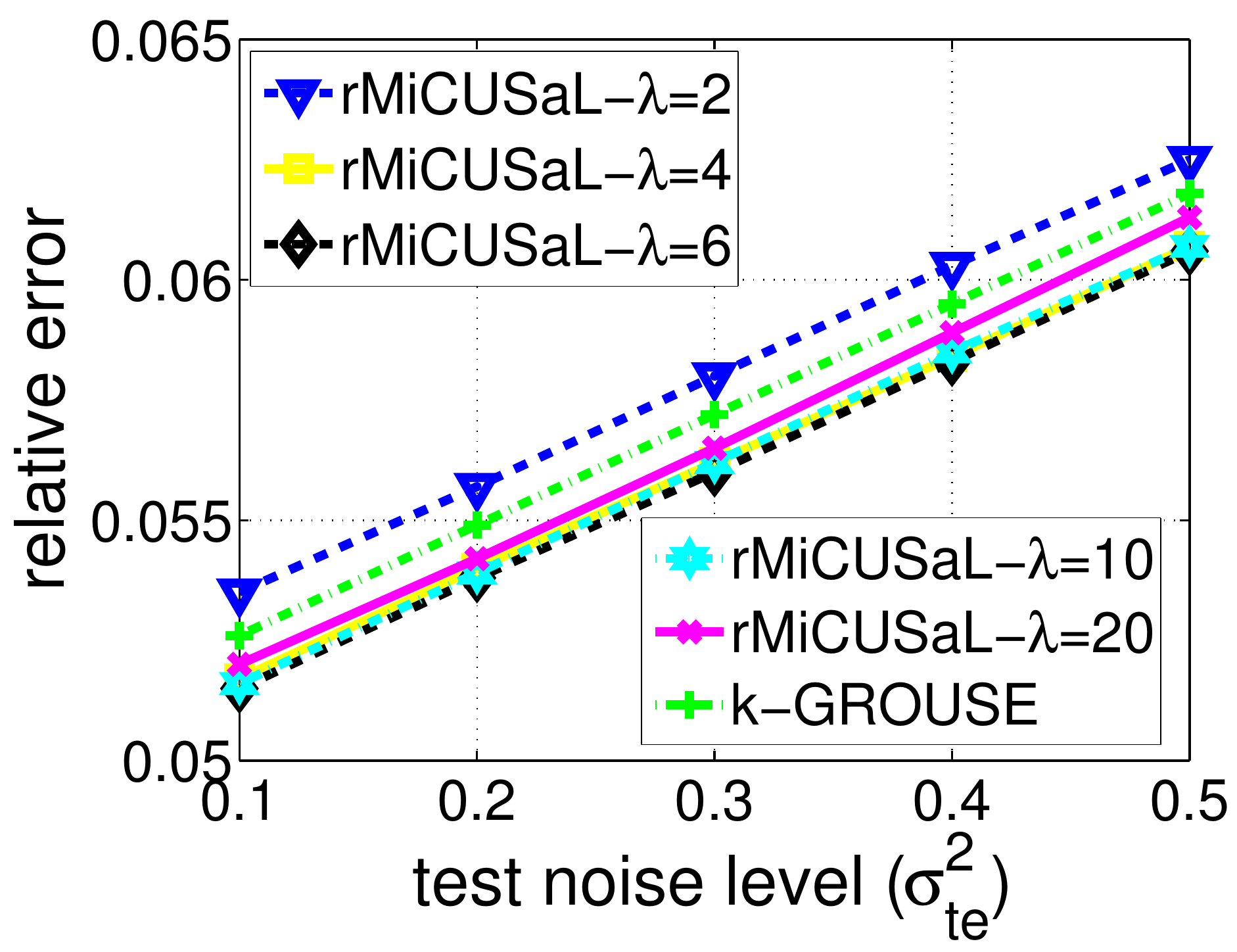} \label{fig:SanfranMissTrace2}} \\
\subfigure[$\sigma_{tr}^2 = 0.05$]{\includegraphics[width=1.6in]{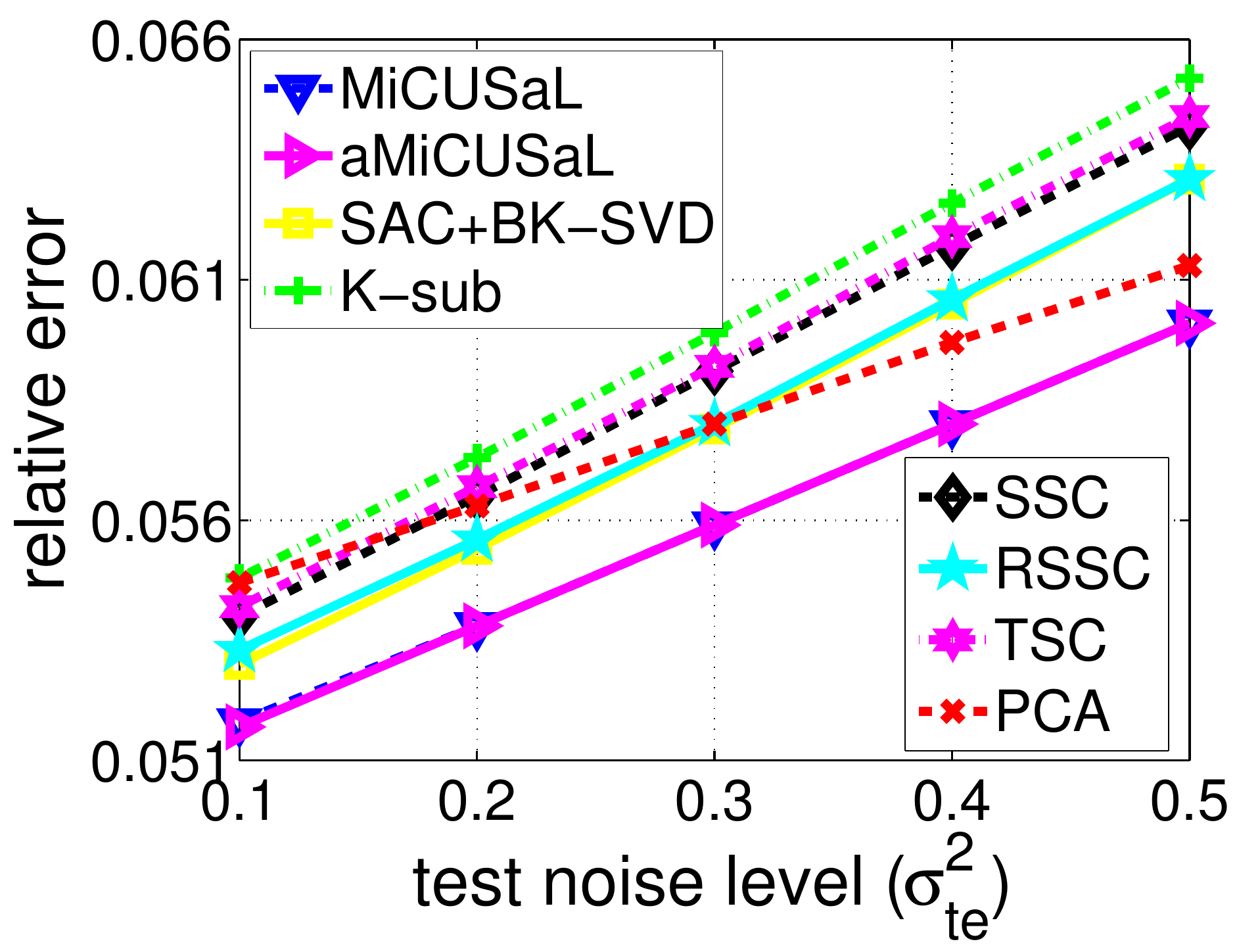} \label{fig:SanfranCompleteDenoise5}}
\quad
\subfigure[$\sigma_{tr}^2 = 0.05$]{\includegraphics[width=1.6in]{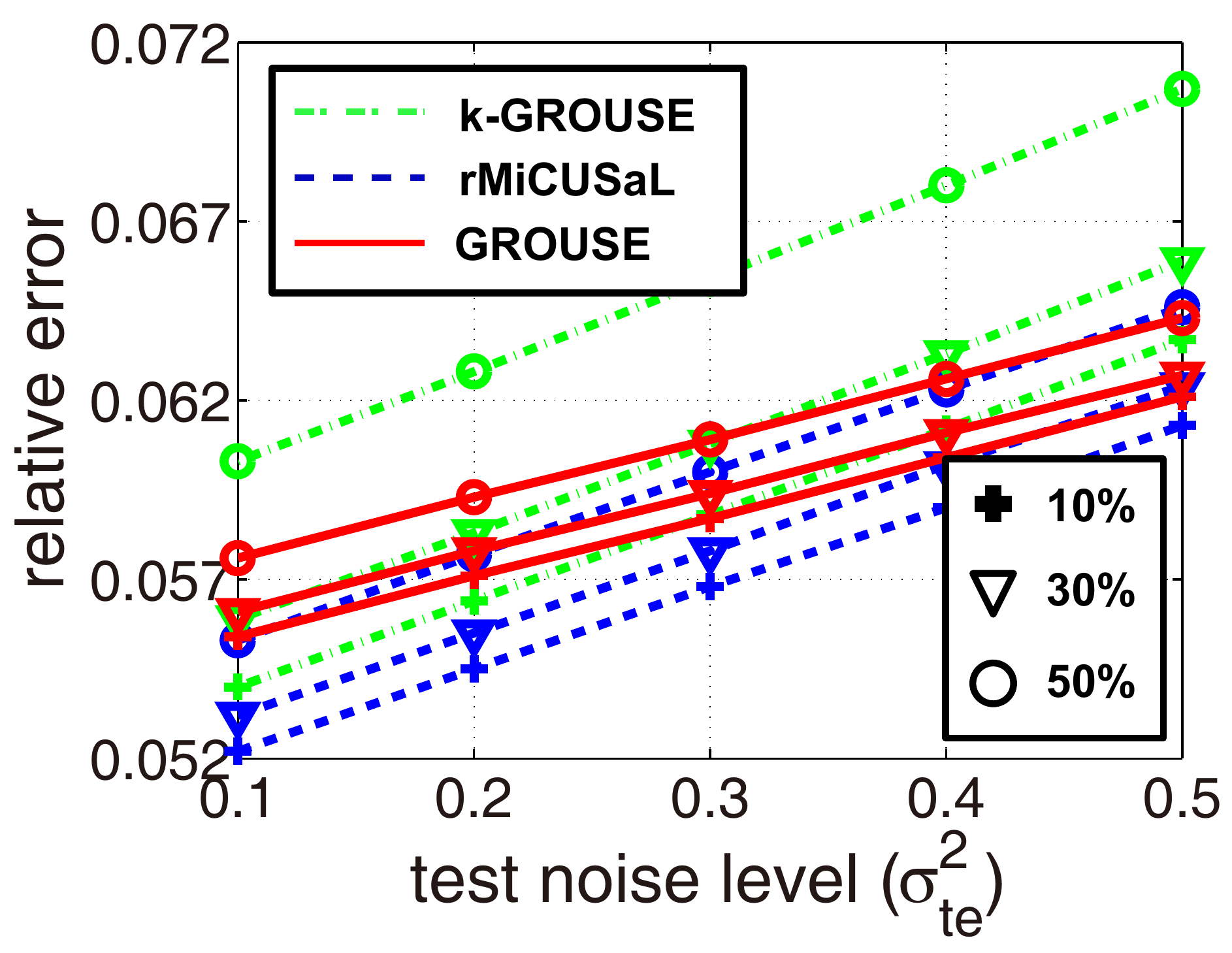} \label{fig:SanfranMissDenoise5}}
\quad
\subfigure[$\sigma_{tr}^2 = 0.05$]{\includegraphics[width=1.6in]{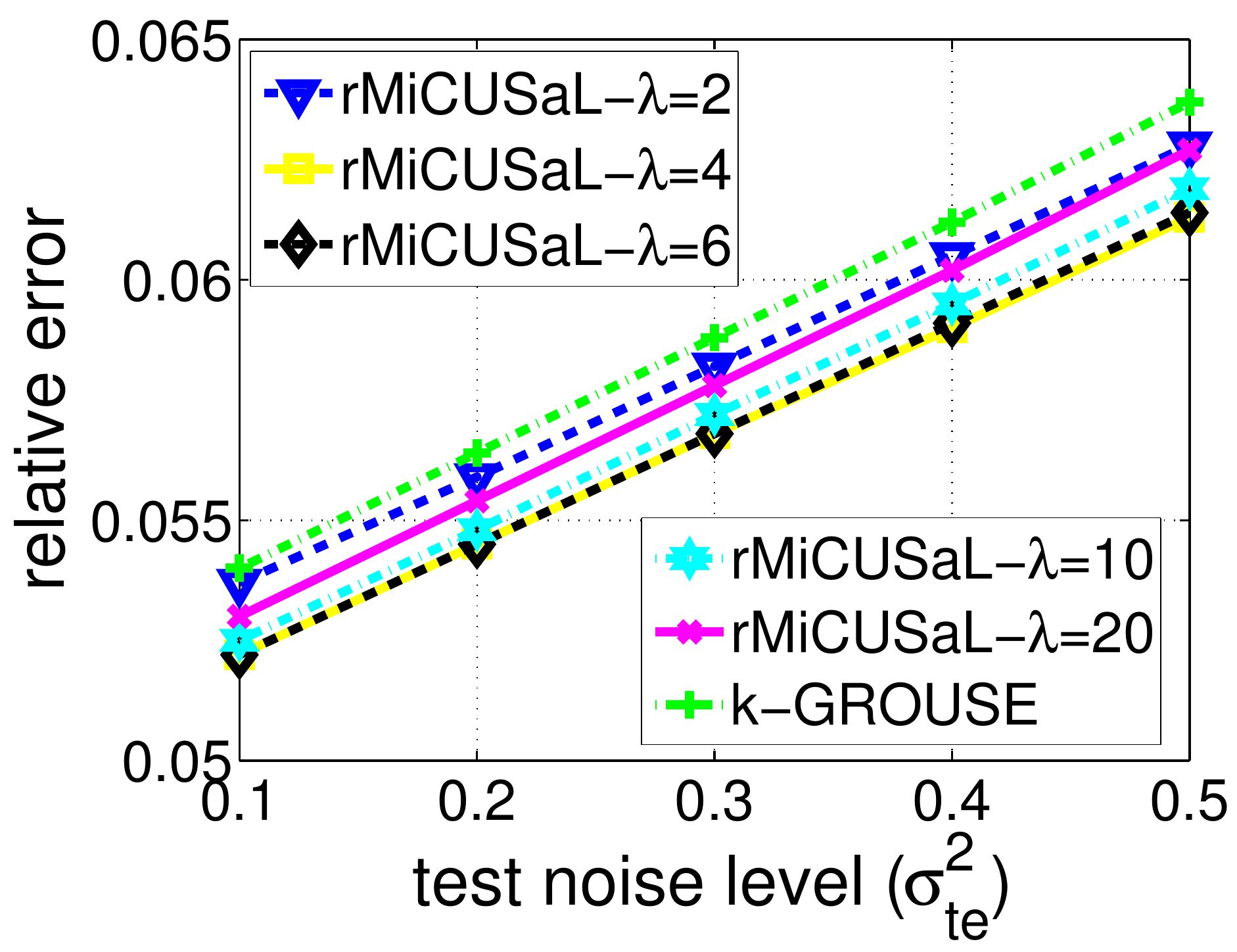} \label{fig:SanfranMissTrace5}}
\caption{Comparison of MC-UoS learning performance on San Francisco City Hall data. (a) and (d) show relative errors of test signals for complete data experiments. (b) and (e) show relative errors of test signals for missing data experiments. The numbers in the legend of (b) and (e) indicate the percentages of missing entries within the training data. (c) and (f) show relative errors of test signals for rMiCUSaL (with $10\%$ missing entries) using different $\lambda$'s.}
\label{fig:sanfranresult}
\end{figure*}

To further show the effectiveness of the proposed approaches, we test our
proposed methods on real-world city scene data. First, we study the
performance of our methods on San Francisco City Hall image, as shown in
Fig.~\ref{fig:Sanfran}. To generate the clean training and test data, we
split the image into left and right subimages of equal size. Then we extract
all $30 \times 20$ nonoverlapping image patches from the left subimage and
reshape them into $N=722$ column vectors of dimension $m=600$. All these
vectors are normalized to have unit $\ell_2$ norms and are then used as
signals in $\mathbf{X}$. Test signals in $\mathbf{X}^{te} \in \R^{600 \times
722}$ are extracted in the same way from the right subimage. White Gaussian
noise is then added to $\mathbf{X}$ and $\mathbf{X}^{te}$ separately, forming
$\mathbf{Y}$ and $\mathbf{Z}$, respectively. In these experiments,
$\sigma_{tr}^2$ is set to be $0.02$ and $0.05$, while $\sigma_{te}^2$ again
ranges from $0.1$ to $0.5$. The Monte Carlo simulations for noisy
data are repeated $50$ times and the results reported here correspond to the
average of these $50$ trials. Note that each patch is treated as a single
signal here, and our goal is to learn an MC-UoS from $\mathbf{Y}$ such that
every test patch can be reliably denoised using the learned subspaces.

We perform aMiCUSaL on the training data $\mathbf{Y}$ with parameters
$L_{max}=8$, $s_{max}=20$, $\lambda = 4$, $k_1=6$, $k_2=10$ and
$\epsilon_{min}=0.08$. The number of random initializations that are
used to arrive at the final MC-UoS structure using aMiCUSaL is $10$ for every
fixed $\mathbf{Y}$. The output $L$ from aMiCUSaL is $4$ or $5$ and $s$ is always between $11$ or
$13$. We also perform MiCUSaL with the same $L$ and $s$ $10$ times. For fair
comparison, we also use the method in this paper to get the dimension of the
subspace for PCA, in which case the estimated $s$ is always $10$. Note that
for all state-of-the-art UoS learning algorithms, we use the same $L$ and $s$
as aMiCUSaL instead of using the $L$ generated by the algorithms themselves.
The reason for this is as follows. The returned $L$ by SSC (with
$\alpha_z=40$) is $1$. Therefore SSC reduces to PCA in this setting. The
output $L$ for RSSC is also $4$ or $5$, which coincides with our algorithm.
The estimation of $L$ (with $q = 2 \max (3, \lceil N/(L \times 20) \rceil)$) for TSC is sensitive to the noise and data.
Specifically, the estimated $L$ is always from $6$ to $9$ for $\sigma_{tr}^2=0.02$ and
$L$ is always $1$ when $\sigma_{tr}^2=0.05$, which results in poorer
performance compared to the case when $L=4$ or $5$ for both training noise
levels. In the missing data experiments, we set $L=5$ and $s=12$ for rMiCUSaL
(with $\lambda = 4$) and $k$-GROUSE, and $s=10$ for GROUSE.
Fig.~\ref{fig:SanfranCompleteDenoise2} and
Fig.~\ref{fig:SanfranCompleteDenoise5} describe the relative reconstruction
errors of test samples when the training data are complete. We see both
MiCUSaL and aMiCUSaL learn a better MC-UoS since they give rise to smaller
relative errors of test data. Further, the average standard deviation
of the mean of relative errors for test data is around $0.00015$ for MiCUSaL and
$0.00045$ for $K$-sub. It can be inferred from
Fig.~\ref{fig:SanfranMissDenoise2} and Fig.~\ref{fig:SanfranMissDenoise5}
that rMiCUSaL also yields better data representation performance for the
missing data case.

To examine the effect of $\lambda$ on the denoising result in both complete
and missing data experiments, we first run aMiCUSaL with $\lambda \in
\{1,2,4,6,8,10\}$ without changing other parameters. When $\lambda=1$ or $2$,
aMiCUSaL always returns $L=2$ or $3$ subspaces, but the reconstruction errors
of the test data are slightly larger than those for $\lambda=4$. When
$\lambda \geq 6$, the distances between the learned subspaces become larger,
and the resulting $L$ will be at least $6$ when $\epsilon_{min}$ is fixed.
However, the relative errors of test data are still very close to the ones
for $\lambda=4$. This suggests that $\lambda=4$ is a good choice in this
setting since it leads to the smallest number of subspaces $L$ and the best
representation performance. We also perform rMiCUSaL with $\lambda \in
\{2,4,6,10,20\}$ while keeping $L$ and $s$ fixed, where the number of missing
entries in the training data is again $10\%$ of the signal dimension. We show
the relative errors of test data in Fig.~\ref{fig:SanfranMissTrace2} and
Fig.~\ref{fig:SanfranMissTrace5}. Similar to the results of the experiments
with synthetic data, we again observe the fact that when $\lambda$ is small
(e.g., $\lambda=2$), the reconstruction errors of the test data are large
because the subspace closeness metric dominates in learning the UoS. The
results for $\lambda=4$ and $6$ are very similar. As $\lambda$ increases
further, the performance of rMiCUSaL gets closer to that of $k$-GROUSE. We
again report the running time of rMiCUSaL and $k$-GROUSE per iteration in the
seventh row of Table~\ref{tab:timecomp}, where we perform the optimization over
each $\mathbf{D}_{\ell}$ for $100$ iterations in each subspace update step
for both rMiCUSaL and $k$-GROUSE. In these experiments, rMiCUSaL appears much
slower than $k$-GROUSE. However, as presented in
Fig.~\ref{fig:SanfranMissDenoise2} and Fig.~\ref{fig:SanfranMissDenoise5},
the performance of rMiCUSaL is significantly better than $k$-GROUSE.

\begin{figure}[t]
\centering
\subfigure[$\sigma_{tr}^2 = 0.02$]{\includegraphics[width=1.6in]{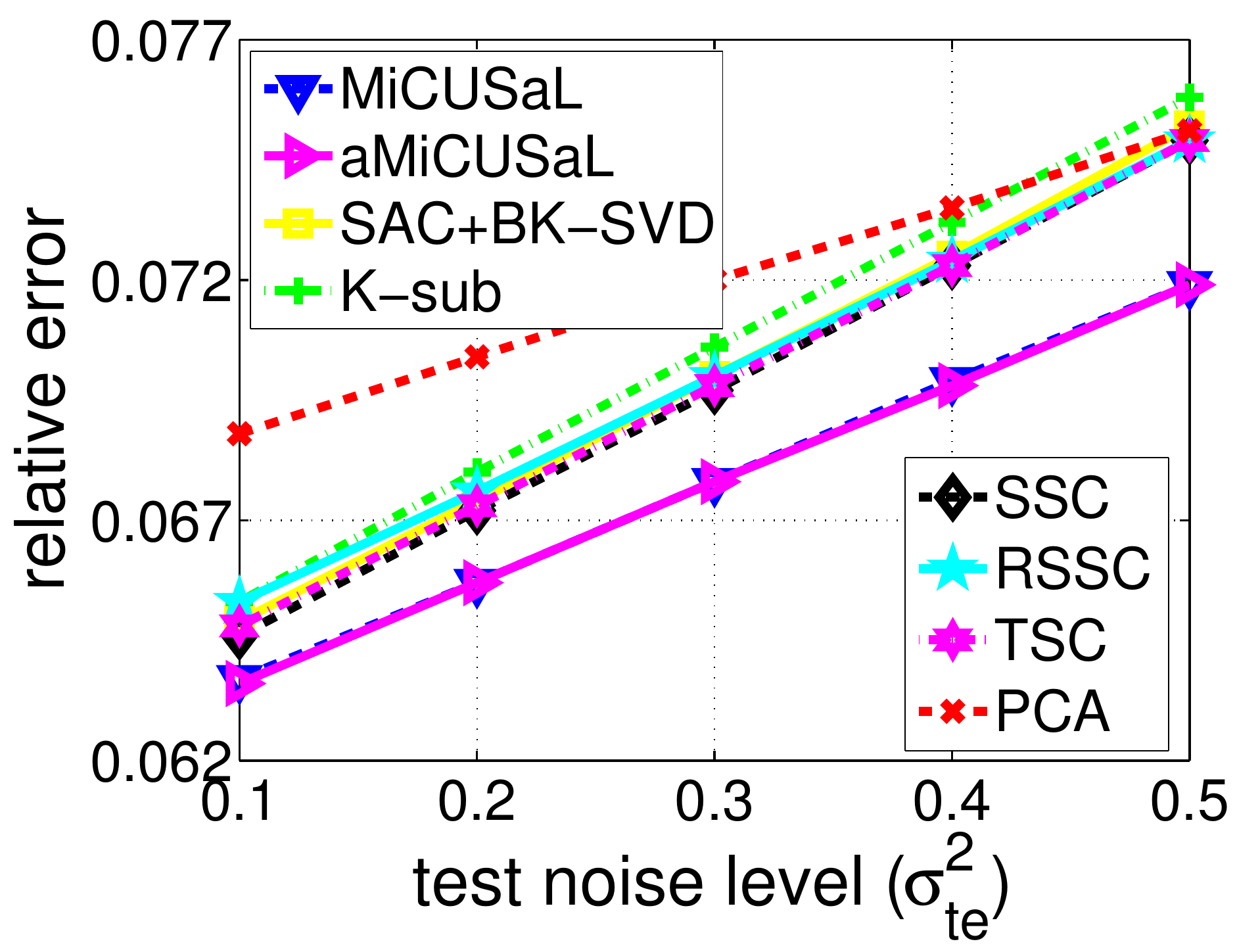} \label{fig:ParisCompleteDenoise2}}
\subfigure[$\sigma_{tr}^2 = 0.05$]{\includegraphics[width=1.6in]{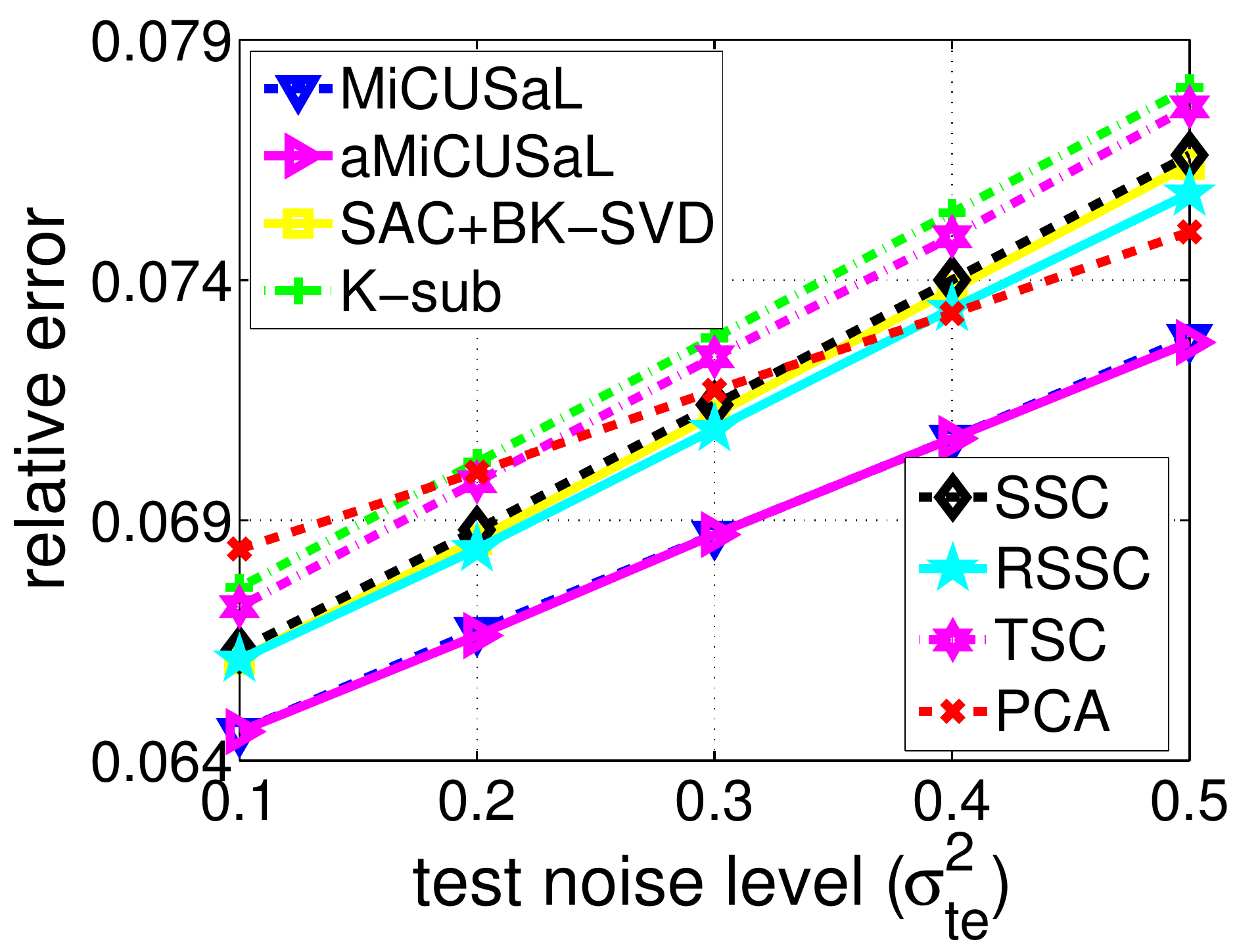} \label{fig:ParisCompleteDenoise5}}
\caption{Comparison of MC-UoS learning performance on Paris City Hall data when the training data are complete.}
\label{fig:parisresult}
\end{figure}

Next, we repeat these experiments for the complete data experiments using Paris
City Hall image in Fig.~\ref{fig:Paris}, forming $\mathbf{X}, \mathbf{X}^{te}
\in \R^{600 \times 950}$. We perform aMiCUSaL using the same parameters
($\lambda=4$) as in the previous experiments. The estimated $L$ in this case
is always between $5$ and $6$ and $s$ is always between $11$ and $12$. The
estimated dimension of the subspace in PCA is $9$ or $10$ when
$\sigma_{tr}^2=0.02$ and it is always $10$ when $\sigma_{tr}^2=0.05$. In these
experiments, we again use the same $L$ and $s$ as aMiCUSaL for all
state-of-the-art UoS learning algorithms. This is because the returned $L$ by
SSC (with $\alpha_z=20$) is again $1$ in this case. The estimated $L$ by RSSC
is usually $7$ or $8$, and the reconstruction errors of test data are very close to
the ones reported here. If we apply TSC using the $L$ estimated by itself
(again, with $q = 2 \max (3, \lceil N/(L \times 20) \rceil)$), we
will have $L=4$ when $\sigma_{tr}^2=0.02$, while the relative
errors of test data are very close to the results shown here. For
$\sigma_{tr}^2=0.05$, TSC will again result in only one subspace. The
relative reconstruction errors of test data with different training noise
levels are shown in Fig.~\ref{fig:parisresult}, from which we make the
conclusion that our methods obtain small errors, thereby outperforming all
other algorithms. The average standard deviation of the mean of
relative errors for test data is also smaller for MiCUSaL (around $0.00023$)
compared to $K$-sub (around $0.00037$).

\subsubsection{Experiments on Face Dataset}
\label{sssec:realface}

\begin{figure}[t]
\centering
\subfigure[]{\includegraphics[width=1.6in]{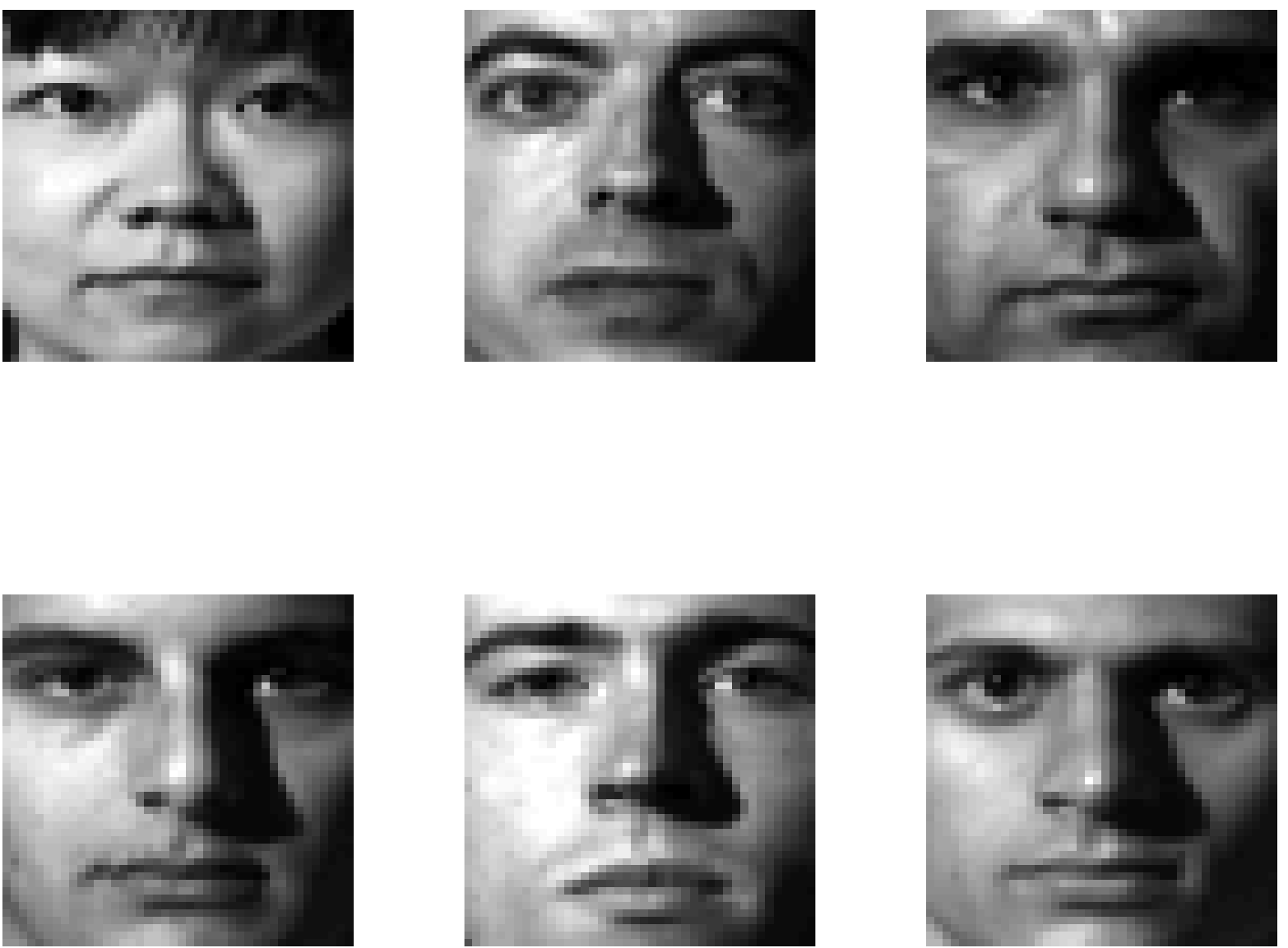} \label{fig:YaleBData}} \\
\subfigure[]{\includegraphics[width=1.6in]{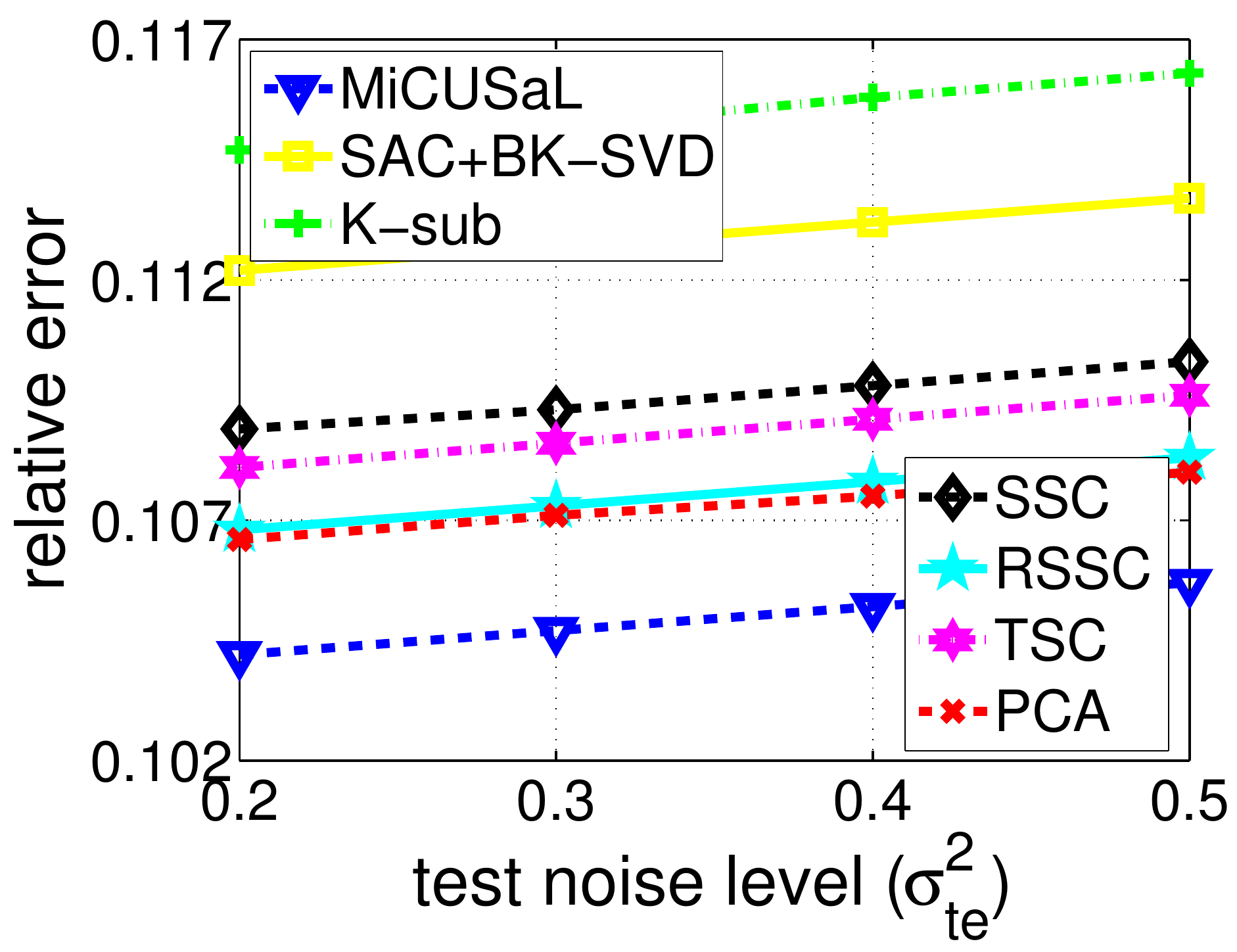} \label{fig:YaleBCompleteData1}}
\subfigure[]{\includegraphics[width=1.6in]{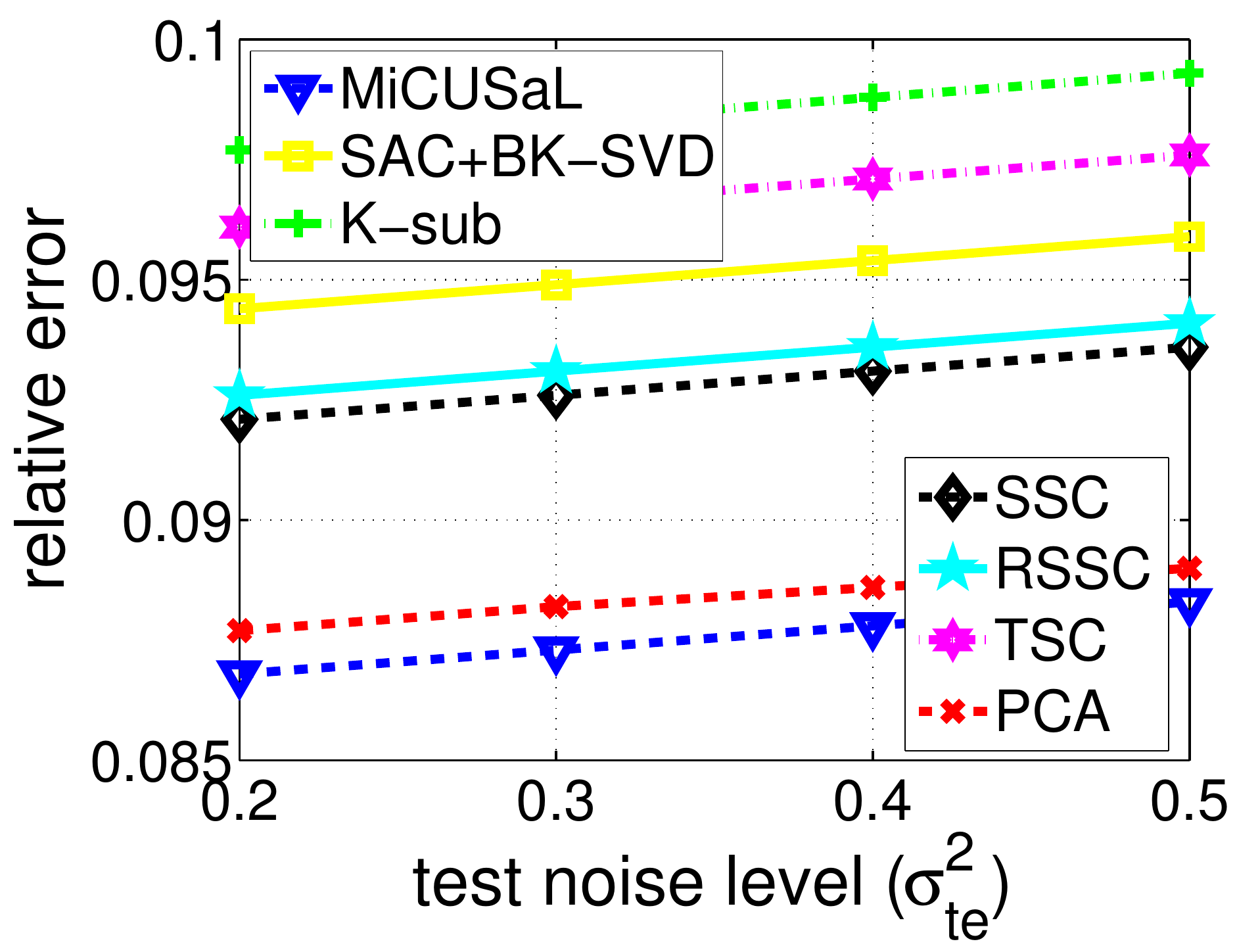} \label{fig:YaleBCompleteData2}}
\caption{Comparison of MC-UoS learning performance on Extended Yale B dataset. The first row of (a) shows some images of subject $5, 6, 8$ and the second row presents some images of subject $22, 28, 30$. (b) and (c) show relative errors of test data in the two experiments.}
\label{fig:yaleB}
\end{figure}

In this section, we work with the Extended Yale B dataset \cite{LeeHK.PAMI2005}, which contains a set of $192 \times 168$ cropped images of $38$ subjects. For each individual, there are $64$ images taken under varying illumination conditions. We downsample the images to $48 \times 42$ pixels and each image is vectorized and treated as a signal; thus, $m=2016$. It has been shown in \cite{BasriJ.PAMI2003} that the set of images of a given subject taken under varying illumination conditions can be well represented by a $9$-dimensional subspace.

We first focus on a collection of images of subjects $5, 6$ and $8$ and normalize
all the images to have unit $\ell_2$ norms. Some representative images are presented in the
first row of Fig.~\ref{fig:YaleBData}. Here we assume the images of
these three subjects lie close to an MC-UoS with $L=3$ and $s=9$. For each
set of images from one subject, we randomly select half of them for training
and the remaining $32$ images belong to test samples; therefore, $\mathbf{X},
\mathbf{X}^{te} \in \R^{2016 \times 96}$. Then we add white Gaussian noise to
both $\mathbf{X}$ and $\mathbf{X}^{te}$ and obtain $\mathbf{Y}$ and $\mathbf{Z}$. The random selection for generating
$\mathbf{X}$ and $\mathbf{X}^{te}$ is repeated $10$ times and we conduct
Monte Carlo simulations for noisy data $10$ times for every fixed
$\mathbf{X}$ and $\mathbf{X}^{te}$. In these experiments, the value
$\sigma_{tr}^2$ is equal to $0.2$ and $\sigma_{te}^2$ is from $0.2$ to $0.5$.
For fair comparison, the dimension of the subspace for PCA is set to be $9$.
We apply MiCUSaL with parameter $\lambda=2$ and SSC with $\alpha_z=40$.
The number of random initializations in these experiments for both
MiCUSaL and $K$-sub is set at $8$ for every fixed $\mathbf{Y}$. Once again,
we observe that MiCUSaL outperforms other learning methods, since it results
in smaller relative errors (cf.~Fig.~\ref{fig:YaleBCompleteData1}).
Moreover, the average standard deviation of MiCUSaL for the $100$
realizations of $\mathbf{Y}$ and $\mathbf{Z}$ is only $0.0013$ for all
$\sigma_{te}$'s, which is again smaller than that of $K$-sub (the
corresponding value is $0.002$).

We then repeat these experiments using a set of images of subjects $22, 28$ and $30$,
and show some image samples in the second row of Fig.~\ref{fig:YaleBData}. We
set $L=3$, $s=9$ and $\lambda=2$ for MiCUSaL and $\alpha_z=40$ for SSC. We
again provide evidence in Fig.~\ref{fig:YaleBCompleteData2} that MiCUSaL
yields better data representation performance in this setting. The average
standard deviation of the mean of the reconstruction errors for test data is
around $0.0012$ for MiCUSaL and $0.0019$ for $K$-sub in this case.

\subsection{Experiments for MC-KUoS Learning}
\label{ssec:nonlinearexperiment}

In this section, we evaluate the performance of the MC-KUoS learning
approaches in terms of the following two problems: image denoising using the
learned MC-KUoS and clustering of training data points. For both these
problems, we consider the USPS dataset
\cite{Hull.PAMI1994},\footnote{Available at:
\url{http://www.cs.nyu.edu/~roweis/data.html}.} which contains a collection of
$m=256$-dimensional handwritten digits. The authors in
\cite{MikaSSMSR.NIPS1998} have demonstrated that using nonlinear features can
improve the denoising performance of this dataset. Unlike the experiments for
MC-UoS learning, the training data we use in this set of experiments are
noiseless. For denoising experiments, we assume every noisy \emph{test}
sample $\mathbf{z} = \mathbf{x} + \boldsymbol{\xi}$, where
$\widetilde{\phi}(\mathbf{x}) = \phi(\mathbf{x}) -
\overline{\boldsymbol{\phi}}$ belongs to one of the $\cS_{\ell}$'s in $\cF$
(again $\|\mathbf{x}\|_2^2=1$) and $\boldsymbol{\xi}$ has $\mathcal{N}(\boldsymbol{0},
(\sigma_{te}^2/m)\mathbf{I}_m)$ distribution. In these experiments, $\sigma_{te}^2$ ranges from $0.2$ to $0.5$.

\subsubsection{Experiments on Image Denoising}
\label{sssec:nonlineardenoise}

For denoising experiments, we compare the result of MC-KUSaL with three other methods: ($i$) kernel $k$-means clustering (kernel $k$-means) \cite{ScholkopfSM.NC1998}, where for each test signal $\mathbf{z}$, we first assign $\phi(\mathbf{z})$ to a cluster whose centroid is closest to $\phi(\mathbf{z})$ in $\cF$, followed by kernel PCA and the method in \cite{RathiDT.SPIE2006} to calculate the pre-image; ($ii$) kernel PCA \cite{ScholkopfSM.Advance1999} with the same number of eigenvectors as in MC-KUSaL (KPCA-Fix); and ($iii$) kernel PCA with the number of eigenvectors chosen by $s = \argmin_{s} || P_{\cS} \phi(\mathbf{z}) - \phi(\mathbf{x})||_2^2$ (KPCA-Oracle), where $\mathbf{x}$ and $\mathbf{z}$ are clean and noisy test samples, respectively. In this manner, the number of eigenvectors $s$ for KPCA-Oracle will be different for different noise levels $\sigma_{te}$'s. We use the same dimension of the subspaces for MC-KUSaL, kernel $k$-means clustering and KPCA-Fix, while the number of subspaces $L$ for kernel $k$-means clustering also equals the one for MC-KUSaL. For the case of missing training data, we report the results of rMC-KUSaL as well as rKPCA. For every fixed test noise level $\sigma_{te}$, we set the dimension of the subspace $s$ for rKPCA to be the same as the one for KPCA-Oracle. The relative reconstruction error of a clean test signal $\mathbf{x} \in \mathbf{X}^{te}$ is calculated by $\frac{\| \mathbf{x} - \widehat{\mathbf{z}} \|_2^2}{\| \mathbf{x} \|_2^2}$, where $\widehat{\mathbf{z}}$ denotes the pre-image with respect to the noisy test sample $\mathbf{z}$.

\begin{figure}[t]
\centering
\subfigure[Complete data]{\includegraphics[width=1.6in]{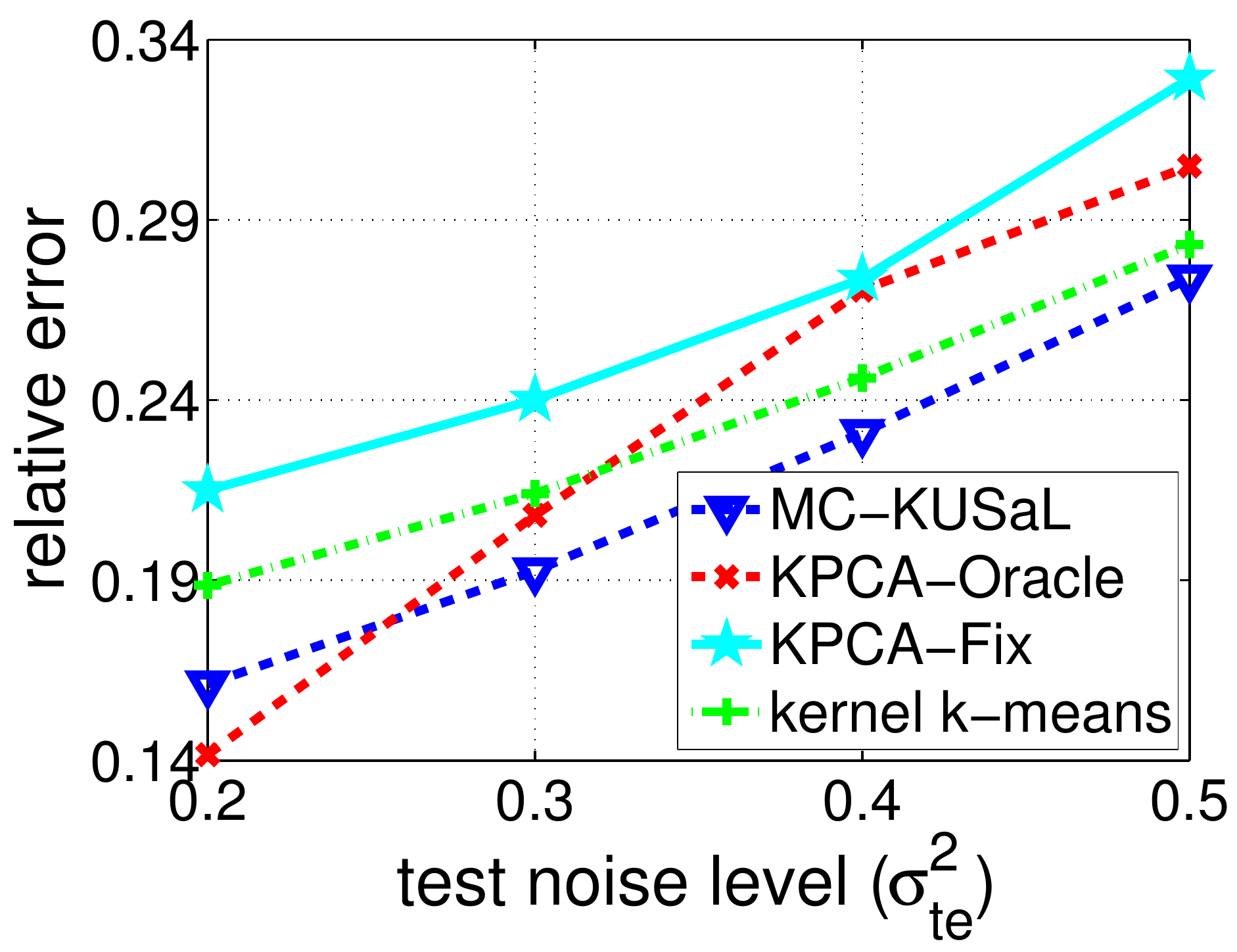} \label{fig:USPSGaussComplete}}
\subfigure[Complete data]{\includegraphics[width=1.6in]{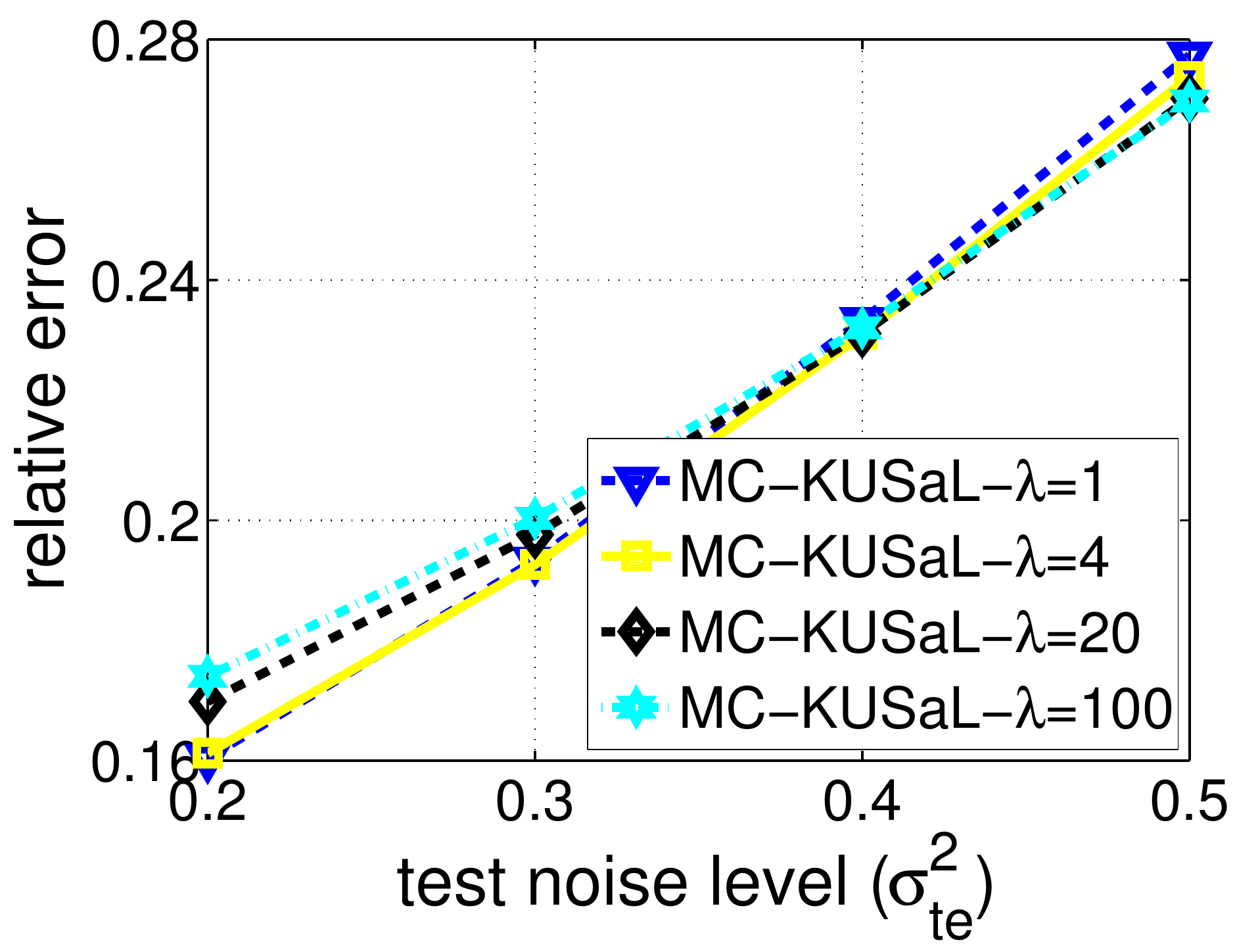} \label{fig:USPSGaussCompleteTrace}} \\
\subfigure[Missing data]{\includegraphics[width=1.6in]{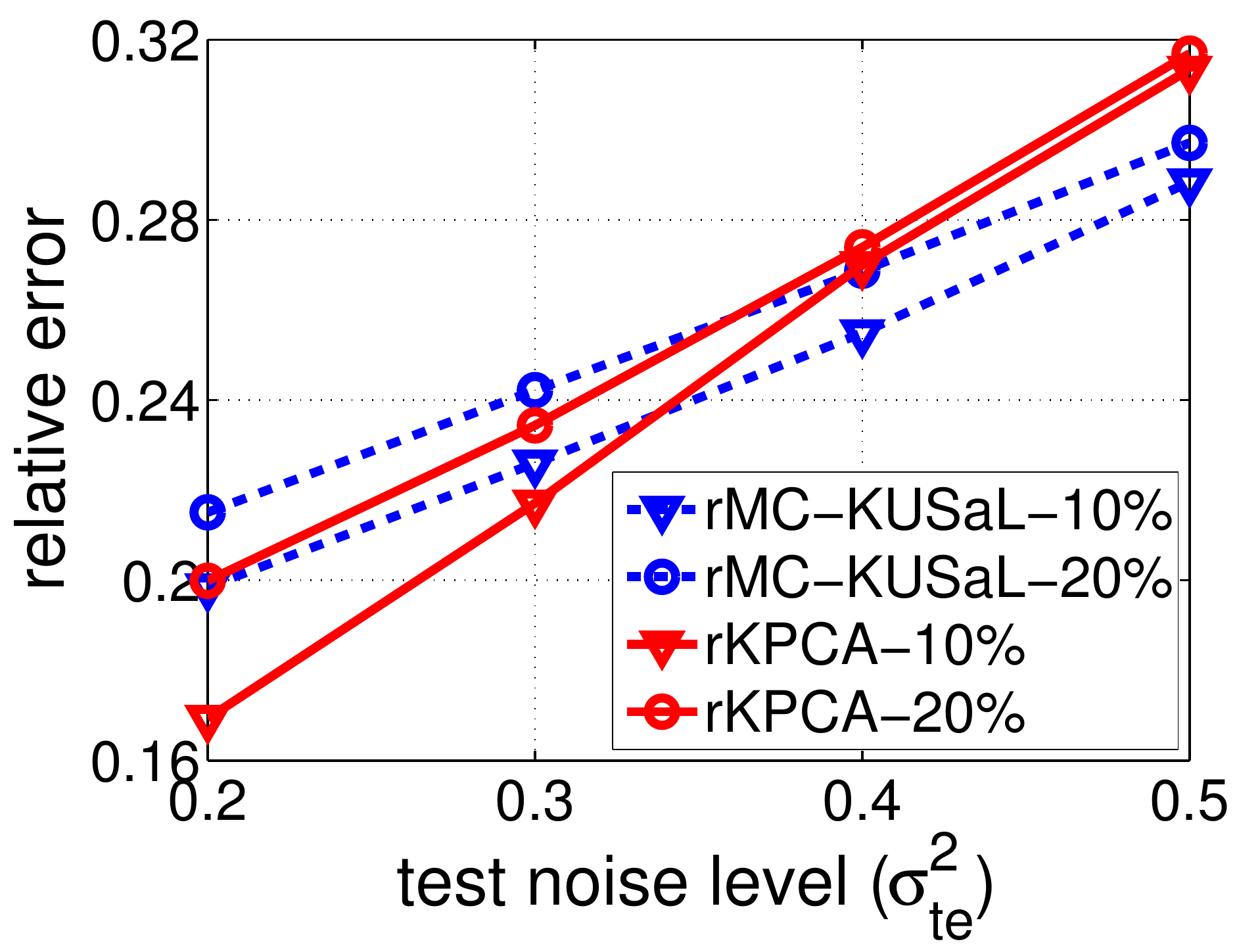} \label{fig:USPSGaussMiss}}
\caption{Comparison of MC-KUoS learning performance on USPS dataset using Gaussian kernel $\kappa(\mathbf{y},\mathbf{y'}) = \exp( - \frac{ \|\mathbf{y}-\mathbf{y'}\|_2^2}{4} )$. In (a), we perform MC-KUSaL with $\lambda=4$. Note that the KPCA-Oracle algorithm is the ideal case of kernel PCA. The numbers in the legend of (c) indicate the percentages of missing entries within the training data.}
\label{fig:usps}
\end{figure}

We experiment with Gaussian kernel with parameter $c=4$. We choose the digits ``$0$'' and ``$4$'' and for each digit we select the first $200$ samples in the dataset ($400$ images in total) for our experiments. All these $400$ samples are then vectorized and normalized to unit $\ell_2$ norms. From these samples, we randomly choose $120$ samples (without replacement) from each class for training and the remaining $80$ samples of each class for testing, forming $\mathbf{X} \in \R^{256 \times 240}$ and $\mathbf{X}^{te} \in \R^{256 \times 160}$. This random selection of test and training samples is repeated $20$ times for cross-validation purposes. We perform $10$ Monte Carlo trials for noisy test data and report the mean over these $200$ random trials.

In these experiments, we implement MC-KUSaL with parameters $L=2$, $s=45$ and $\lambda \in \{1,4,20,100\}$ to learn an MC-UoS in the feature space $\cF$. Fig.~\ref{fig:USPSGaussComplete} shows the mean of relative reconstruction errors of test data for different methods in the presence of complete training data, where we use the result of MC-KUSaL with $\lambda=4$ for comparison with other methods. We observe that for almost all noise levels, our method produces better results than other methods. The only exception is when $\sigma_{te}^2 = 0.2$, in which case MC-KUSaL is the second best of all methods. The caveat here is that in practice, we cannot know beforehand the dimension of the subspace in the feature space for kernel PCA, which yields the best denoising result at this particular noise level. We show the denoising performance of MC-KUSaL with different $\lambda$'s in Fig.~\ref{fig:USPSGaussCompleteTrace}, and we observe that a small $\lambda$ usually results in good performance when $\sigma_{te}^2$ is relatively small, while increasing the $\lambda$ will slightly improve the denoising performance when the SNR of test data gets small.

In the missing data experiments, we set the number of missing entries in the training data to be $10\%$ and $20\%$ of the signal dimension. We use parameters $L=2$, $s=45$ and $\lambda=4$ for rMC-KUSaL. It can be inferred from Fig.~\ref{fig:USPSGaussMiss} that ($i$) the performance of rKPCA and rMC-KUSaL is comparable for all noise levels; and ($ii$) when the number of missing elements is fixed, rMC-KUSaL outperforms the rKPCA when the SNR of the test data is small and vice versa.

\subsubsection{Experiments on Clustering}
\label{sssec:nonlinearcluster}

\begin{table*}[t]
\centering \caption{Clustering error ($\%$) on the USPS dataset}
\begin{tabular}{c|c|c|c|c|c||c|c}
\hline
\multirow{2}{*}{Digits} & \multirow{2}{*}{Kernel Function} & \multicolumn{6}{c}{Algorithms} \\
\cline{3-8}
  &  & MC-KUSaL & kernel $k$-means & $k$-means & SSC & rMC-KUSaL($10\%$) & rMC-KUSaL($20\%$)  \\
\hline
\multirow{2}{*}{1,7} & $\kappa(\mathbf{y},\mathbf{y'}) = \exp( - \frac{\|\mathbf{y}-\mathbf{y'}\|_2^2}{8} )$  & \textbf{7.21} & 20.94 & \multirow{2}{*}{21.19} & \multirow{2}{*}{12.13} & 11.52 & 12.69  \\
\cline{2-4}  \cline{7-8}
 & $\kappa(\mathbf{y},\mathbf{y'}) = (\langle \mathbf{y}, \mathbf{y'} \rangle + 2)^3$  & \textbf{8.23} & 20.54 & & & 10.60 & 11.85  \\
\hline
\multirow{2}{*}{1,6} & $\kappa(\mathbf{y},\mathbf{y'}) = \exp( - \frac{\|\mathbf{y}-\mathbf{y'}\|_2^2}{4} )$  & \textbf{5.00} & 11.04 & \multirow{2}{*}{11.29} & \multirow{2}{*}{8.71} & 6.27 & 7.88  \\
\cline{2-4}  \cline{7-8}
 & $\kappa(\mathbf{y},\mathbf{y'}) = (\langle \mathbf{y}, \mathbf{y'} \rangle + 1)^3$  & \textbf{4.85} & 10.60 & & & 7.54 & 8.04  \\
\hline
\end{tabular}
\label{tab:clusterresult}
\end{table*}

In this section, we empirically compare the clustering performance of
MC-KUSaL with ($i$) kernel $k$-means clustering (kernel $k$-means)
\cite{ScholkopfSM.NC1998}, ($ii$) standard $k$-means clustering ($k$-means)
\cite{Lloyd.TIT1982}, and ($iii$) spectral clustering \cite{Luxburg.SaC2007}
when the training data are complete. In the case of spectral clustering, we
make use of the similarity matrix returned by the noisy variant of the SSC
optimization program in \cite{ElhamifarV.PAMI2013}. We also present the
clustering performance of rMC-KUSaL, with the number of missing entries in
the training data being set to $10\%$ and $20\%$ of the signal dimension.
We compute the clustering error for MC-KUSaL/rMC-KUSaL by using the final
kernel subspace assignment labels $\{l_i\}_{i=1}^N$. For all the following
experiments, we select $L=2$ (since we only have $2$ classes) and
$\lambda=200$.

We first experiment with digits ``$1$'' and ``$7$'' in the USPS dataset, where in
every trial we randomly choose $120$ $\ell_2$ normalized samples from the
first $200$ samples of these two digits and use these $240$ samples
in these experiments. This random selection is repeated $20$ times. We perform
MC-KUSaL and rMC-KUSaL using Gaussian kernel $\kappa(\mathbf{y},\mathbf{y'}) = \exp( - \frac{
\|\mathbf{y}-\mathbf{y'}\|_2^2}{8} )$ with $s=35$ and polynomial kernel
$\kappa(\mathbf{y},\mathbf{y'}) = (\langle \mathbf{y}, \mathbf{y'} \rangle +
2)^3$ with $s=40$. The parameter $\alpha_z$ for SSC is set to be $20$. The
clustering results are listed in Table~\ref{tab:clusterresult}, where we can
see the clustering error for MC-KUSaL is roughly $40\%$ of the ones for
kernel/standard $k$-means clustering and MC-KUSaL is much better than SSC
(with $32\%$ reduction) in these experiments. In addition, the clustering error
for rMC-KUSaL is an increasing function of the number of missing entries for
both Gaussian kernel and polynomial kernel.

As another example, we repeat the above experiments using digits ``$1$'' and
``$6$'', where we again apply MC-KUSaL and rMC-KUSaL using Gaussian kernel
$\kappa(\mathbf{y},\mathbf{y'}) = \exp( - \frac{\|\mathbf{y}-\mathbf{y'}\|_2^2}{4} )$ with $s=35$ and polynomial kernel
$\kappa(\mathbf{y},\mathbf{y'}) = (\langle \mathbf{y}, \mathbf{y'} \rangle +
1)^3$ with $s=40$. In these experiments, SSC is performed with $\alpha_z=10$.
From Table~\ref{tab:clusterresult}, we again observe that MC-KUSaL
outperforms other clustering algorithms with $42\%$ reduction (compared to
SSC) in the clustering error. In the missing data experiments, the clustering
performance of rMC-KUSaL using Gaussian kernel degrades as the number of
missing entries of the data increases. When we use polynomial kernel for
rMC-KUSaL, increasing the number of entries in the missing data does not
result in much degradation of the clustering performance.

We conclude by noting that the choice of kernels in these experiments
is agnostic to the training data. Nonetheless, data-driven learning of
kernels is an active area of research, which is sometimes studied under the
rubric of \emph{multiple kernel learning}
\cite{BachLJ.ICML2004,LanckrietCBGJ.JMLR2004,CortesMR.NIPS2009,KloftBSZ.JMLR2011}.
While some of these works can be leveraged to further improve the performance
of our proposed algorithms, a careful investigation of this is beyond the
scope of this work.

\section{Conclusion}
\label{sec:conclusion}

In this paper, we proposed a novel extension of the canonical union-of-subspaces model, termed the metric-constrained union-of-subspaces (MC-UoS) model. We first proposed several efficient iterative approaches for learning of an MC-UoS in the ambient space using both complete and missing data. Moreover, these methods are extended to the case of a higher-dimensional feature space such that one can deal with MC-UoS learning problem in the feature space using complete and missing data. Experiments on both synthetic and real data showed the effectiveness of our algorithms and their superiority over the state-of-the-art union-of-subspaces learning algorithms. Our future work includes estimation of the number and dimension of the subspaces from the training data for MC-UoS learning in the feature space.

\appendix[Proof of Lemma \ref{lemma:innerproduct}]  \label{append:proof}
\begin{IEEEproof}
First, we have $\langle \mathbf{y}_i, \mathbf{y}_j \rangle = \sum_{u=1}^{m} {\mathbf{z}_{ij}^{*}}_{(u)}$ and $\langle [\mathbf{y}_i]_{\mathbf{\Omega}_{ij}}, [\mathbf{y}_j]_{\mathbf{\Omega}_{ij}} \rangle = \sum_{v=1}^{n} {\mathbf{z}_{ij}^{*}}_{({\mathbf{\Omega}_{ij}}_{(v)})}$ with $n = |\mathbf{\Omega}_{ij}|$. Here, ${\mathbf{z}_{ij}^{*}}_{(u)}$ denotes the $u$-th entry of a vector $\mathbf{z}_{ij}^{*}$ and ${\mathbf{\Omega}_{ij}}_{(v)}$ denotes the $v$-th element of $\mathbf{\Omega}_{ij}$. Let $\hbar(Z_1, \dots, Z_n) = \sum_{v=1}^{n} Z_v$ be the sum of $n$ random variables and $Z_v = {\mathbf{z}_{ij}^{*}}_{({\mathbf{\Omega}_{ij}}_{(v)})}$. We prove the bound under the assumption that these $n$ variables are drawn uniformly from a set $\{ {\mathbf{z}_{ij}^{*}}_{(1)}, \dots, {\mathbf{z}_{ij}^{*}}_{(m)}\}$ with replacement. This means they are independent and we have $ \E [\sum_{v=1}^{n} Z_v] = \E [ \sum_{v=1}^{n} {\mathbf{z}_{ij}^{*}}_{({\mathbf{\Omega}_{ij}}_{(v)})} ] = \frac{n}{m} \sum_{u=1}^{m} {\mathbf{z}_{ij}^{*}}_{(u)} $. If the value of one variable in the sum is replaced by any other of its possible values, the sum changes at most $2\|\mathbf{z}_{ij}^{*}\|_{\infty}$, i.e., $ | \sum_{v=1}^{n} Z_v - \sum_{v \neq v'} Z_v - \widehat{Z}_{v'} | = | Z_{v'} - \widehat{Z}_{v'} | \leq 2\|\mathbf{z}_{ij}^{*}\|_{\infty} $ for any $v' \in \{ 1,\dots, n \}$. Therefore, McDiarmid's Inequality \cite{McDiarmid.Survey1989} implies that for $\beta > 0$,
\begin{align*}
\mathbb{P} \Big[\vert \sum_{v=1}^{n} Z_v - \frac{n}{m} \sum_{u=1}^{m} {\mathbf{z}_{ij}^{*}}_{(u)} \vert \geq \frac{n}{m} \beta \Big] \leq 2 \exp \Big( \frac{ - n \beta^2  }    {2 m^2 \|\mathbf{z}_{ij}^{*}\|_{\infty}^2}  \Big),
\end{align*}
or equivalently,
\begin{align*}
& \mathbb{P} \Big[ \sum_{u=1}^{m} {\mathbf{z}_{ij}^{*}}_{(u)} - \beta \leq \frac{m}{n} \sum_{v=1}^{n} Z_v \leq \sum_{u=1}^{m} {\mathbf{z}_{ij}^{*}}_{(u)} + \beta \Big]  \\
&\qquad\qquad \geq 1 - 2 \exp \Big( \frac{ - n \beta^2  }    {2 m^2 \|\mathbf{z}_{ij}^{*}\|_{\infty}^2}  \Big).
\end{align*}
Taking the definition of $\beta = \sqrt{\frac{2 m^2 \| \mathbf{z}_{ij}^{*} \|_{\infty}^2}{|\mathbf{\Omega}_{ij}|} \log(\frac{1}{\delta})}$ yields the result.
\end{IEEEproof}


%
%
%




%

\begin{IEEEbiography}[{\includegraphics[width=1in,height=1.25in,clip,keepaspectratio]{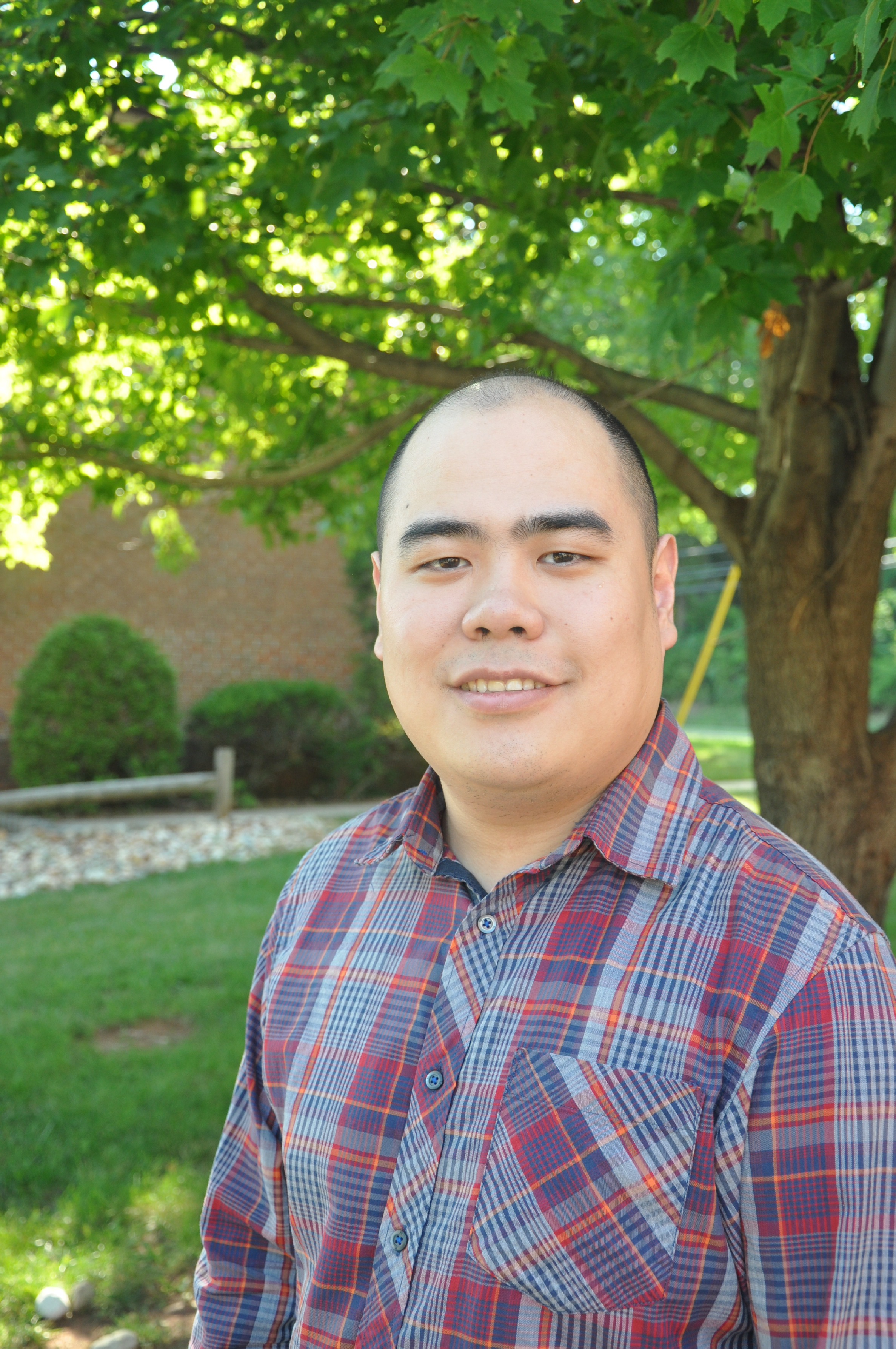}}]{Tong Wu}
received BE degree in Instrument Science and Engineering from Shanghai Jiao Tong University, China in 2009, and MS degree in Electrical Engineering from Duke University in 2011. Since September 2012, he has been working towards the PhD degree at the Department of Electrical and Computer Engineering, Rutgers, The State University of New Jersey. His research interests include high-dimensional data analysis, statistical signal processing, image and video processing, and machine learning.
\end{IEEEbiography}

\begin{IEEEbiography}[{\includegraphics[width=1in,height=1.25in,clip,keepaspectratio]{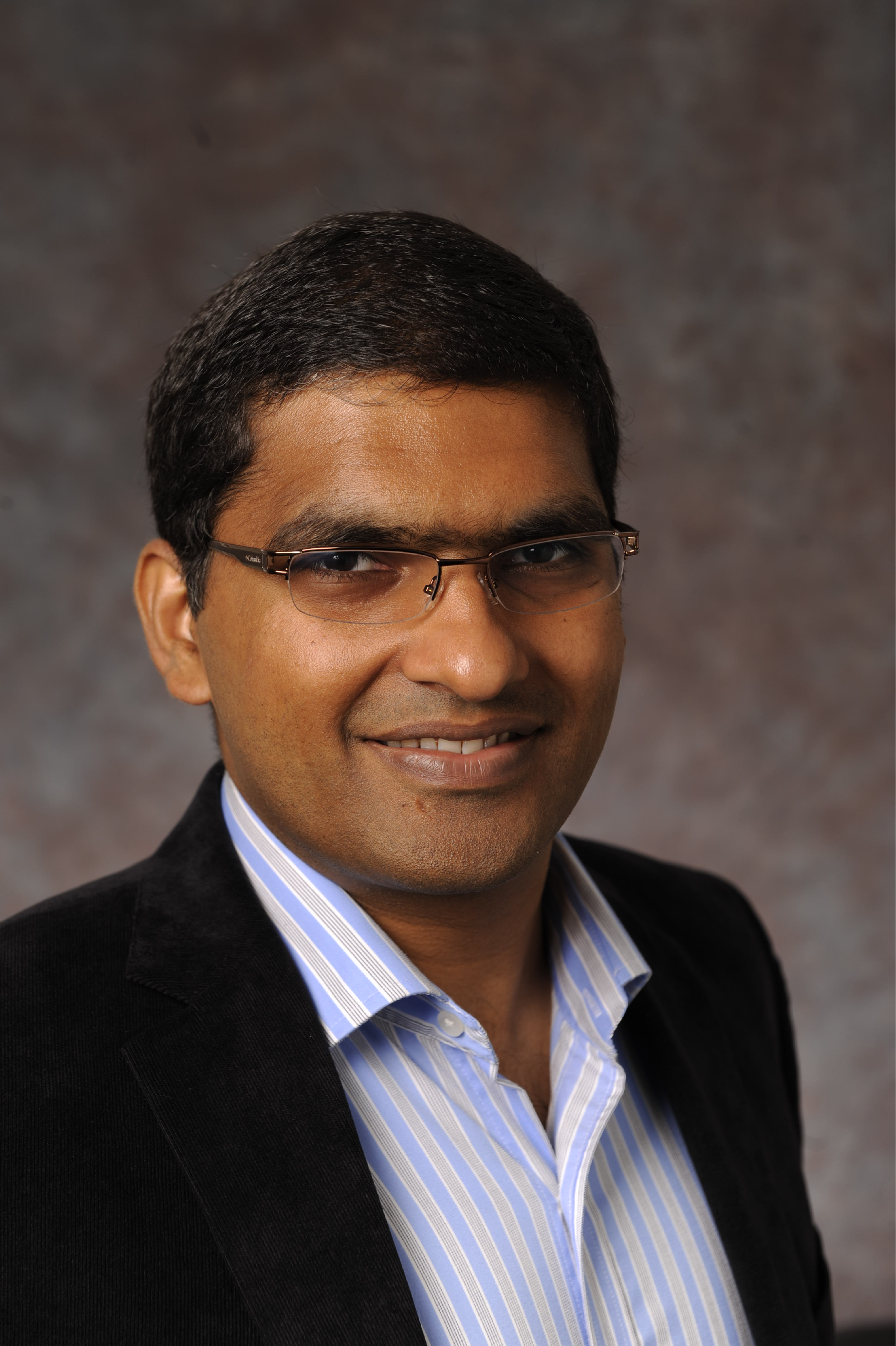}}]{Waheed U. Bajwa}
received BE (with Honors) degree in electrical engineering from the National University of Sciences and Technology, Pakistan in 2001, and MS and PhD degrees in electrical engineering from the University of Wisconsin-Madison in 2005 and 2009, respectively. He was a Postdoctoral Research Associate in the Program in Applied and Computational Mathematics at Princeton University from 2009 to 2010, and a Research Scientist in the Department of Electrical and Computer Engineering at Duke University from 2010 to 2011. He is currently an Assistant Professor in the Department of Electrical and Computer Engineering at Rutgers University. His research interests include harmonic analysis, high-dimensional statistics, machine learning, statistical signal processing, and wireless communications.

Dr. Bajwa has more than three years of industry experience, including a summer position at GE Global Research, Niskayuna, NY. He received the Best in Academics Gold Medal and President's Gold Medal in Electrical Engineering from the National University of Sciences and Technology in 2001, the Morgridge Distinguished Graduate Fellowship from the University of Wisconsin-Madison in 2003, the Army Research Office Young Investigator Award in 2014, and the National Science Foundation CAREER Award in 2015. He co-guest edited a special issue of Elsevier Physical Communication Journal on ``Compressive Sensing in Communications'' (2012), and co-chaired CPSWeek 2013 Workshop on Signal Processing Advances in Sensor Networks and IEEE GlobalSIP 2013 Symposium on New Sensing and Statistical Inference Methods. He currently serves as the Publicity and Publications Chair of IEEE CAMSAP 2015, and is an Associate Editor of the IEEE Signal Processing Letters.
\end{IEEEbiography}




\end{document}